\documentclass[letterpaper]{article}

\usepackage{uai2019}
\usepackage[margin=1in]{geometry}

\usepackage[utf8]{inputenc} 
\usepackage[T1]{fontenc}    
\usepackage{lmodern}
\usepackage{hyperref}      
\usepackage{url}            
\usepackage{booktabs}       
\usepackage{amsfonts}       
\usepackage{nicefrac}       
\usepackage{microtype}     
\usepackage{natbib}
\usepackage{bm}
\usepackage{soul}

\usepackage{algorithm}
\usepackage{algorithmic}

\usepackage[pdftex]{graphicx}
\usepackage{xcolor}

\usepackage{amsmath}
\usepackage{verbatim}

\newcommand{\R}{\mathbb{R}}

\newcommand{\Nor}{\mathcal{N}}

\usepackage{subcaption}

\usepackage[flushleft]{threeparttable}
\usepackage{makecell}

\newlength{\mzerolen}

\definecolor{dark-blue}{rgb}{0.15,0.15,0.4}
\definecolor{medium-blue}{rgb}{0,0,0.5}
\hypersetup{
   colorlinks, linkcolor={dark-blue},
   citecolor={dark-blue}, urlcolor={medium-blue}
}

\usepackage{times}

\title{Subspace Inference for Bayesian Deep Learning}

\author{
	Pavel Izmailov\thanks{\text{ }\text{ }Equal contribution.}\, $^1$,
	Wesley J. Maddox$^*$$^1$,
	Polina Kirichenko$^*$$^1$,
	Timur Garipov$^*$$^{4}$ \\
	\textbf{Dmitry Vetrov}$^{2,3}$,
	\textbf{Andrew Gordon Wilson}$^{1,5}$ \\
	$^1$Cornell University,
	$^2$Higher School of Economics,
	$^3$Samsung-HSE Laboratory, \\
	$^4$Samsung AI Center in Moscow,
	$^5$Courant Institute and Center for Data Science, New York University}

\begin{document}

\maketitle

\begin{abstract}

Bayesian inference was once a gold standard for learning with neural networks, providing accurate full predictive distributions and well calibrated uncertainty. 
However, scaling Bayesian inference techniques to deep neural networks is challenging due to the high dimensionality of the parameter space. 
In this paper, we construct low-dimensional subspaces of parameter space, such as the first principal components of the stochastic gradient 
descent (SGD) trajectory, which contain diverse sets of high performing models. 
In these subspaces, we are able to apply elliptical slice sampling and variational inference, which struggle in the full parameter space. 
We show that Bayesian model averaging over the induced posterior in these subspaces produces accurate predictions and well-calibrated predictive uncertainty for both regression and image classification.

\end{abstract}

\section{INTRODUCTION} \label{sec:intro}

Bayesian methods were once the state-of-the-art approach for inference with neural networks 
\citep{mackay2003information,neal1996bayesian}. However, the parameter spaces for modern
deep neural networks are extremely high dimensional, posing challenges to standard Bayesian 
inference procedures.

\begin{figure*}
	\centering
	\begin{subfigure}{0.22\textwidth}
		\centering
		\includegraphics[height=\textwidth]{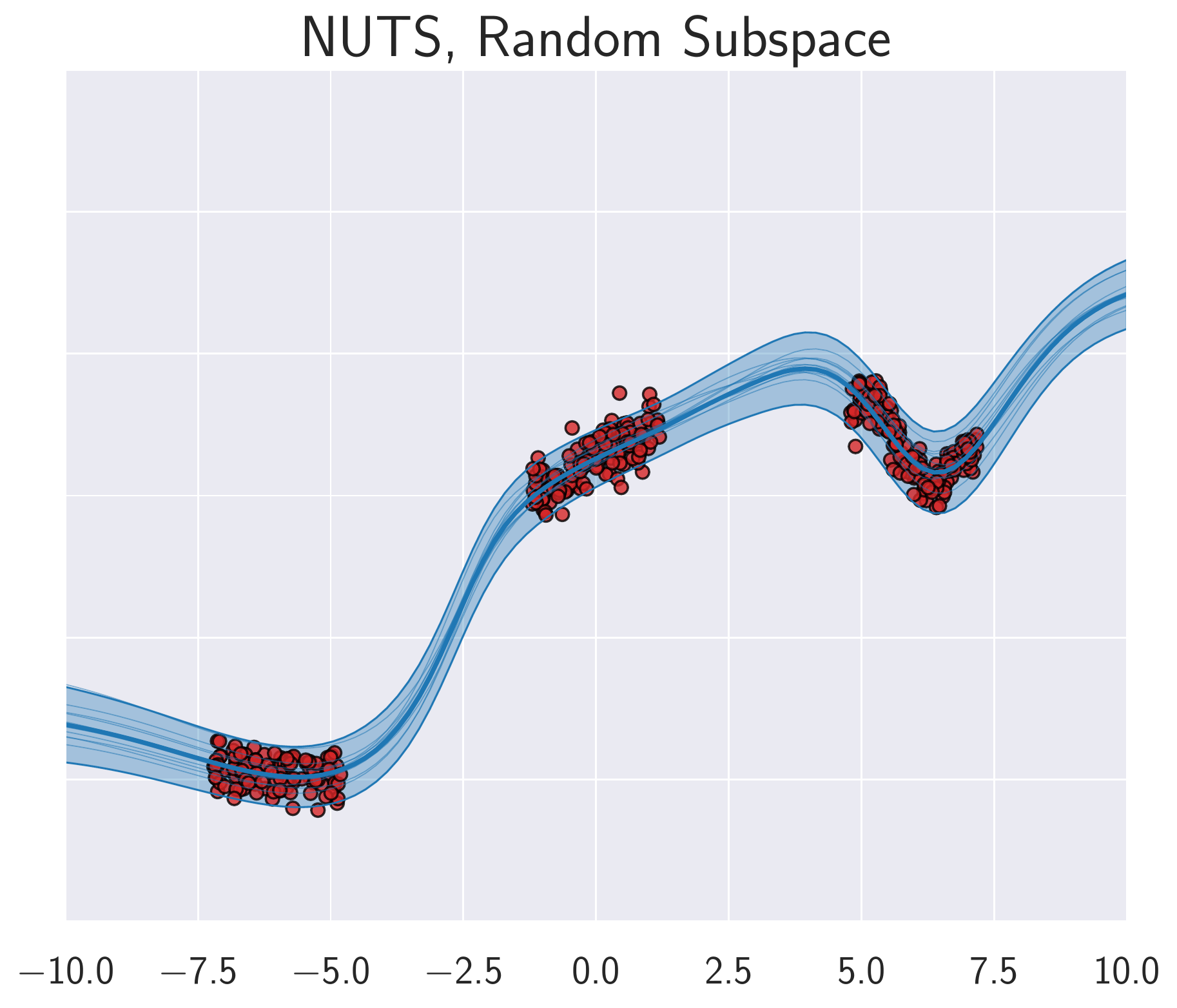}
        \caption{}
	\end{subfigure}
	\quad
	\begin{subfigure}{0.22\textwidth}
		\centering
		\includegraphics[height=\textwidth]{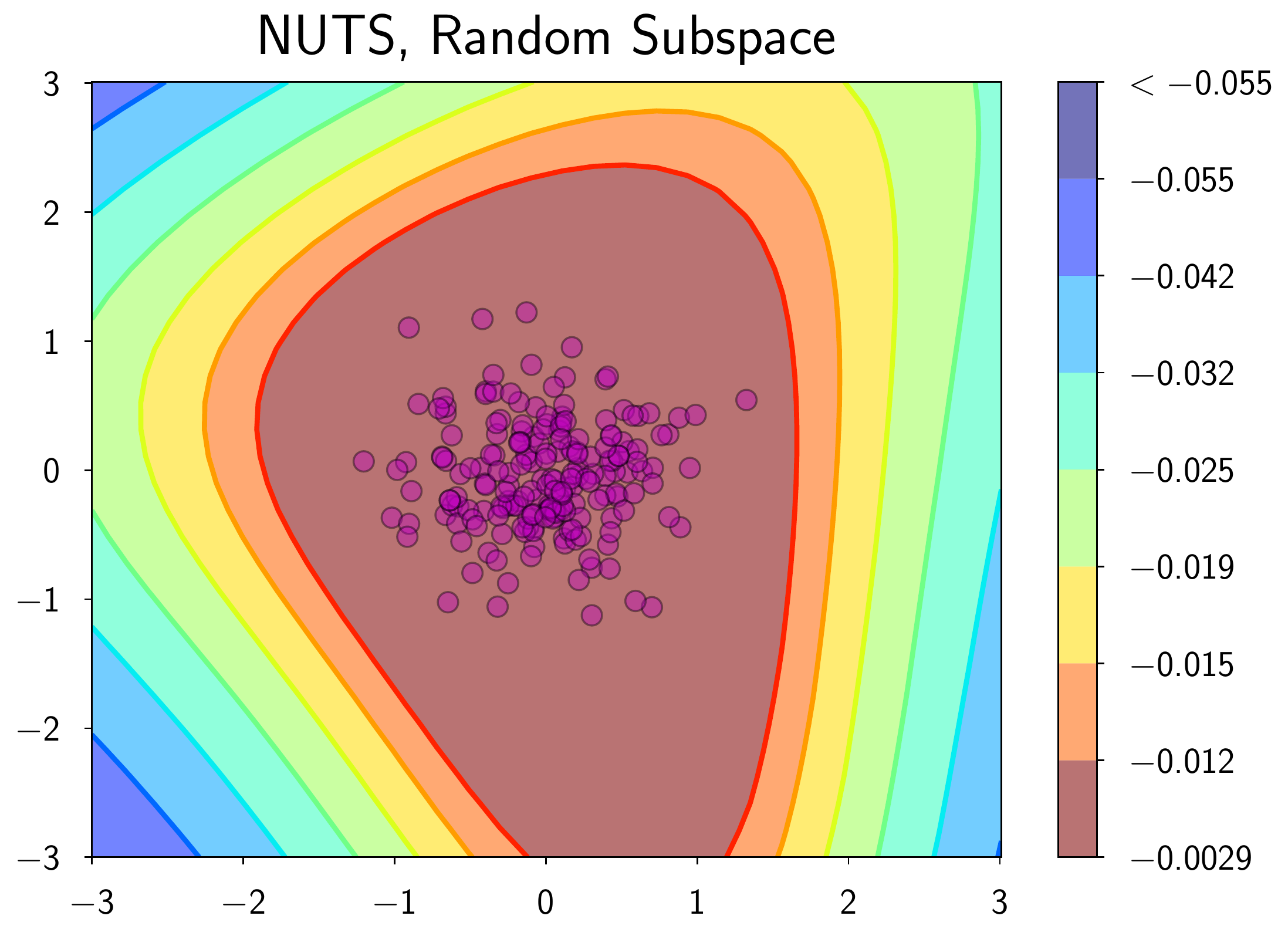}
        \caption{}
	\end{subfigure}
	\quad\quad
	\begin{subfigure}{0.22\textwidth}
		\centering
		\includegraphics[height=\textwidth]{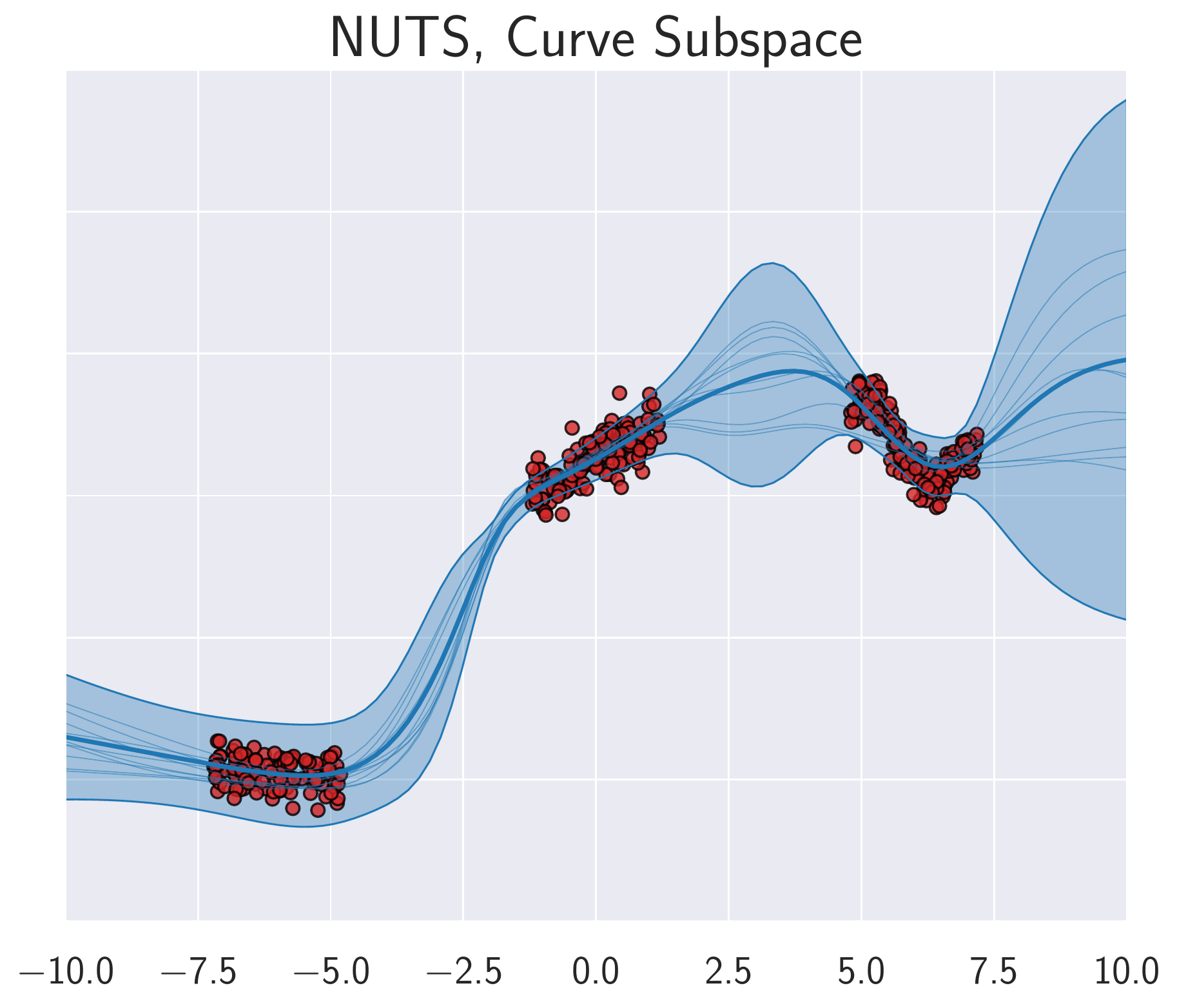}
        \caption{}
	\end{subfigure}
	\quad
	\begin{subfigure}{0.22\textwidth}
		\centering
		\includegraphics[height=\textwidth]{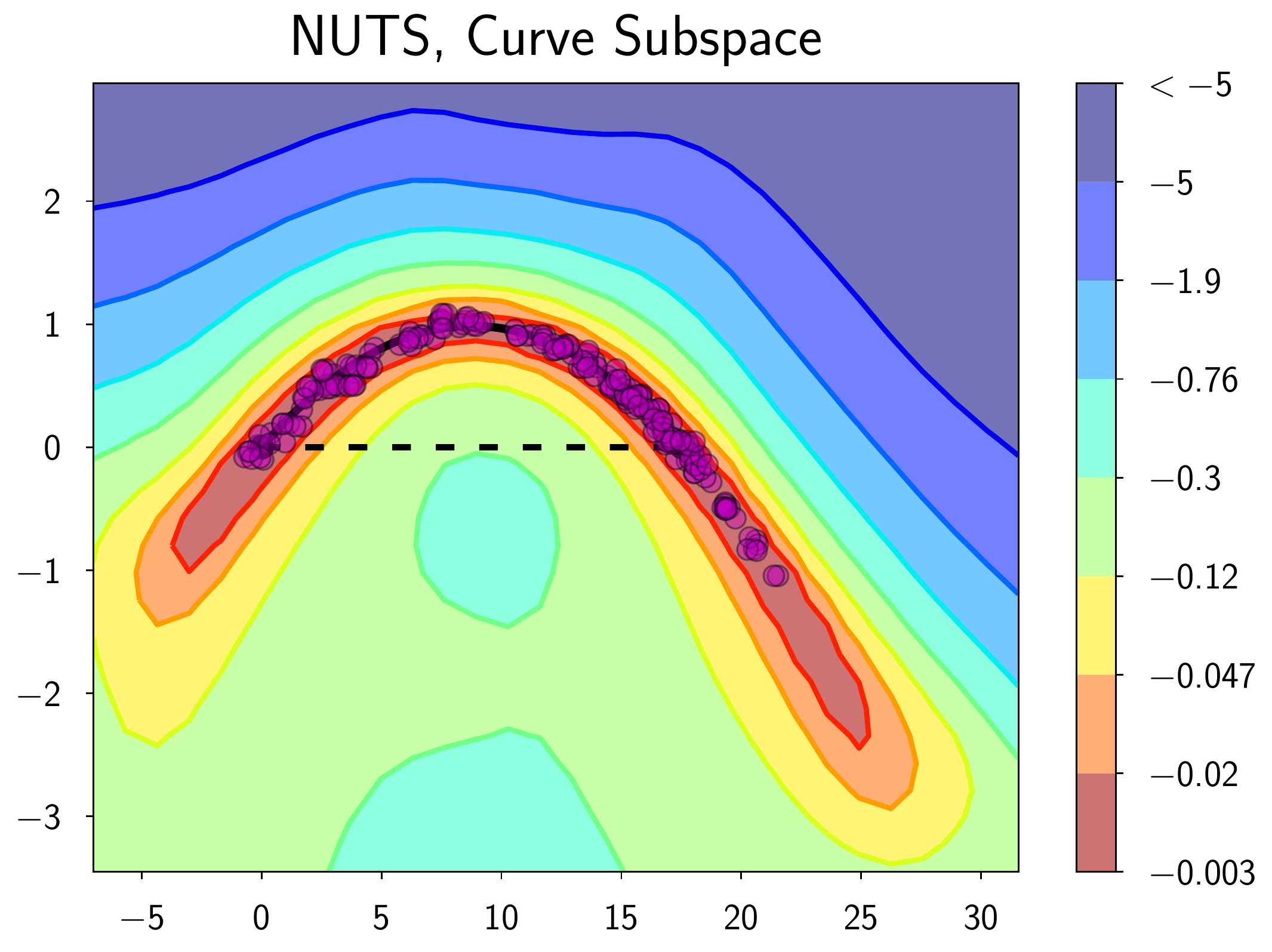}
        \caption{}
	\end{subfigure}
	\caption{Predictive distribution and samples in the parameter space for subspace inference 
    on a synthetic regression problem in a random subspace \textbf{(a, b)} 
    and subspace containing a near-constant loss (log posterior) curve between two 
    independently trained solutions \textbf{(c, d)} \citep[see][ for details]{fge}.
    On the plots \textbf{(a, c)} data points
    are shown with red circles, the shaded region represents the $3 \sigma$-region
    of the predictive distribution at each point, the predictive mean is shown with
    a thick blue line and sample trajectories are shown with thin blue lines.
    Panels \textbf{(b, d)} show the contour plots of the posterior log-density within the corresponding
    subspace; magenta circles represent samples from the posterior in the subspace.
    In the rich subspace containing the near-constant loss curve, the samples produce
    better uncertainty estimates and more diverse trajectories. We use a 
    small fully-connected network with $4$ hidden layers. See Section \ref{sec:toyreg}
    for more details.
    }
	\label{fig:intro}
	\vspace{-0.5cm}
\end{figure*}

In this paper, we propose a different
approach to approximate Bayesian inference in deep learning models: we
design a low-dimensional subspace $\mathcal{S}$ of the weight space and
perform posterior inference over the parameters within this subspace. We call this approach \emph{Subspace Inference} (SI).\footnote{PyTorch code is available at \url{https://github.com/wjmaddox/drbayes}.}

It is our contention that the subspace $\mathcal{S}$ can be chosen to contain
a diverse variety of representations, corresponding to different high quality 
predictions, over which Bayesian model averaging leads to accuracy gains and
well-calibrated uncertainties.

In Figure \ref{fig:intro}, we visualize the samples from the approximate posterior
and the corresponding predictive distributions in performing subspace inference for
a ten-dimensional random subspace, and a rich two-dimensional subspace containing a low-loss curve between two
independently trained SGD solutions \citep[see][]{fge} on a synthetic one-dimensional
regression problem. As we can see, the predictive distribution corresponding to 
a random subspace does not capture a diverse set of possible functions required for greater uncertainty away from the data, but 
sampling from the posterior in the rich curve subspace provides meaningful uncertainty over functions.

Our paper is structured as follows. We begin with a discussion of related work in 
Section \ref{sec:motivation}. In Section \ref{sec:inf}, we describe the proposed method for inference
in low-dimensional subspaces of the parameter space. In Section~\ref{sec:subspace}, we discuss possible choices 
of the low-dimensional subspaces. In particular, we consider random subspaces, subspaces 
corresponding to the first principal components of the SGD trajectory \citep{maddox2019simple}, 
and subspaces containing low-loss curves between independently trained
solutions \citep{fge}. 

We analyze the effects of using different subspaces and approximate
inference methods, by visualizing uncertainty on a regression problem in 
Section \ref{sec:toyreg}.  We then apply the proposed method to a range 
of UCI regression datasets in Section \ref{sec:ucireg}, as well as CIFAR-10 and 
CIFAR-100 classification problems in Section \ref{sec:cifar}, achieving consistently
strong performance in terms of both test accuracy and likelihood. Although the dimensionality 
of the weight space for modern neural networks is extraordinarily large, we show that surprisingly
low dimensional subspaces contain a rich diversity of representations. 
For example, we can construct \textit{5 dimensional} subspaces where Bayesian model averaging leads to 
notable performance gains on a 
\emph{36 million dimensional} WideResNet trained on CIFAR-100.
    
    We summarize subspace inference in Algorithm \ref{alg:inf}. We note that this procedure uses three modular steps:
    (1) construct a subspace; (2) posterior inference in the subspace; and (3) form a Bayesian model average. Different design choices are possible for each step. For example, choices for the subspace include a random subspace, a PCA subspace, or a mode connected subspace. Many other choices are also possible. For posterior inference, one can use deterministic approximations over the parameters in the subspace, such as a variational method, or MCMC.

\begin{algorithm}
	\caption{\label{alg:inf} Bayesian Subspace Inference}
	\begin{algorithmic}
		\STATE \textbf{Input:} data, $\mathcal{D};$ model, $\mathcal{M};$
		\STATE 1. Construct subspace, i.e. using Alg. \ref{alg:swag}, Section \ref{sec:subspace}
		\STATE 2. Posterior inference within subspace (Section \ref{sec:inf})
		\STATE 3.\ Form a Bayesian model average (Section \ref{sec:sampling}, \ref{sec:approxinf})
	\end{algorithmic}
\end{algorithm}

\section{RELATED WORK} \label{sec:motivation}

\citet{maddox2019simple} proposed SWAG, which forms an approximate Gaussian posterior over neural network weights, with a mean and low rank plus diagonal covariance matrix formed from a partial trajectory of the SGD iterates with a modified learning rate schedule. SWAG provides scalable Bayesian model averaging, with compelling accuracy and calibration results on CIFAR and ImageNet. The low-rank part of the SWAG covariance defines a distribution over a low-dimensional subspace spanned by the first principal components of the SGD iterates. 

\citet{silva2015bayesian} consider the related problem of Bayesian inference using projected methods for constrained latent variable models, with applications to probabilistic PCA 
\citep{roweis1998algorithms, bishop1999bayesian}.

\citet{pradier2019projected} propose to perform variational inference (VI) in a subspace formed by an auto-encoder trained on a set of models generated from fast geometric ensembling \citep{fge}; this approach requires training several models and fitting an auto-encoder, leading to limited scalability.

Similarly, \citet{karaletsos2018probabilistic} propose to use a meta-prior in a low-dimensional space to perform variational inference for BNNs. This approach can be viewed as a generalization of hyper-networks \citep{ha2016hypernetworks}. Alternatively, both \citet{titsias2017learning} and \citet{krueger2017bayesian} propose Bayesian versions of hyper-networks to store meta-models of parameters.

\cite{patra2018constrained} provide theoretical guarantees for Bayesian inference in the setting of constrained posteriors.
Their method samples from the unconstrained posterior before using a mapping into the constrained parameter space.
In their setting, the constraints are chosen a priori; on the other hand, we choose the constraints (e.g.\ the subspace) after performing unconstrained inference via SGD.

Bayesian coresets \citep{huggins2016coresets} use a weighted combination of the full dataset and Bayesian compressed regression \citep{guhaniyogi2015bayesian} uses random projections of the data inputs in linear regression settings; both are designed for the purpose of efficient inference, but unlike our subspace inference, these methods operate solely in data space, rather than in parameter space.

\section{INFERENCE WITHIN A SUBSPACE} \label{sec:inf}

In this section we discuss how to perform Bayesian inference within a given subspace of a 
neural network. In Section \ref{sec:subspace} we will propose approaches for effectively 
constructing such subspaces.

\subsection{MODEL DEFINITION}
\label{sec:model}

We consider a model, $\mathcal{M}$, with weight parameters $w \in \mathbb{R}^d$. The model has an associated likelihood for the dataset, $\mathcal{D}$, given by $p_\mathcal{M}(\mathcal{D} | w).$ 

We perform inference in a $K$-dimensional subspace $\mathcal{S}$ defined by 
\begin{align}\label{eq:subspace}
    \mathcal{S} &= \{w \vert w = \hat w + z_1 v_1 + \ldots z_K v_K\} \notag \\ 
    &= \{w \vert w = \hat w + P z\},
\end{align}
where $P = (v_1^{\top}, \ldots, v_K^{\top}) \in \R^{d \times K}$, $\hat w \in \R^{d}$, $z = (z_1, \ldots, z_K)^{\top} \in \mathbb{R}^K$.
With a fixed $\hat{w}$ and projection matrix $P$, which assign the subspace, 
the free parameters of the model, over which we perform inference, are now simply $z \in \mathbb{R}^K$.
We describe choices for $\hat w$ and $P$ in Section \ref{sec:subspace}.

\begin{figure}[!h]
\vspace{-0.2in}
	\centering
	\begin{subfigure}{0.27\textwidth}
		\centering
		\includegraphics[width=\textwidth,trim={0.45cm 0.4cm .45cm 0.25cm},clip]{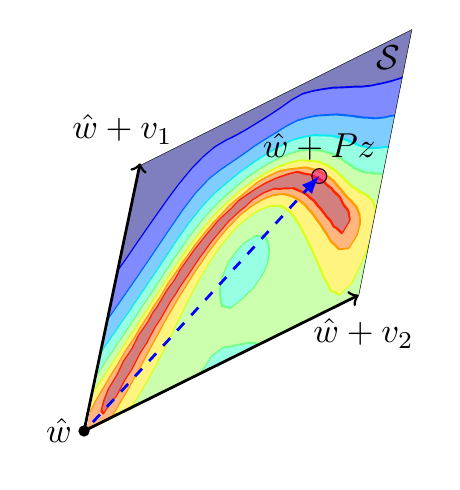}
	\end{subfigure}
	\caption{Illustration of subspace $\mathcal{S}$ with shift vector $\hat w$ and basis vectors $v_1, v_2$, with a contour plot of the posterior log-density over parameters $z$.
	} 
	\label{fig:subspace}
	\vspace{-0.05in}
\end{figure}

The new model has the likelihood function:
\begin{align}
    p(\mathcal{D} \vert z) = 
    p_\mathcal{M}(\mathcal{D} \vert w = \hat w +  P z),
    \label{eq:like}
\end{align}
where the right-hand side represents the likelihood for the model, $\mathcal{M}$, with parameters $\hat w + P z$  and data $\mathcal{D}$. 
We can then perform Bayesian inference over the low-dimensional subspace parameters $z$.
We illustrate the subspace parameterization as well as the posterior log-density over parameters $z$ in Figure \ref{fig:subspace}.

We emphasize that the new model \eqref{eq:like} is \textit{not} a reparameterization of the original model, as the 
mapping from the full parameter space to the subspace is not invertible. 
For this reason, we consider the subspace model parameterized by $z$ as a \textit{different model} that shares many functional properties with the original model $\mathcal M$ (see Section \ref{sec:loss_measure} for an extended discussion). 
We discuss potential benefits of using the subspace model \eqref{eq:like} in Section \ref{sec:appendix_benefits}.

\subsection{BAYESIAN MODEL AVERAGING}\label{sec:sampling}

We can sample from an induced posterior over the weight parameters in the original space by first sampling
from the posterior over the parameters in the subspace $\tilde{z} \sim p(z|\mathcal{D})$, using an approximate 
inference method of choice, and then transforming
those samples into the original space as $\tilde w = \hat w + P \tilde z$.

To perform Bayesian model averaging on new test data points, $\mathcal{D}^*$, we can compute a Monte Carlo estimate of the integral
\begin{align}
\label{eq:bmaint}
p(\mathcal{D}^*|\mathcal{D}) &= \int p_\mathcal{M}(\mathcal{D}^*|\tilde w = \hat w +  P z) p(z |\mathcal{D}) dz  \\
&\approx  \frac{1}{J} \sum_{j=1}^J p_\mathcal{M}(\mathcal{D}^*|\tilde w = \hat w +  P \tilde{z}_j) \,, \tilde{z}_j \sim p(z | \mathcal{D}) \,. \notag
\end{align}
Using the Monte Carlo estimate of the integral in \eqref{eq:bmaint} produces mixtures of Gaussian predictive distributions for regression tasks with Gaussian likelihoods, and categorical distributions for classification tasks.

\subsection{APPROXIMATE INFERENCE PROCEDURES}\label{sec:approxinf}

Our goal is to approximate the posterior $p(z | \mathcal{D})$ over the free parameters
$z$ in the subspace $\mathcal{S}$, in order to perform a Bayesian model average.
As we can set the number of parameters, $K$, to be much smaller than the dimensionality $d$ of the full parameter space, performing Bayesian inference becomes considerably more tractable in the subspace. 
We can make use of a wide range of 
approximate inference procedures, even if we are working with a large modern 
neural network.

In particular, we can use powerful and exact full-batch MCMC methods to approximately sample from 
$p(z | \mathcal{D})$, such as Hamiltonian Monte Carlo (HMC) \citep{neal2011mcmc}  or 
elliptical slice sampling (ESS) \citep{murray2010elliptical}. ESS relies heavily on prior sampling, initially
introduced for sampling from posteriors with informative Gaussian process priors; however, ESS has 
special relevance for subspace inference, since these subspaces are specifically constructed to be 
centred on good regions of the loss, where a wide range of priors will provide reasonable samples.
Alternatively, we can perform a deterministic approximation 
$q(z | \mathcal{D}) \approx p(z | \mathcal{D})$, for example 
using Laplace or a variational approach, and then sample from $q$. 
The low dimensionality of the problem allows us to choose very flexible variational families such as RealNVP \citep{realnvp}
to approximate the posterior.

Ultimately, the inference procedure is an experimental design choice, and we are free to use a wide range of approximate 
inference techniques. 

\subsection{PRIOR CHOICE} \label{sec:priorchoice}

There is a significant practical difference between \emph{Bayesian model averaging} (Section \ref{sec:sampling}) and standard training
(regularized maximum likelihood estimation) for a range of priors $p(z)$, including vague priors.
The exact specification of the prior itself, $p(z)$, if sufficiently diffuse, is not crucial for good performance or for
the benefits of Bayesian model averaging in deep learning. What matters is not the prior over parameters in isolation, but how 
this prior interacts with the functional form of the model.
The neural network induces a structured prior distribution over functions, even when combined with a vague prior over its
parameters. For subspace inference specifically, the subspace is constructed to be centred on a good region of the loss, such that a wide 
range of priors will provide coverage for weights corresponding to high performing networks. 
We discuss reasonable choices of priors for various subspaces in Section \ref{sec:subspace}.

\subsection{PREVENTING POSTERIOR CONCENTRATION WITH FIXED TEMPERATURE POSTERIORS}\label{sec:temp} 

In the model proposed in Section \ref{sec:model}, there are only $K \ll N$ parameters
as opposed to $d \gg N$ parameters in the full weight space, while the number of observed
data points $N$ is constant. In this setting, 
the posterior can overly concentrate around the maximum likelihood estimate (MLE), 
becoming too constrained by the data, leading to overconfident uncertainty estimates.

To address the issue of premature posterior concentration in the subspace, we propose to 
introduce a temperature hyperparameter that scales the likelihood. In particular, 
we use the {\it tempered} posterior:
\begin{align}
	p_T(z | \mathcal{D}) \propto 
    {\underbrace{p(\mathcal{D}| z)}_{\text{likelihood}}}^{1/T} \underbrace{p(z)}_{\text{prior}}.
	\label{eq:tempered_post}
\end{align}
When $T = 1$ the true posterior is recovered, and as $T \rightarrow \infty$, 
the tempered posterior $p_T(z | \mathcal{D})$ approaches the prior $p(z)$. 

The temperature $T$ is a hyper-parameter that can be determined through cross-validation. We study the effect of temperature on the performance of subspace inference in Section \ref{sec:temp_effect}. When the temperature is close to $1$ the posterior concentrates around the MLE and subspace inference fails to improve upon maximum likelihood training. As $T$ becomes large, subspace inference produces increasingly less confident predictions. In Section \ref{sec:temp_effect}, good performance can be achieved with a broad range of $T$.

Tempered posteriors are often used in Bayesian inference algorithms 
to enhance multi-modal explorations \citep[e.g.,][]{geyer1991markov,neal1996sampling}. 
Similarly, \citet{watanabe2013widely} uses a tempered posterior to recover an expected generalization error of Bayesian models.

\section{SUBSPACE CONSTRUCTION} \label{sec:subspace}

In the previous section we showed how to perform inference in a given subspace 
$\mathcal{S}$. We now discuss various ways to construct $\mathcal{S}$.

\subsection{RANDOM SUBSPACES}\label{sec:random}

To construct a simple random subspace, $\mathcal{S}$,  
we draw $K$ random $v_1, \ldots, v_K \sim \Nor(0, I_p)$
in the weight space. We then rescale each of the vectors to have norm $1$.
Random subspaces require only drawing $Kp$ random normal numbers and so are quick to generate and form, but contain little information about the model.
In related work, \citet{li_measuring_2018} train networks from scratch in a random subspace without a shift vector, 
requiring projections into much higher dimensions than are considered in this paper.

We use the weights of a network pre-trained with stochastic weight averaging (SWA) \citep{swa} as the shift vector $\hat w = w_\text{SWA}$. 
In particular, we run SGD with a high constant learning rate from a pre-trained solution, and form the average
$w_\text{SWA} =  \frac 1 L \sum_i w_i$ from the SGD iterates $w_i$.

Since the log likelihoods as a function of neural network parameters 
for random subspaces appear approximately quadratic \citep{swa},
and the subspace is centred on a good solution, 
a reasonable prior for $p(z)$ is $\Nor(0, \sigma_K^2 I_K)$.

\subsection{PCA OF THE SGD TRAJECTORY}\label{sec:pca}

Intuitively, we want the subspace $\mathcal{S}$ over which we perform inference to 
(1) contain a diverse 
set of models that produce meaningfully different predictions and (2) be cheap to construct. 
\citet{fge} and \citet{swa} argue that the
subspace spanned by the SGD trajectory satisfies both (1) and (2). 
They run SGD starting from a pre-trained solution with a high constant learning rate
and then ensemble predictions or average the weights of the iterates. 
Further, \citet{maddox2019simple}
showed fitting the SGD iterates with a Gaussian distribution with a 
low-rank plus diagonal covariance for scalable Bayesian
model averaging provides well-calibrated uncertainty estimates.
Finally, \citet{li_visualizing_2018} and \citet{maddox2019simple} used the
first few PCA components of the SGD trajectory for loss surface visualization.
These observations motivate inference \emph{directly}
in the subspace spanned by the SGD trajectory.

We propose to use the first few PCA components $v_i$ of the SGD
trajectory to define the basis of the subspace.
As in \citet{swa}, we run SGD with a high
constant learning rate from a pre-trained solution and capture snapshots 
$w_i$ of weights at the end of each of $T$ epochs. We store the deviations
$a_i = w_\text{SWA} - w_i$ for the last $M$ epochs. The number $M$ here is determined
by the amount of memory we can use.\footnote{We use $M=20$ in our experiments. To side-step any memory issues, 
we could use any online PCA technique instead, such as 
frequent directions \citep{ghashami2016frequent}.}
We
then run PCA based on randomized SVD \citep{halko2011finding}\footnote{Implemented in sklearn.decomposition.TruncatedSVD.}
on the matrix $A$ comprised of vectors $a_1, \ldots, a_M$ and use the first $K$ principal components
$v_1, \ldots, v_K$ to define the subspace \eqref{eq:subspace}. 
Like for the random subspace, we use the SWA solution \citep{swa} for the shift vector, $\hat{w}.$
We summarize this procedure in Algorithm \ref{alg:swag}.
\begin{minipage}[t]{0.5\textwidth}

	\centering
	\begin{algorithm}[H]
		\caption{Subspace Construction with PCA}\label{alg:swag}
		\begin{algorithmic}
			\STATE $w_0$:~pretrained weights; $\eta$:~learning rate; 
			$T$:~number of steps; 
			$c$:~moment update frequency; 
			$M$:~maximum number of columns in deviation matrix; 
			$K$:~rank of PCA approximation;
			$P$:~projection matrix for subspace
			\STATE $w_{\text{SWA}} \leftarrow w_0$\hfill [Initialize mean]
			\STATE \textbf{for} $i \leftarrow 1, 2, ..., T$ \textbf{do}
			\STATE ~~~~$w_i \leftarrow  w_{i-1} - \eta \nabla_{w} \mathcal{L}(w_{i-1})$\hfill [SGD update]
			\STATE ~~~~\textbf{if}~$\mathtt{MOD}(i,c) = 0$ \textbf{then}
			\STATE ~~~~~~~~$n \leftarrow i/c$\hfill [Number of models]
			\STATE ~~~~~~~~$w_{\text{SWA}} \leftarrow \dfrac{n w_{\text{SWA}} + w_{i}}{n+1}$\hfill [Update mean]
			\STATE ~~~~~~~~\textbf{if} $\mathtt{NUM\_COLS}(A) = M$ \textbf{then}
			\STATE ~~~~~~~~~~~~$\mathtt{REMOVE\_COL}(A[:, 1])$
			\STATE ~~~~~~~~$\mathtt{APPEND\_COL}(A, w_i - \overline{w})$\hfill [Store deviation] 
			\STATE  $U, S, V^{\top} \leftarrow \text{SVD}(A)$ \hfill [Truncated SVD]
			\STATE \textbf{return} $\hat w = w_{\text{SWA}},~~ P=SV^{\top}$
		\end{algorithmic}
	\end{algorithm}
\end{minipage}

\quad

\citet{maddox2019simple} showed empirically that the log likelihood in the subspace looks locally approximately quadratic,
and so a reasonable choice of prior $p(z)$ is $\Nor(0,\sigma_K I_K)$, when scaling PCA vectors 
$v_i$ to have norms proportional to the singular values of the matrix $A$ as in Algorithm \ref{alg:swag}. We note that this
prior would be centred around a set of good solutions because of the shift parameter $\hat{w}$ in constructing the subspace.

\paragraph{Relationship to Eigenvalues of the Hessian}
\citet{li_visualizing_2018} and \citet{gurari2019gradient} argue that the first principal
components of the SGD trajectory correspond to the top eigenvectors of the 
Hessian of the loss, and that these eigenvectors change slowly during training. 
This observation suggests that these principal components captures many of the sharp directions of the loss surface,
corresponding to large Hessian eigenvalues.
We expect then that our PCA subspace should include variation in the type of functions that it contains.
See Appendix \ref{sec:appendix_eigen} for more details as well as a computation of Hessian and Fisher eigenvalues 
through a GPU accelerated Lanczos algorithm \citep{gardner2018gpytorch}.

\subsection{CURVE SUBSPACES}\label{sec:curve}

\citet{fge} proposed a method to find paths of near-constant low loss (and consequently
high posterior density) in the weight space between converged SGD solutions starting 
from different random initializations. These curves lie in $2$-dimensional subspaces of the weight space. 
We visualize the loss surface in such a space for a synthetic regression problem
in Figure~\ref{fig:intro}~(d). 
This \emph{curve subspace} provides an example of a rich subspace containing 
diverse high performing models, and stress-tests the inference procedure for
effectively exploring a highly non-Gaussian distribution.

To parameterize the curve subspace we set $\hat w = (w_1 + w_0) / 2$, 
$v_1 = (w_0 - \hat w) / \|w_0 - \hat w\|, v_2 = (w_{1/2} - \hat w) / \|w_{1/2} - \hat w\|$, where $w_0$ and $w_1$ are the endpoints, and $w_{1/2}$ is the 
midpoint of the curve. 

In this case, the posterior in the subspace is clearly non-Gaussian. However, a vague but centred Gaussian prior $\mathcal{N}(0, \sigma^2_K I)$ is 
reasonable as a simple choice with our parameterization of the curve subspace. 

\subsection{COMPUTATIONAL COST OF SUBSPACE CONSTRUCTION}
We consider the cost of constructing each of the subspaces described in Section \ref{sec:random}-\ref{sec:curve}.
We note that constructing any subspace in our approach is a \emph{one-time} computation.

The random subspace (Section \ref{sec:random}) is virtually free to construct, as it only requires sampling $K$ independent Gaussian vectors. 

To construct the PCA subspace (Section \ref{sec:pca}), we run SVD on the deviation matrix $A$, which is a \emph{one-time} computation and is very fast. 
In particular, exact SVD takes $\mathcal O(\min(M^2 d, d^2 M))$, while randomized SVD takes $\mathcal O(Md \log{K} + (M+d) K^2)$ (see Section 1.4.1 of \citet{halko2011finding}). For our largest examples, while the number of parameters in the model $d$ is on the order of $10^7$, $M$ is on the order of $10^2$; thus, taking the exact SVD is linear in the number of parameters. 
For example, using standard hardware (a Dell XPS-15 laptop, with Intel core i7, 16Gb RAM), it takes 4 minutes to perform \textit{exact} SVD on our largest model, WideResNet on CIFAR-100, with 36 million parameters. By comparison, it takes approximately eight hours on an NVIDIA 1080Ti GPU to train the same WideResNet on CIFAR-100 to completion.

The curve subspace (Section \ref{sec:curve}) is the most expensive to construct, as it requires pre-training the two solutions corresponding to the 
endpoints of the curve, and then running the curve-finding procedure in \citet{fge},
which in total is roughly $3 \times $ the cost of training a single DNN of the same architecture.

Constructing the subspace is in general very fast and readily applicable to large deep networks, with minimal
overhead compared to standard training.

\section{EXPERIMENTS}
\label{sec:exps}

We evaluate subspace inference empirically using the random, PCA, and curve subspace construction methods
discussed in Section \ref{sec:subspace}, in conjunction with the 
approximate posterior inference methods discussed in Section \ref{sec:approxinf}. 
In particular, we experiment with the No-U-Turn-Sampler (NUTS) \citep{hoffman2014no}, Elliptical Slice Sampling (ESS) \citep{murray2010elliptical} 
and Variational Inference with fully factorized Gaussian approximation family (VI) and Real-valued Non-Volume Preserving flow (RealNVP) \citep{realnvp} family. 
Section \ref{sec:appendix_approxinf} contains more details on each of the approximate inference methods. We use the priors for $p(z)$ discussed in Section \ref{sec:subspace}.

We show that approximate Bayesian inference 
within a subspace provides good predictive uncertainties on regression problems, first visually in Section \ref{sec:toyreg} and then quantitatively on 
UCI datasets in Section \ref{sec:ucireg}. 

We then apply subspace inference to large-scale image classification on CIFAR-10 and CIFAR-100 and obtain results competitive with state-of-the-art scalable Bayesian deep learning methods.

\subsection{VISUALIZING REGRESSION UNCERTAINTY}
\label{sec:toyreg}

\begin{figure*}
	\centering
	\begin{subfigure}{0.28\textwidth}
		\centering
		\includegraphics[width=\textwidth]{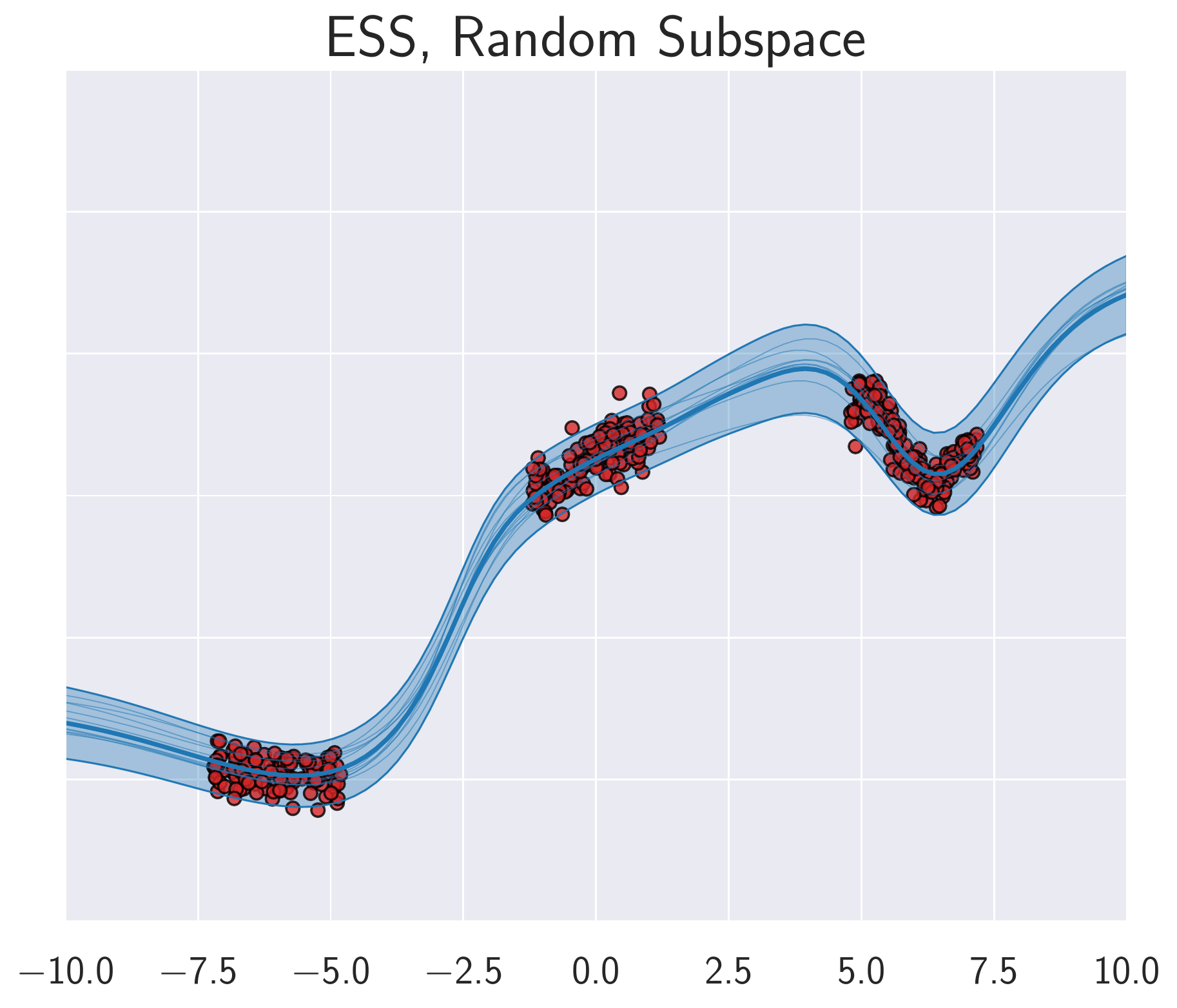}
	\end{subfigure}
	\quad\quad
	\begin{subfigure}{0.28\textwidth}
		\centering
		\includegraphics[width=\textwidth]{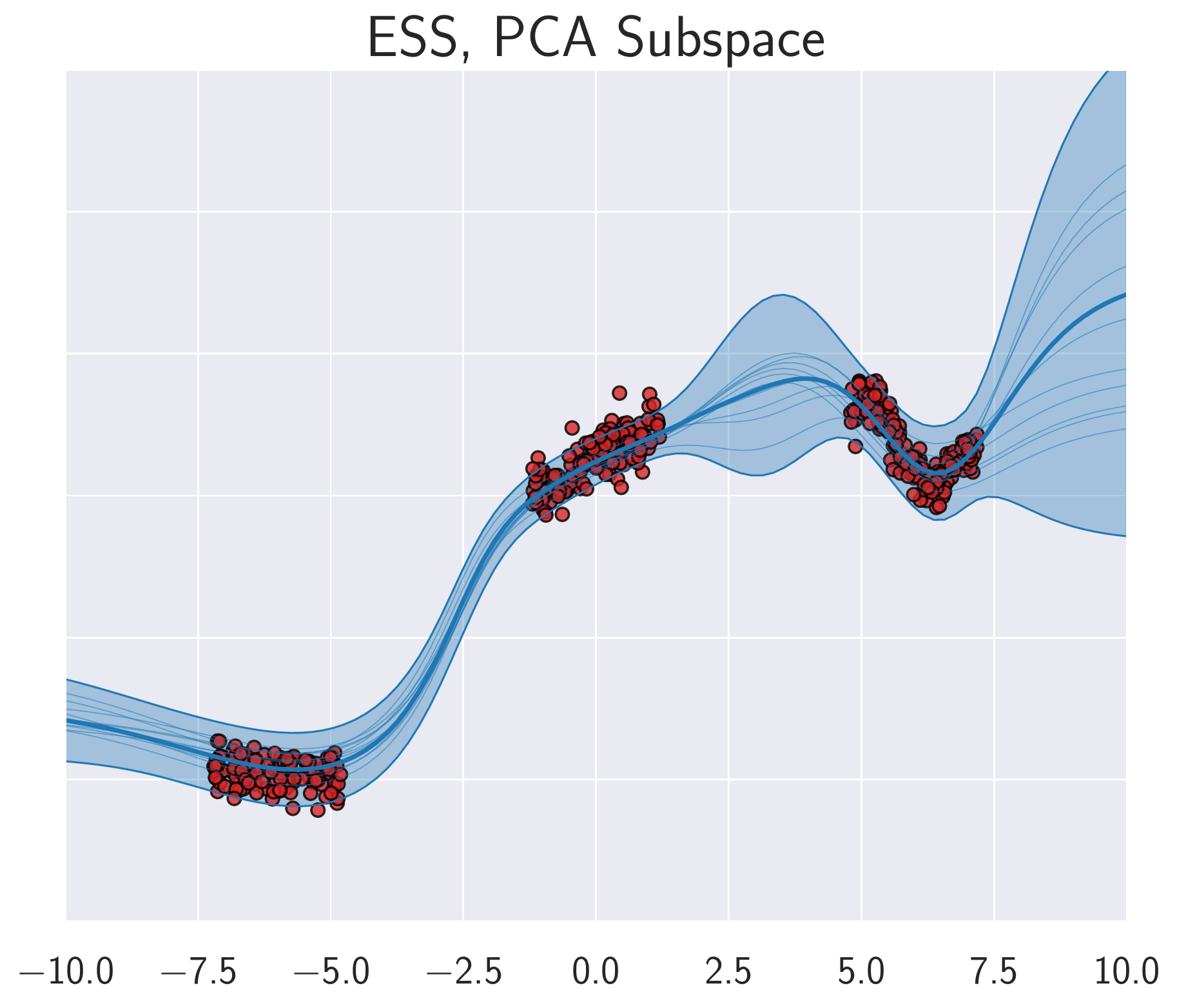}
	\end{subfigure}
	\quad\quad
	\begin{subfigure}{0.28\textwidth}
		\centering
		\includegraphics[width=\textwidth]{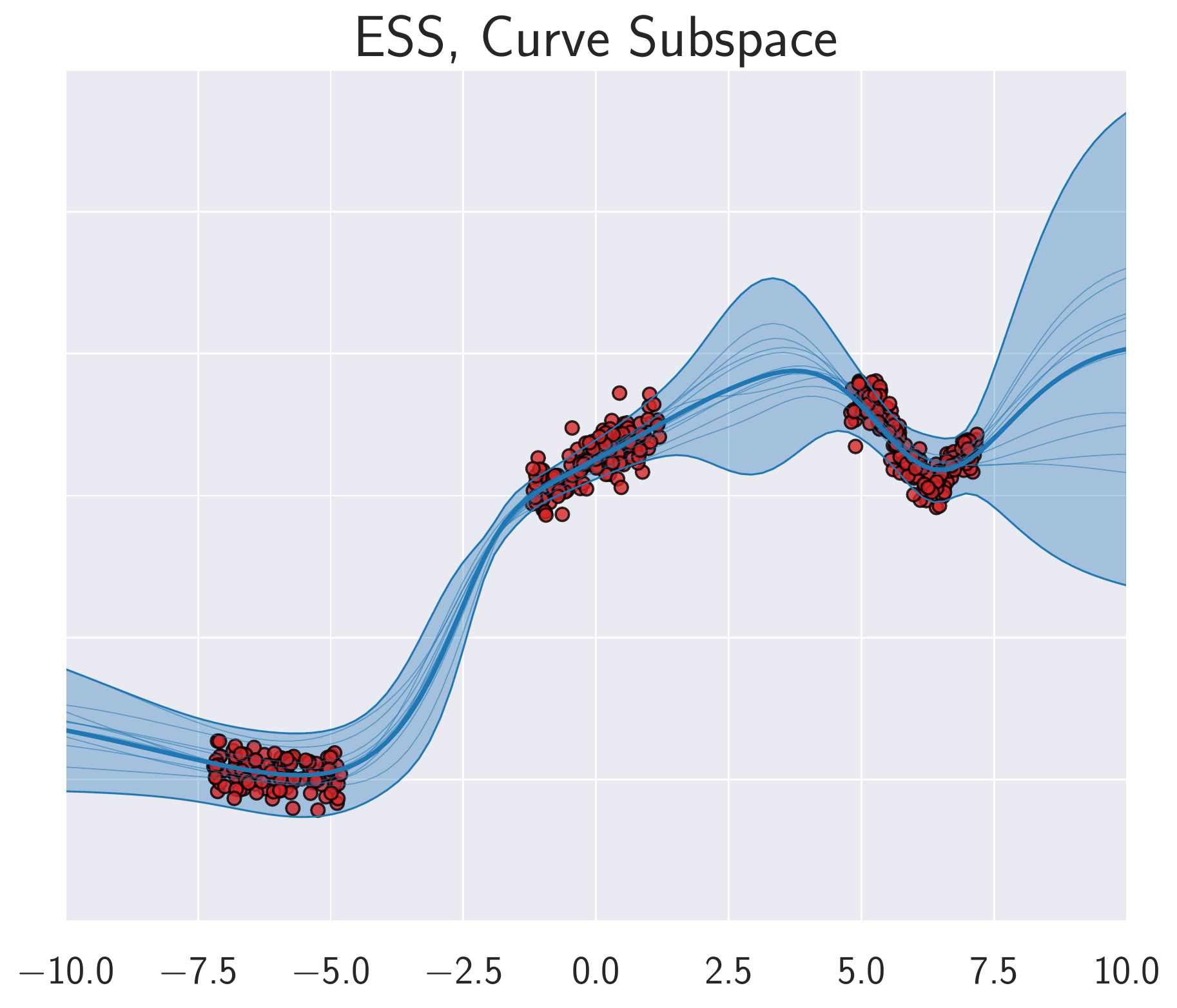}
	\end{subfigure}
	\vspace{0.3cm}
	
	\begin{subfigure}{0.28\textwidth}
		\centering
		\includegraphics[width=\textwidth]{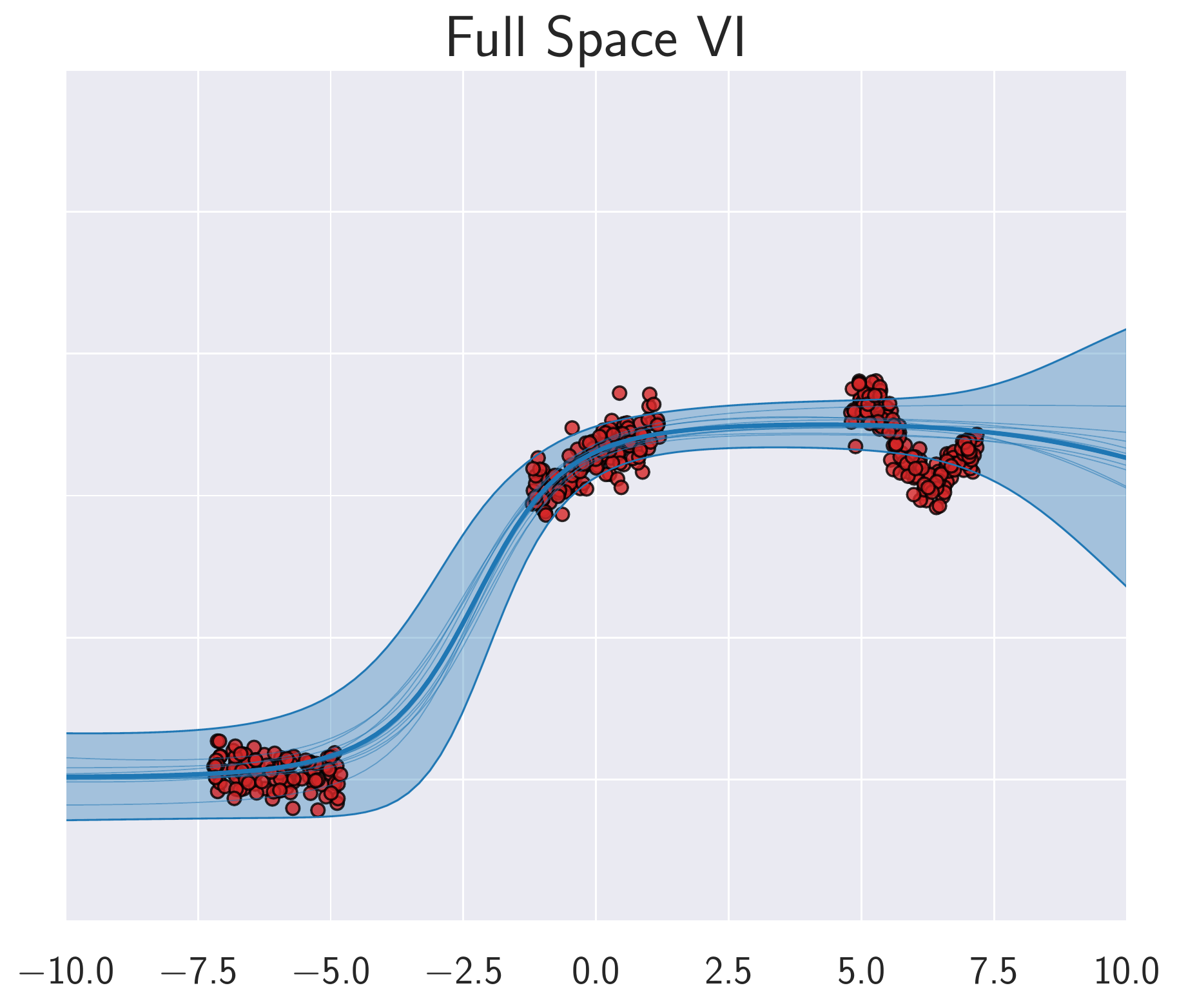}
	\end{subfigure}
	\quad\quad
	\begin{subfigure}{0.28\textwidth}
		\centering
		\includegraphics[width=\textwidth]{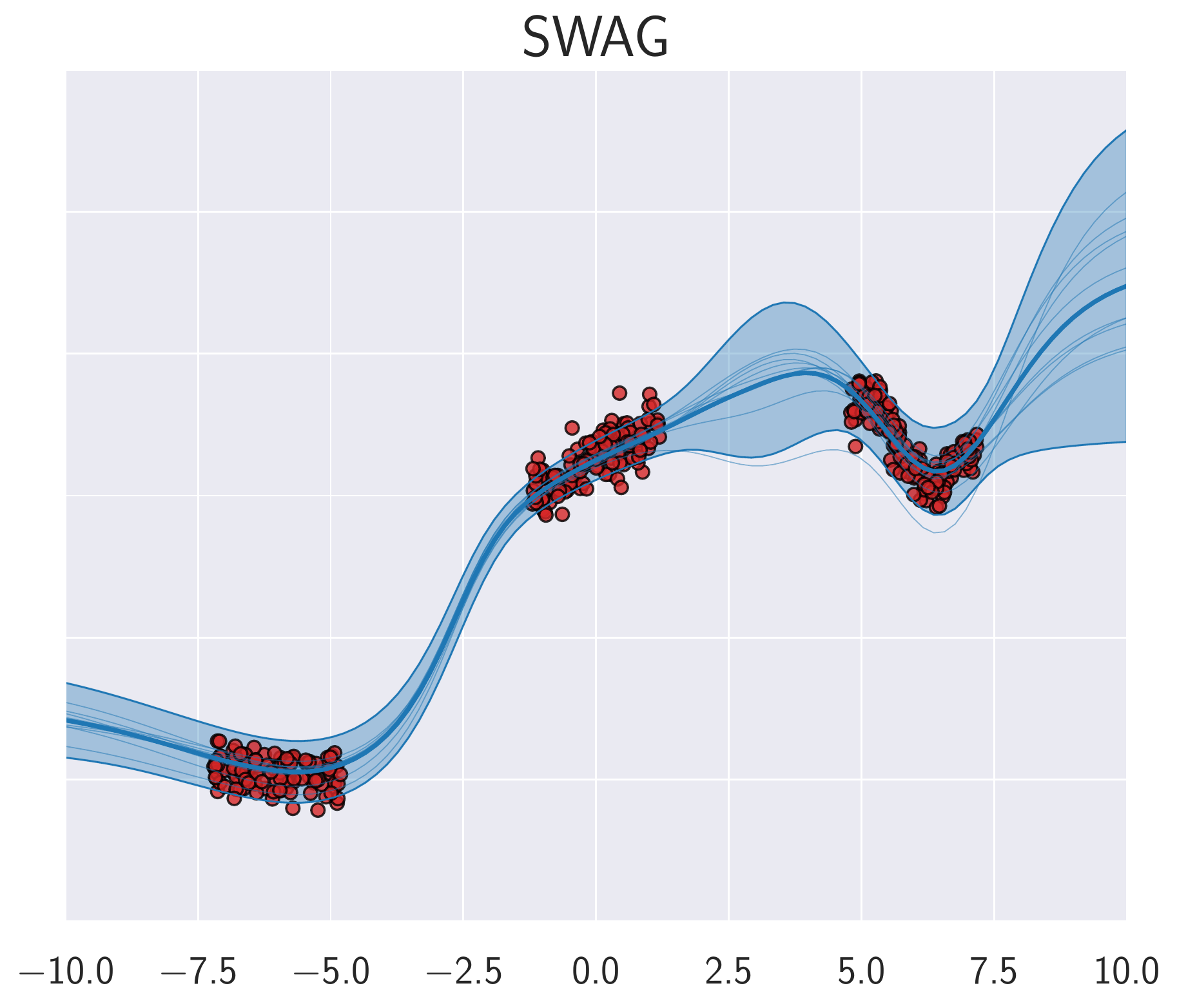}
	\end{subfigure}
	\quad\quad
	\begin{subfigure}{0.28\textwidth}
		\centering
		\includegraphics[width=\textwidth]{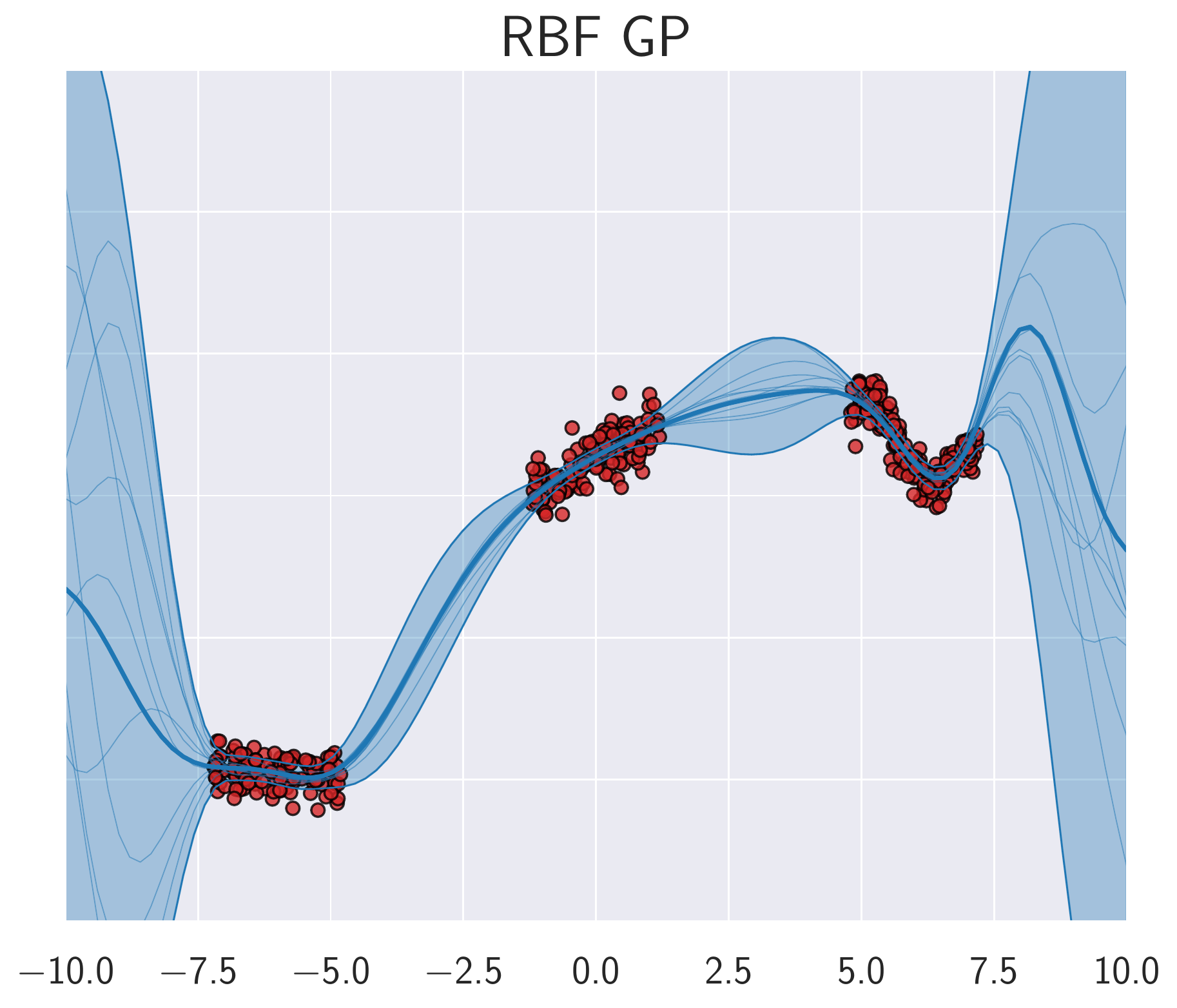}
	\end{subfigure}
	\caption{
	Predictive distribution for visualizing uncertainty in regression. Data (red circles), predictive mean (dark blue line), 
	sample posterior functions (light blue lines), $\pm 3$ standard deviations about the mean (shaded region).
	Elliptical slice sampling (ESS) with either a PCA or curve subspace provides uncertainty that intuitively grows
	away from the data, unlike variational
	inference in the full parameter space or ESS with a random subspace.
	This intuitive behaviour matches a GP with an RBF kernel, except the GP is less 
	confident for extrapolation.}
	\label{fig:toyreg_trajectories}
	\vspace{-0.3cm}
\end{figure*}

We want predicted uncertainty to grow as we move away from the data. Far away from the data there are many possible functions that are consistent with our observations and thus there should be greater uncertainty.
However, this intuitive behaviour is difficult to achieve with Bayesian neural networks \citep{inbetween}.
Further, log-likelihoods on benchmark datasets (see Section \ref{sec:ucireg}) do not necessarily test this behaviour, where 
over-confident methods can obtain better likelihoods \citep{weiwei}.

We use a fully-connected architecture with hidden layers that have
$[200, 50, 50, 50]$ neurons respectively. The network 
takes two inputs, $x$ and $x^2$ (the redundancy helps with 
training),  and outputs a single real value $y = f(x)$.
To generate the data we set the weights of the network
with this same architecture randomly, and evaluate the predictions $f(x)$ for
$400$ points sampled uniformly in intervals $[-7.2, -4.8]$, $[-1.2, 1.2]$, 
$[4.8, 7.2]$. We add Gaussian noise to the outputs $y = f(x) + \epsilon(x)$.
We show the data with red circles in Figure \ref{fig:toyreg_trajectories}.

We train an SWA solution \citep{swa}, and construct three subspaces: a ten-dimensional
random subspace, ten-dimensional PCA-subspace and a two-dimensional curve subspace
(see Section \ref{sec:subspace}). 
We then run each of the inference methods 
listed in Section \ref{sec:appendix_approxinf} in each of the subspaces. 
We visualize the predictive distributions 
in the observed space for each combination of method and subspace 
in Appendix \ref{sec:appendix_regvis} -
Figure \ref{fig:supp_toyreg_trajectories} -
and the posterior density overlayed by 
samples of the subspace parameters in Figure \ref{fig:supp_toyreg_samples}.

In Figure~\ref{fig:supp_toyreg_samples} the shape of the posterior
in random and PCA subspaces is close to Gaussian, and all approximate inference
methods produce reasonable samples. In the 
mode connecting curve subspace the posterior has a more complex shape. 
The variational methods were unable to produce a reasonable fit. The simple
variational approach is highly constrained in its Gaussian representation of the posterior.
However, RealNVP in principle has the flexibility to fit many posterior approximating 
distributions, but perhaps lacks the inductive biases to easily fit these types of 
curvy distributions in practice, especially when trained with the variational ELBO.
On the other hand, certain MCMC methods, such as elliptical slice sampling, can effectively
navigate these distributions.

In the top row of Figure \ref{fig:toyreg_trajectories} we visualize the 
predictive distributions for elliptical slice sampling in each of the subspaces.
In order to represent this uncertainty, the posterior in the subspace must
assign mass to settings of the weights that give rise to 
models making a diversity of predictions in these regions. 
In the random subspace, the predictive uncertainty does not 
significantly grow far away from the data, suggesting that it does
not capture a diversity of models. On the other hand, the PCA 
subspace captures a diverse collection of models, with uncertainty
growing away from the data. Finally, the predictive distribution for the curve
subspace is the most adaptive, suggesting that it contains the greatest
variety of models corresponding to weights with high posterior probabilities.

In the bottom row of Figure \ref{fig:toyreg_trajectories} we visualize the predictive
distributions for simple variational inference applied in the original parameter
space (as opposed to a subspace), SWAG 
\citep{maddox2019simple}, and a Gaussian process with an RBF kernel. The SWAG predictive distribution is similar
to the predictive distribution of ESS in the PCA subspace, as both methods attempt
to approximate the posterior in the subspace containing the principal components
of the SGD trajectory; however, ESS operates directly in the subspace,
and is not constrained to a Gaussian representation. Variational inference interestingly
appears to \emph{underfit} the data, failing to adapt the posterior mean to structured variations
in the observed points, and provides relatively homogenous uncertainty estimates, except 
very far from the data. 
We hypothesize that underfitting happens because the KL divergence term between approximate posterior and prior distributions overpowers the data fit term in the variational lower bound objective, as we have many more parameters than data points in this problem (see \cite{blundell2015weight} for details about VI).
In the bottom right panel of Figure~\ref{fig:toyreg_trajectories} we also consider a Gaussian
process (GP) with an RBF kernel~--- presently the gold standard for regression uncertainty. The GP provides
a reasonable predictive distribution in a neighbourhood of the data, but arguably becomes underconfident
for extrapolation, with uncertainty quickly blowing up away from the data.

\subsection{UCI REGRESSION}\label{sec:ucireg}

We next compare our subspace inference methods on UCI regression tasks to a variety of methods for approximate Bayesian inference with
neural networks.\footnote{We use the pre-processing from \url{https://github.com/hughsalimbeni/bayesian_benchmarks}.}
To measure performance we compute Gaussian test likelihood (details in Appendix \ref{sec:testlike}). 
We follow convention and parameterize our neural network models so that for an input $x$ they produce two outputs, predictive mean $\mu(x, w)$ and predictive variance $\sigma(x, w)$.
For all datasets we tune the temperature $T$ in \eqref{eq:like} by maximizing the average likelihood over $3$ random validations splits (see Appendix \ref{sec:temp_effect} for a discussion of the effect of temperature).

\begin{figure*}
    \hspace{-0.035\textwidth}
	\includegraphics[width=1.05\textwidth]{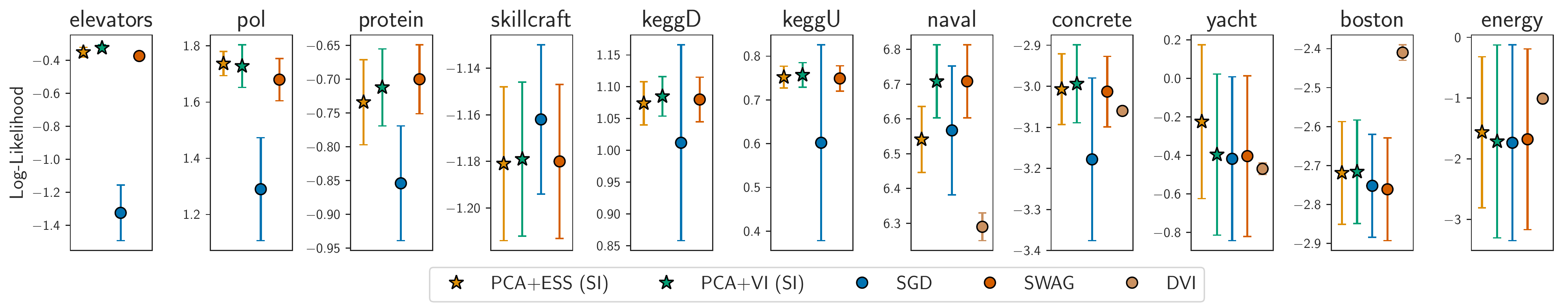}
	\caption{
	Test log-likelihoods for subspace inference and baselines on UCI regression datasets. Subspace inference (SI) with PCA achieves as good or better test log-likelihoods compared to SGD and SWAG, and is competitive with DVI (which does not immediately scale to larger problems or networks).
	We report mean over $20$ random splits of the data $\pm$ one standard deviation.
	For consistency with the literature, we report 
	normalized log-likelihoods on large datasets (elevators, pol, protein, 
	skillcraft, keggD, keggU; see Section \ref{sec:uci_large}) and unnormalized log-likelihoods on small datasets (naval, concrete, yacht, boston, energy; see Section \ref{sec:uci_small}).}
	\label{fig:testll}
	\vspace{-0.3cm}
\end{figure*}

\subsubsection{Large UCI Regression Datasets}
\label{sec:uci_large}
We experiment with $5$ large regression datasets from UCI: \emph{elevators}, \emph{keggdirected}, \emph{keggundirected}, \emph{pol}, \emph{protein} and \emph{skillcraft}.
We follow the experimental framework of \citet{wilson2016deep}.

On all datasets except skillcraft we use a feedforward network with five hidden layers of sizes [1000, 1000, 500, 50, 2], ReLU activations and two outputs $\mu(x, w)$ and $\sigma_w(x)$ parameterizing predictive mean and variance.
On  skillcraft, we use a smaller architecture [1000, 500, 50, 2] like in \citet{wilson2016deep}.
We additionally learn a global noise variance $s^2$, so that the predictive variance at $x$ is $\sigma^2(x, w, s) = s^2 + \sigma_w^2(x)$, where $ \sigma_w^2(x)$ is the variance output from the final layer of the network. 
We use softplus parameterizations for both $\sigma_w$ and $s$ to ensure positiveness of the variance, initializing the global variance at $s^2 = 1$ (the total variance in the dataset).

We compare subspace inference to deep kernel learning (DKL) \citep{wilson2016deep} with a spectral mixture base kernel \citep{wilson2013gaussian}, SGD trained networks with a Bayesian final layer as in \citet{riquelme2018deep} and two approximate Gaussian processes: orthogonally decoupled variational Gaussian processes (OrthVGP) \citep{salimbeni2018orthogonally} and Fastfood approximate kernels (FF)\footnote{Results for FF are from \citet{wilson2016deep}.} \citep{yang2015carte}.
We show the test log-likelihoods and RMSEs in Appendix \ref{sec:appendix_ucilarge} -
Tables \ref{tab:bigrmse} and \ref{tab:bigtestll}, and the 95\% credible intervals in Table \ref{tab:bigcalibration}.
We summarize the test log-likelihoods in Figure \ref{fig:testll}. 
Subspace inference outperforms SGD by a large margin on elevators, pol and protein, and is competitive on the other datasets. 
Compared to SWAG, subspace inference typically improves results by a small margin.

Finally, we plot the coverage of the 95\% predictive intervals in Figure \ref{fig:biguci_calibration}. 
Again, subspace inference performs at least as well as the SGD baseline, and has substantially better calibration on elevators and protein.

\subsubsection{Small UCI Regression Datasets}
\label{sec:uci_small}
We compare subspace inference to the state-of-the-art approximate BNN inference methods including deterministic variational inference (DVI) \citep{wu2018fixing}, deep GPs (DGP) with two GP layers trained via expectation propagation \citep{bui2016deep}, and variational inference with the re-parameterization trick \citep{kingma2013auto}.
We follow the set-up of \citet{bui2016deep} and use a fully-connected network with a single hidden layer with 50 units.
We present the test log-likelihoods, RMSEs and test calibration results in Appendix \ref{sec:appendix_ucismall} - Tables \ref{tab:small_ll}, \ref{tab:small_rmse} and \ref{tab:small_ca}. 
We additionally visualize test log-likelihoods in Figure \ref{fig:testll} and calibrations in Appendix Figure \ref{fig:biguci_calibration}.

Our first observation is that SGD provides a surprisingly strong baseline, sometimes outperforming DVI. 
The subspace methods outperform SGD and DVI on naval, concrete and yacht and are competitive on the other datasets.

\subsection{IMAGE CLASSIFICATION}
\label{sec:cifar}

\begin{figure*}
	\centering
	\begin{subfigure}{0.29\textwidth}
		\centering
		\includegraphics[width=\textwidth]{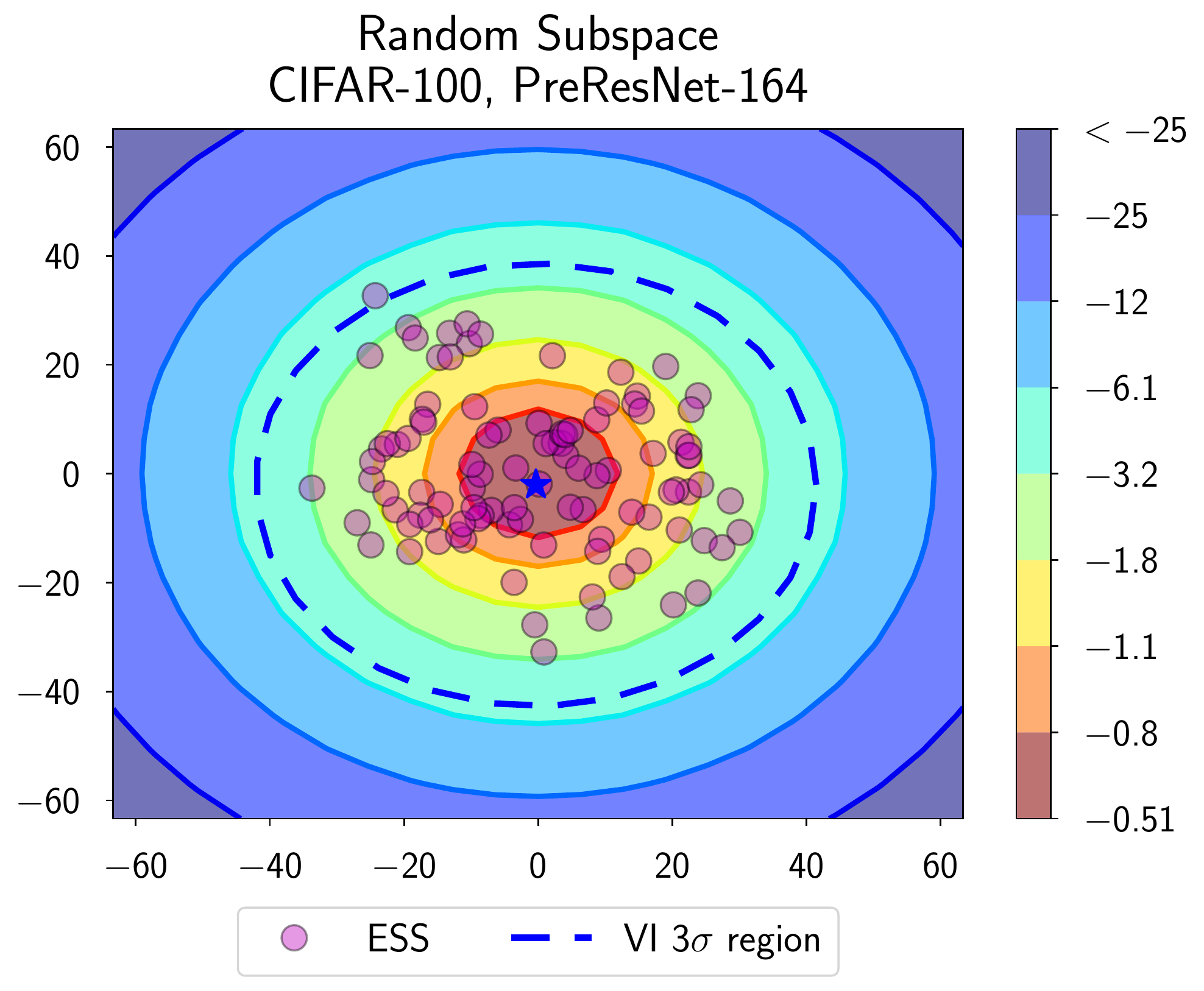}
        \caption{}
	\end{subfigure}
	\quad
	\begin{subfigure}{0.29\textwidth}
		\centering
		\includegraphics[width=\textwidth]{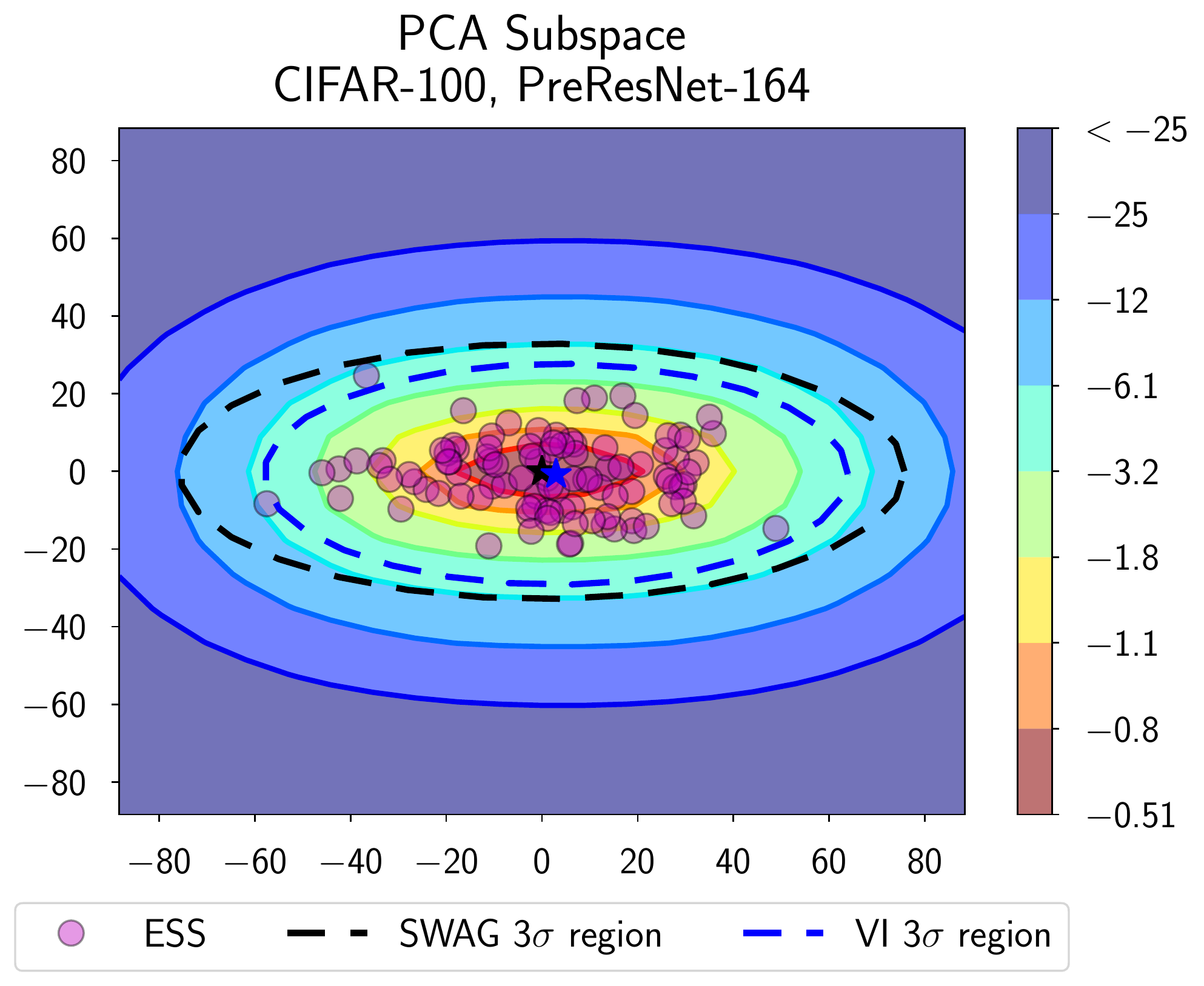}
        \caption{}
	\end{subfigure}
	\quad
	\begin{subfigure}{0.29\textwidth}
		\centering
		\includegraphics[width=\textwidth]{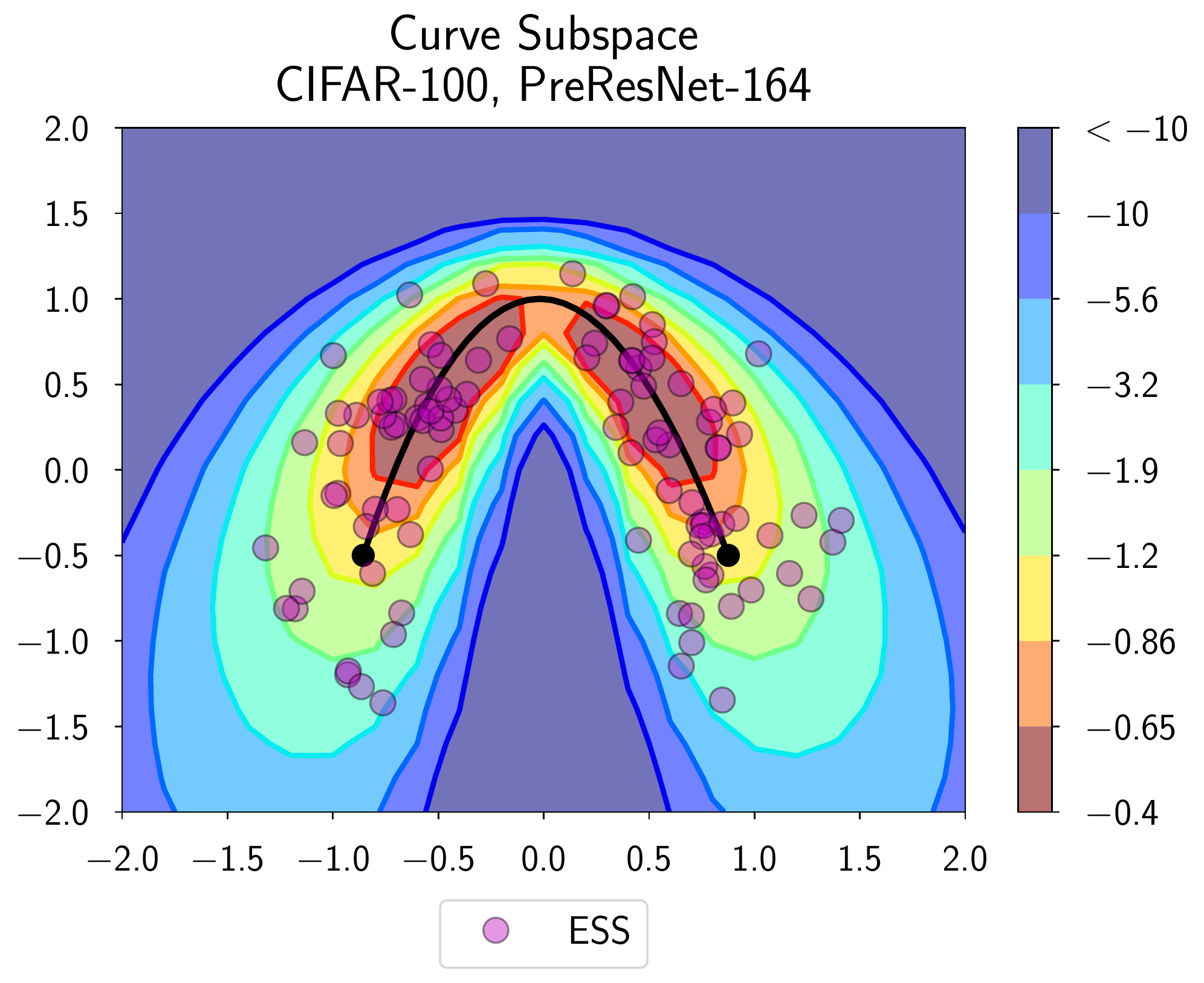}
        \caption{}
	\end{subfigure}
	\caption{Posterior log-density surfaces, ESS samples (shown with magenta circles),
    and VI approximation posterior distribution ($3\sigma$-region shown with blue 
    dashed line)
    in \textbf{(a)} random, 
    \textbf{(b)} PCA and
    \textbf{(c)} curve subspaces for PreResNet-164 on CIFAR-100. 
    In panel (b) the dashed black line shows the $3\sigma$-region of the SWAG
    predictive distribution.
    } 
	\label{fig:preresnet_surfaces}
	\vspace{-0.3cm}
\end{figure*}

Next, we test the proposed method on state-of-the-art convolutional networks on CIFAR
datasets. Similarly to Section \ref{sec:toyreg}, we construct five-dimensional random and five-dimensional PCA subspaces around
a trained SWA solution. We also construct a two-dimensional curve subspace by connecting our SWA
solution to another independently trained SWA solution. 
We use the value $T = 5000$ for temperature in all the image classification experiments with random and PCA subspaces, and $T=10000$ in the curve subspace 
(Section \ref{sec:temp}), from cross-validation using VGG-16 on CIFAR-100.

We visualize the samples
from ESS in each of the subspaces in Figure \ref{fig:preresnet_surfaces}. For VI 
we also visualize the
$3\sigma$-region of the closed-form approximate posterior in the random and PCA subspaces.
As we can see, ESS is able to capture the shape of the posterior in each
of the subspaces.

In the PCA subspace we also visualize the SWAG approximate posterior distribution.
SWAG overestimates the variance along the first principle component (horizontal axis in Figure \ref{fig:preresnet_surfaces}b) of the SGD trajectory\footnote{See also Figure 5 in \citet{maddox2019simple}.}. 
VI in the subspace is able
to provide a better fit for the posterior distribution, as it is directly approximating the posterior while SWAG is approximating the SGD trajectory.

We report the accuracy and negative log-likelihood for each
of the subspaces in Table \ref{tab:cifar_subspaces}. Going from
random, to PCA, to curve subspaces provides progressively better results,
due to increasing diversity and quality of models within each subspace.
In the remaining experiments we use the PCA subspace, as it generally
provides good performance at a much lower computational cost than the curve 
subspace. 

\begin{table}[!h]
	
	\centering
	\caption{Negative log-likelihood and Accuracy for PreResNet-164 for 
    $10$-dimensional random, $10$-dimensional PCA, and $2$-dimensional
    curve subspaces.
    We report mean and stdev over $3$ independent runs.}
	\label{tab:cifar_subspaces}
	\resizebox{0.475\textwidth}{!}{%
	\begin{tabular}{lccccc}
		\toprule
		 & Random & PCA & Curve \\
		\midrule
        NLL & $0.6858 \pm 0.0052$  & $0.6652 \pm 0.004$ &  $0.6493 \pm 0.01$ \\
        Accuracy (\%) & $80.17 \pm 0.03$ & $80.54 \pm 0.13$ & $81.55 \pm 0.26$ \\ 
        \bottomrule
	\end{tabular}
	}
	\vspace{-0.1in}
\end{table}

We next apply ESS and simple VI in the PCA subspace on VGG-16, PreResNet-164
and WideResNet28x10 on CIFAR-10 and CIFAR-100. We report the results in Appendix Tables
\ref{tab:cifar_nll}, \ref{tab:cifar_acc}. 
Subspace inference is competitive with SWAG and consistently outperforms
most of the other baselines, including MC-dropout \citep{gal2016dropout}, temperature 
scaling \citep{guo2017calibration} and KFAC-Laplace \citet{ritter2018scalable}.

\section{CONCLUSION}

Bayesian methods were once a gold standard for inference with neural networks. Indeed, a Bayesian model average provides an entirely different mechanism for making predictions than standard training, with benefits in accuracy and uncertainty representation. However, efficient and practically useful Bayesian inference has remained a critical challenge for the exceptionally high dimensional parameter spaces in modern deep neural networks. In this paper, we have developed a subspace inference approach to side-step the dimensionality challenges in Bayesian deep learning: by constructing subspaces where the neural network has sufficient variability, we can easily apply standard approximate Bayesian inference methods, with strong practical results. 

In particular, we have demonstrated that simple affine subspaces constructed from the principal components of the SGD trajectory contain enough variability for practically effective Bayesian model averaging, often out-performing full parameter space inference techniques. Such subspaces can be surprisingly low dimensional (e.g., 5 dimensions), and combined with approximate inference methods, such as slice sampling, which are good at automatically exploring posterior densities, but would be entirely intractable in the original parameter space. We further improve performance with subspaces constructed from low-loss curves connecting independently trained solutions \citep{fge}, with additional computation. Subspace inference is particularly effective at representing growing and shrinking uncertainty as we move away and towards data, which has been a particular challenge for Bayesian deep learning methods. 

Crucially, our approach is modular and easily adaptable for exploring both different subspaces and approximate inference techniques, enabling fast exploration for new problems and architectures. There are many exciting directions for future work. It could be possible to explicitly train subspaces to select high variability in the functional outputs of the model, while retaining quality solutions. One could also develop Bayesian PCA approaches \citep{minka2001automatic} to automatically select the dimensionality of the subspace. Moreover, the intuitive behaviour of the regression uncertainty for subspace inference could be harnessed in a number of active learning problems, such as Bayesian optimization and probabilistic model-based reinforcement learning, helping to address the current issues of input dimensionality in these settings.

The ability to perform approximate Bayesian inference in low-dimensional subspaces of deep neural networks is a step towards, scalable, modular, and interpretable Bayesian deep learning.

\subsubsection*{Acknowledgements}
WJM, PI, PK and AGW were supported by an Amazon Research Award, Facebook Research, and NSF IIS-1563887.
WJM was additionally supported by an NSF Graduate Research Fellowship under Grant No. DGE-1650441.

\bibliographystyle{apalike}
\bibliography{refs}

\clearpage
\appendix
 
\section{ADDITIONAL DISCUSSION OF SUBSPACE INFERENCE}

\subsection{LOSS OF MEASURE}
\label{sec:loss_measure}

When mapping into a lower-dimensional subspace of the true parameter space, we lose the ability to invert the transform and thus measure (i.e. volume of the distribution) is lost. 
Consider the following simple example in $\mathbb{R}^2$.
First, form a spherical density, $p(x,y) = \mathcal{N}(0, I_2)$, and then fix $x$ and $y$ along a slice, such that $x - y = c$.
The support of the resulting distribution has \textit{no area}, since it represents a line with no width.
For this reason, it is more correct to consider the subspace model \eqref{eq:like} as a different model that shares many of the same functional properties as the fully parametrized model, rather than a re-parametrized version of the same model. Indeed, we cannot construct a Jacobian matrix to represent the density in the subspace. 

\subsection{POTENTIAL BENEFITS OF SUBSPACE INFERENCE}
\label{sec:appendix_benefits}

\paragraph{Quicker Exploration of the Posterior} Reducing the dimensionality of parameter space enables significantly faster mixing of MCMC chains. For example, the expected number of likelihood evaluations 
needed for an accepted sample drawn using Metropolis-Hastings or Hamiltonian Monte Carlo (HMC) respectively grow as $d^2$ and $d^{5/4}$. 
If the dimensionality of the subspace grows as $K=\log{d}$, for example, then we would 
expect the runtime of Metropolis-Hastings to produce independent samples to grow at a modest $2 \log{d}$. 
We also note that the structure of the subspace may be much more amenable to exploration than the original posterior,
requiring less time to effectively cover. For example, \citet{maddox2019simple} show that the loss in the subspace constructed from the principal components of SGD iterates is approximately locally quadratic. On the other hand, if the subspace has a complicated structure, it can now be traversed using ambitious exploration methods which do not scale to higher dimensional spaces, such as parallel tempering \citep{geyer1991markov}.

\paragraph{Potential Lack of Degeneracies} 
In the subspace, the model will be absent of many of degeneracies present for most DNNs. We would expect therefore the posterior to concentrate to a single point, and be more amenable in general to approximate inference strategies.

\section{APPROXIMATE INFERENCE METHODS}\label{sec:appendix_approxinf}

We can use MCMC methods to approximately sample from 
$p(z | \mathcal{D})$, or we can perform a deterministic approximation 
$q(z | \mathcal{D}) \approx p(z | \mathcal{D})$, for example using Laplace
or a variational approach,
and then sample from $q$. 
We particularly consider the following methods in our experiments, 
although there are many other possibilities. 
The inference procedure is an experimental design choice.

\paragraph{Slice Sampling}
As the dimensionality of the subspace $K$ is low, gradient-free methods
such as slice sampling \citep{neal2003slice} and elliptical slice sampling (ESS) \citep{murray2010elliptical} can be used to 
sample from the projected posterior distribution.
Elliptical slice sampling is designed to have no tuning parameters, and only requires a Gaussian prior in the subspace.
\footnote{We use the Python implementation at \url{https://github.com/jobovy/bovy_mcmc/blob/master/bovy_mcmc/elliptical_slice.py}.}
For networks that cannot evaluate all of the training data in memory at a single time, it is easily possible to sum the loss over mini-batches computing a full log probability, without storing gradients.

\paragraph{NUTS}
The No-U-Turn Sampler (NUTS) \citep{hoffman2014no} is an HMC method \citep{neal2011mcmc} that dynamically tunes the hyper-parameters (step-size and leapfrog steps) of HMC. \footnote{Implemented in Pyro \citep{bingham2018pyro}.}
 NUTS has the advantage of being nearly black-box: only a joint likelihood and its gradients need to be defined. 
However, full gradient calls are required, which can be difficult to cache and a constant factor slower than a full likelihood calculation.

\paragraph{Simple Variational Inference}

One can perform variational inference in the subspace using the fully-factorized
Gaussian posterior approximation family for $p(z | \mathcal{D})$, from which we can 
sample to form a Bayesian model average. Fully-factorized Gaussians are among 
the simplest and the most common variational families. Unlike ESS or NUTS, VI
can be trained with mini-batches \citep{hoffman2013stochastic}, but is often 
practically constrained in the distributions it 
can represent.

\paragraph{RealNVP}

Normalizing flows, such as RealNVP \citep{realnvp}, parametrize
the variational distribution family with invertible neural networks, 
for flexible non-Gaussian posterior approximations.

\section{EIGEN-GAPS OF THE FISHER AND HESSIAN MATRICES}\label{sec:appendix_eigen}

\begin{figure}
	\centering
	\includegraphics[width=0.35\textwidth]{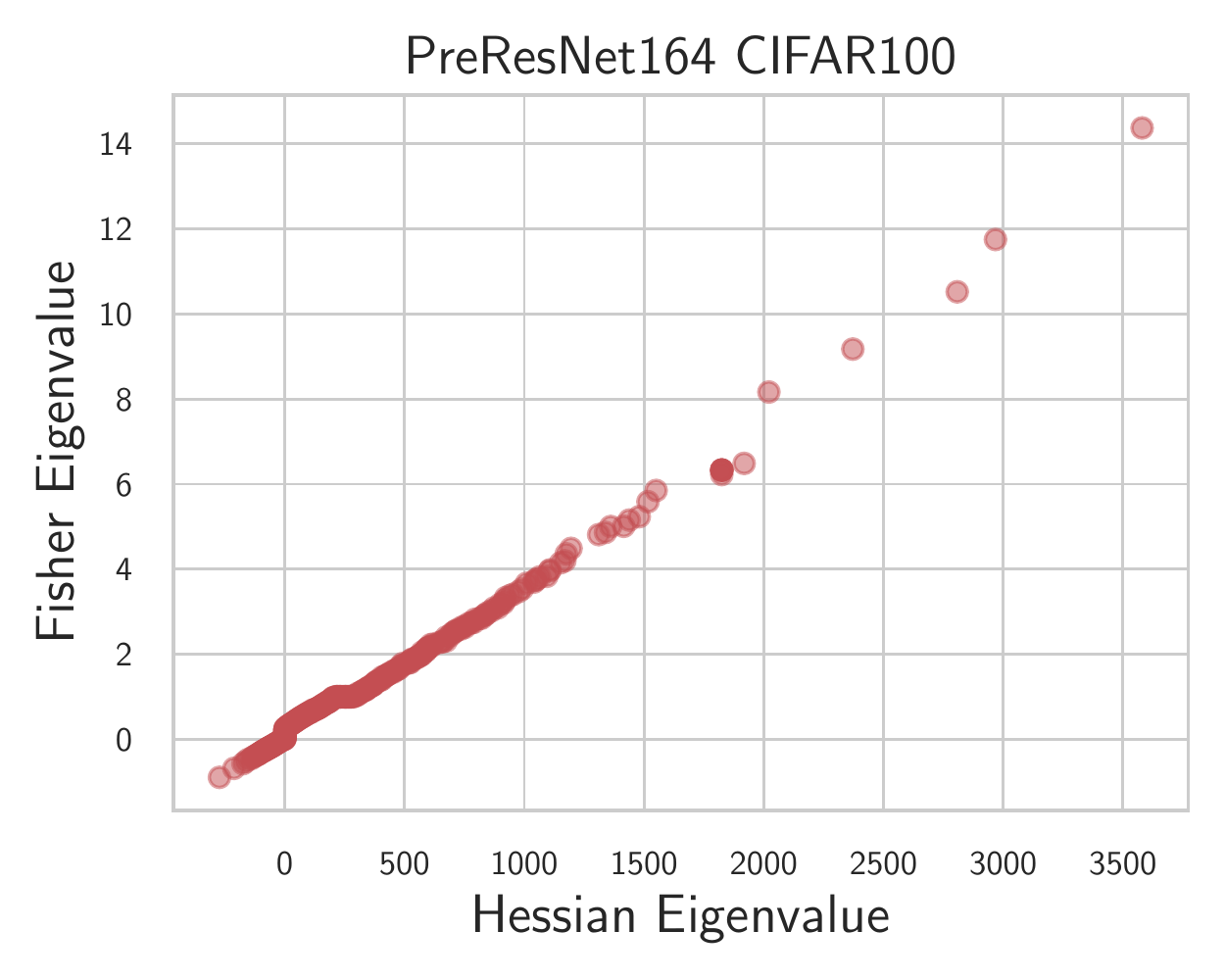}
	\caption{Plot of 300 eigenvalues of the Fisher and Hessian matrices for a PreResNet164 on CIFAR100. A clear separation exists between the top 20 or so eigenvalues and the rest, which are crowded together.}
	\label{fig:evals}
\end{figure}

\begin{figure}
 	\centering
 	\includegraphics[width=0.38\textwidth]{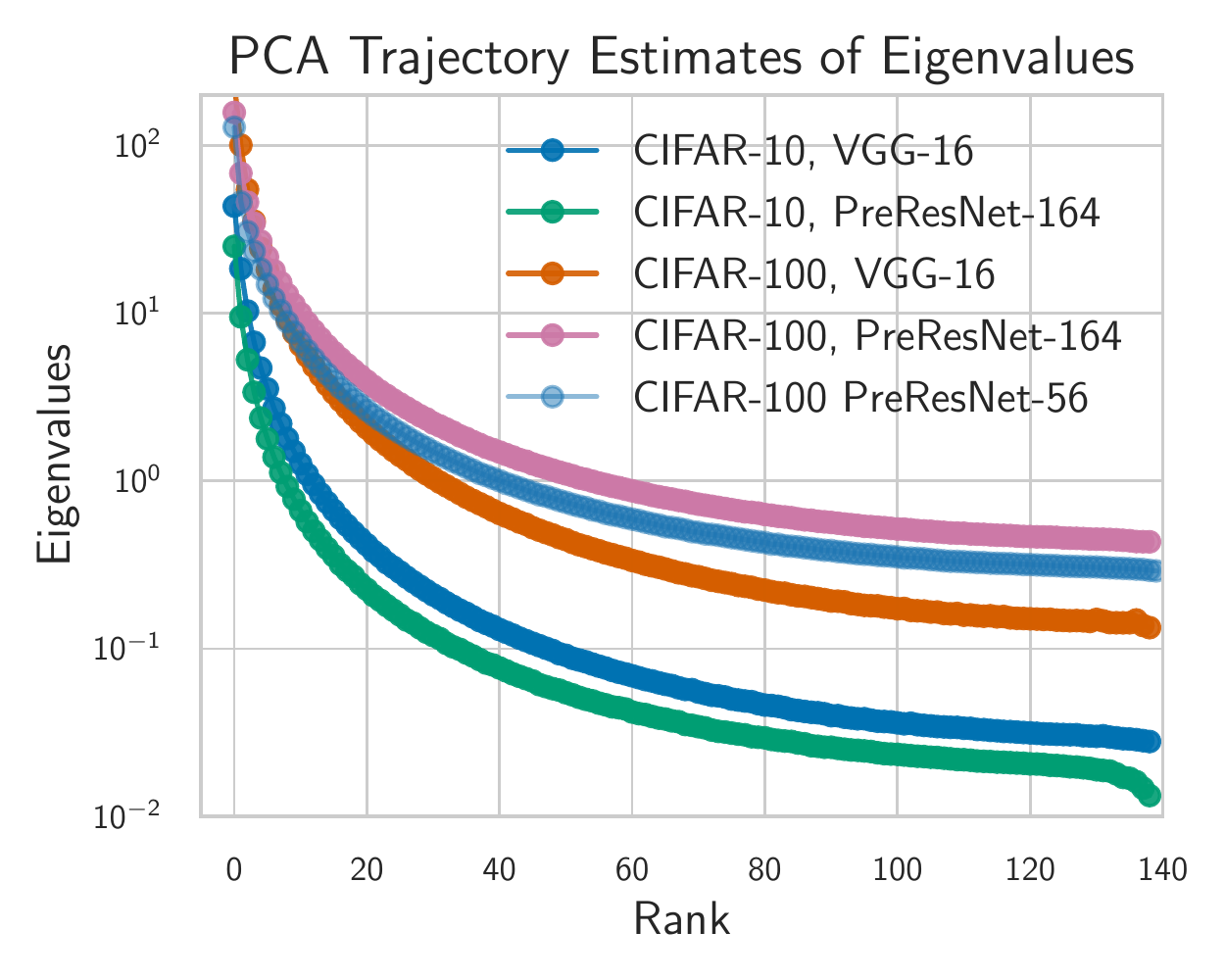}
 	\caption{Eigenvalues of trajectory covariance (explained variance proportion) estimated from randomized SVD across three architectures on CIFAR-10 and CIFAR-100 plotted on a log-scale. The trajectory decays extremely quickly, decaying towards 0 at around 10-20 steps. 
 	}
 	\label{fig:traj_est}
 \end{figure}

In Figure \ref{fig:evals} we can see similar behavior for the eigenvalues of both the Hessian and the empirical Fisher information matrix at the end of training.
To compute these eigenvalues, we used a GPU-enabled Lanczos method in GPyTorch \citep{gardner2018gpytorch} on a pre-trained PreResNet164.
Given the gap between the top eigenvalues and the rest for both the Hessian and Fisher matrices, we might expect that the parameter updates of gradient descent would primarily lie in the subspace spanned by the top eigenvalues. \citet{li_measuring_2018} and \citet{gurari2019gradient} empirically (and with a similar theoretical basis) found that the training dynamics of SGD (and thus the parameter updates) primarily lie within the subspace formed by the top eigenvalues (and eigenvectors).
Additionally, in Figure \ref{fig:traj_est} we show that the eigenvalues of the trajectory decay rapidly (on a log scale) across several different architectures; this rapid decay suggests that most of the parameter updates occur in a low dimensional subspace (at least near convergence), supporting the claims of \citet{gurari2019gradient}. 
In future work it would be interesting to further explore the connections between the eigenvalues of the Hessian (and Fisher) matrices of DNNs and the covariance matrix of the trajectory of the SGD iterates.

\section{ADDITIONAL REGRESSION UNCERTAINTY VISUALIZATIONS}
\label{sec:appendix_regvis}
\begin{figure*}
	\centering
	\begin{subfigure}{0.30\textwidth}
		\centering
		\includegraphics[width=\textwidth]{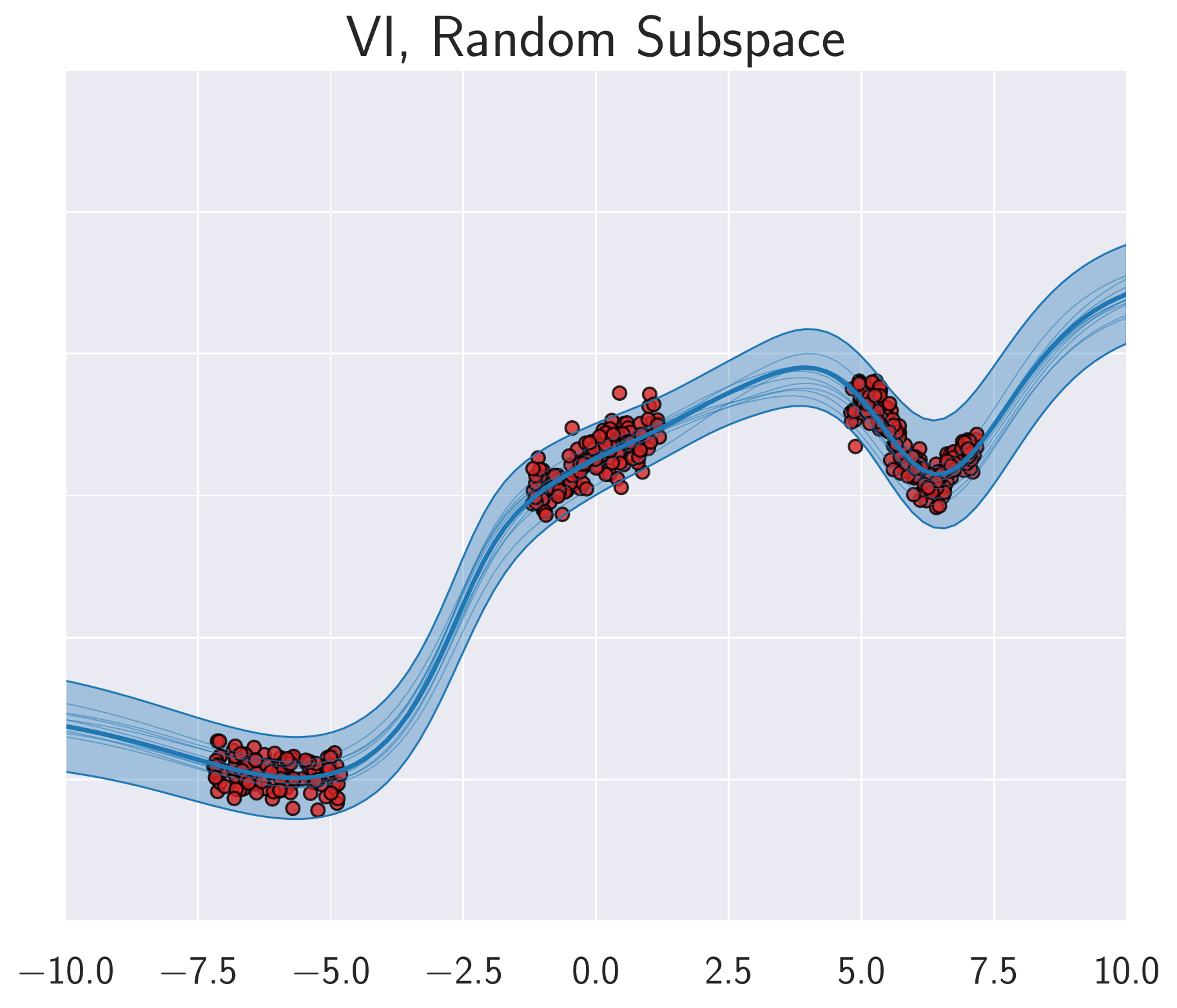}
	\end{subfigure}
	\quad
	\begin{subfigure}{0.30\textwidth}
		\centering
		\includegraphics[width=\textwidth]{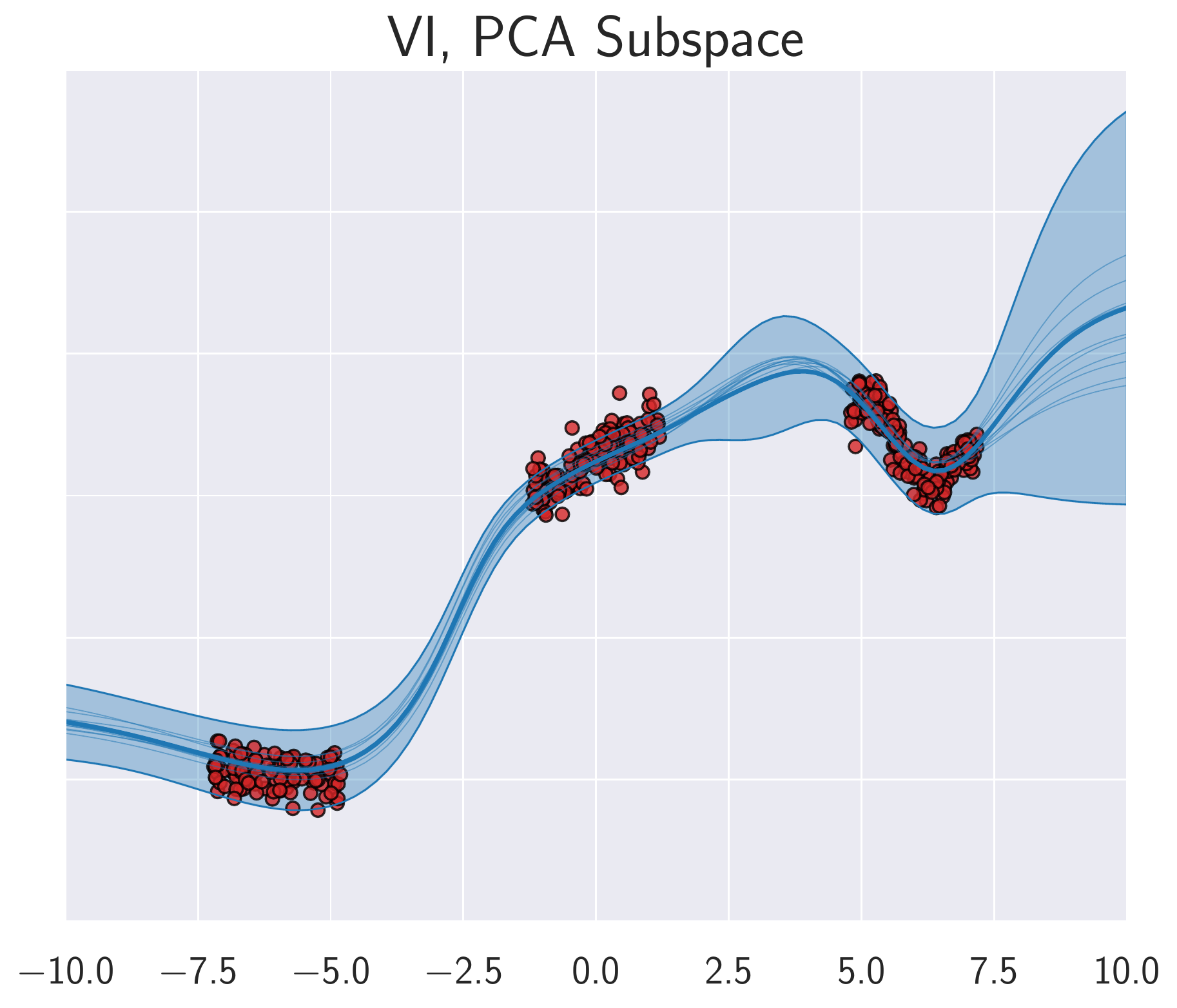}
	\end{subfigure}
	\quad
	\begin{subfigure}{0.30\textwidth}
		\centering
		\includegraphics[width=\textwidth]{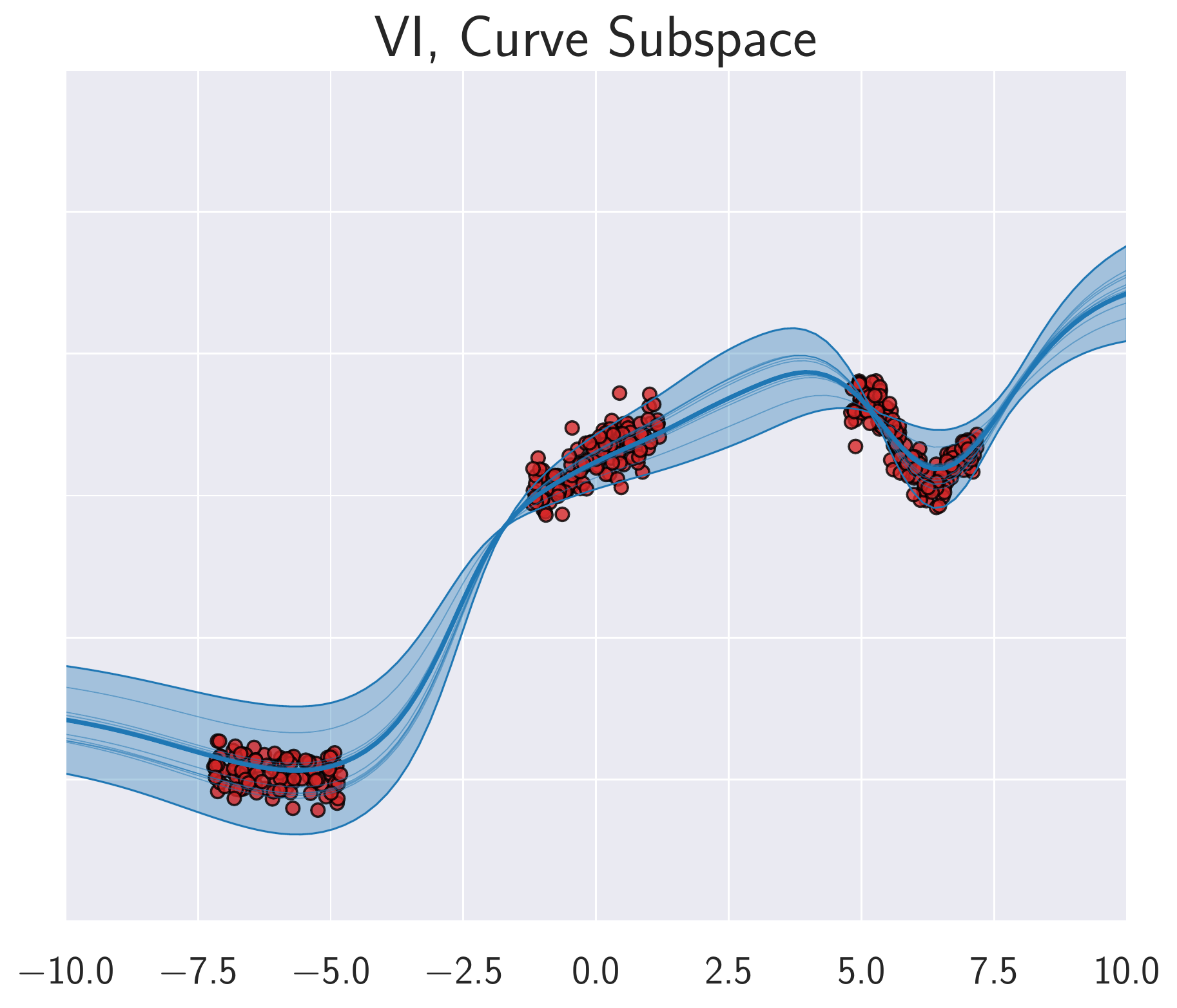}
	\end{subfigure}
	\begin{subfigure}{0.30\textwidth}
		\centering
		\includegraphics[width=\textwidth]{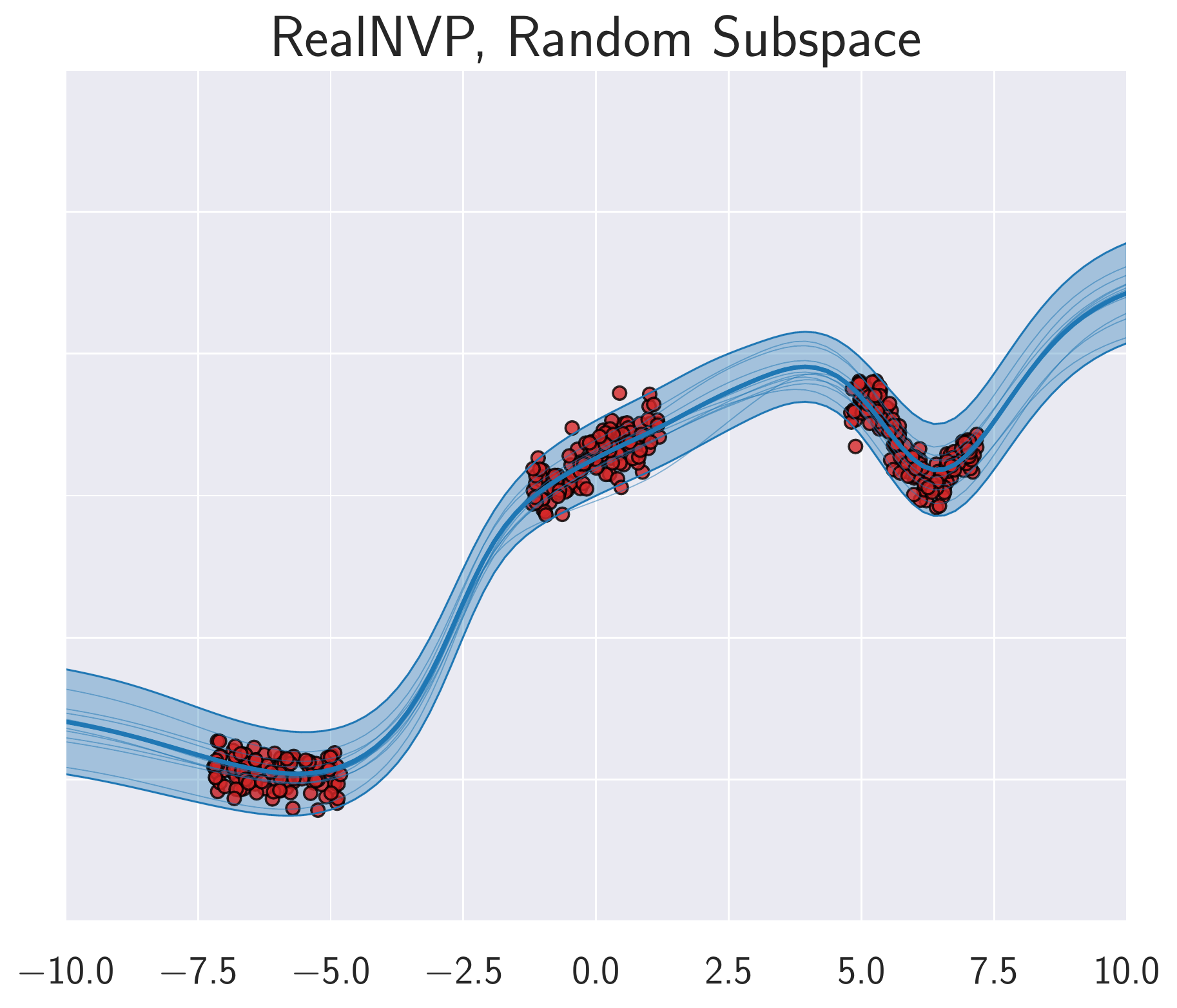}
	\end{subfigure}
	\quad
	\begin{subfigure}{0.30\textwidth}
		\centering
		\includegraphics[width=\textwidth]{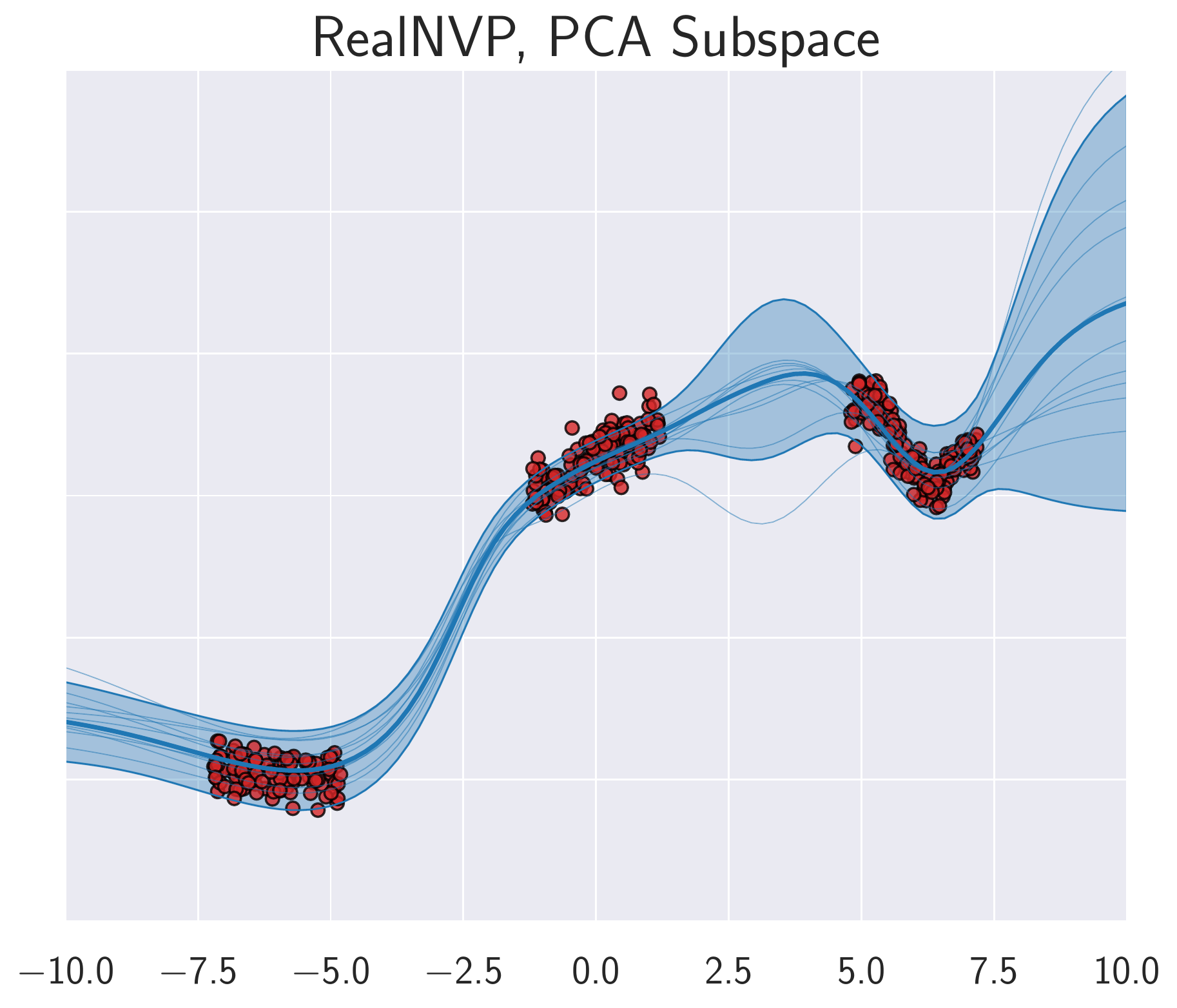}
	\end{subfigure}
	\quad
	\begin{subfigure}{0.30\textwidth}
		\centering
		\includegraphics[width=\textwidth]{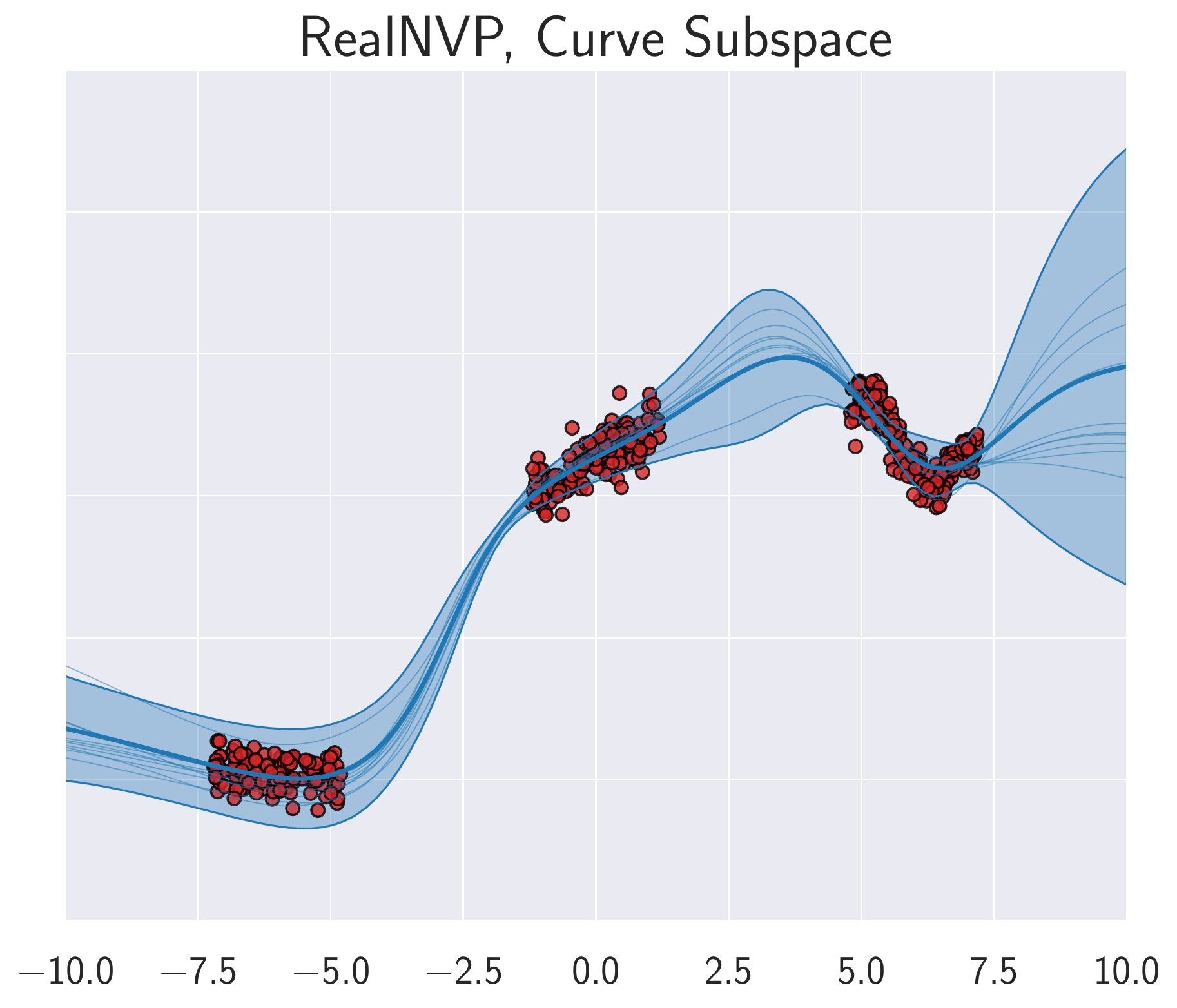}
	\end{subfigure}
	\begin{subfigure}{0.30\textwidth}
		\centering
		\includegraphics[width=\textwidth]{figs/toyreg_nutsrand.pdf}
	\end{subfigure}
	\quad
	\begin{subfigure}{0.30\textwidth}
		\centering
		\includegraphics[width=\textwidth]{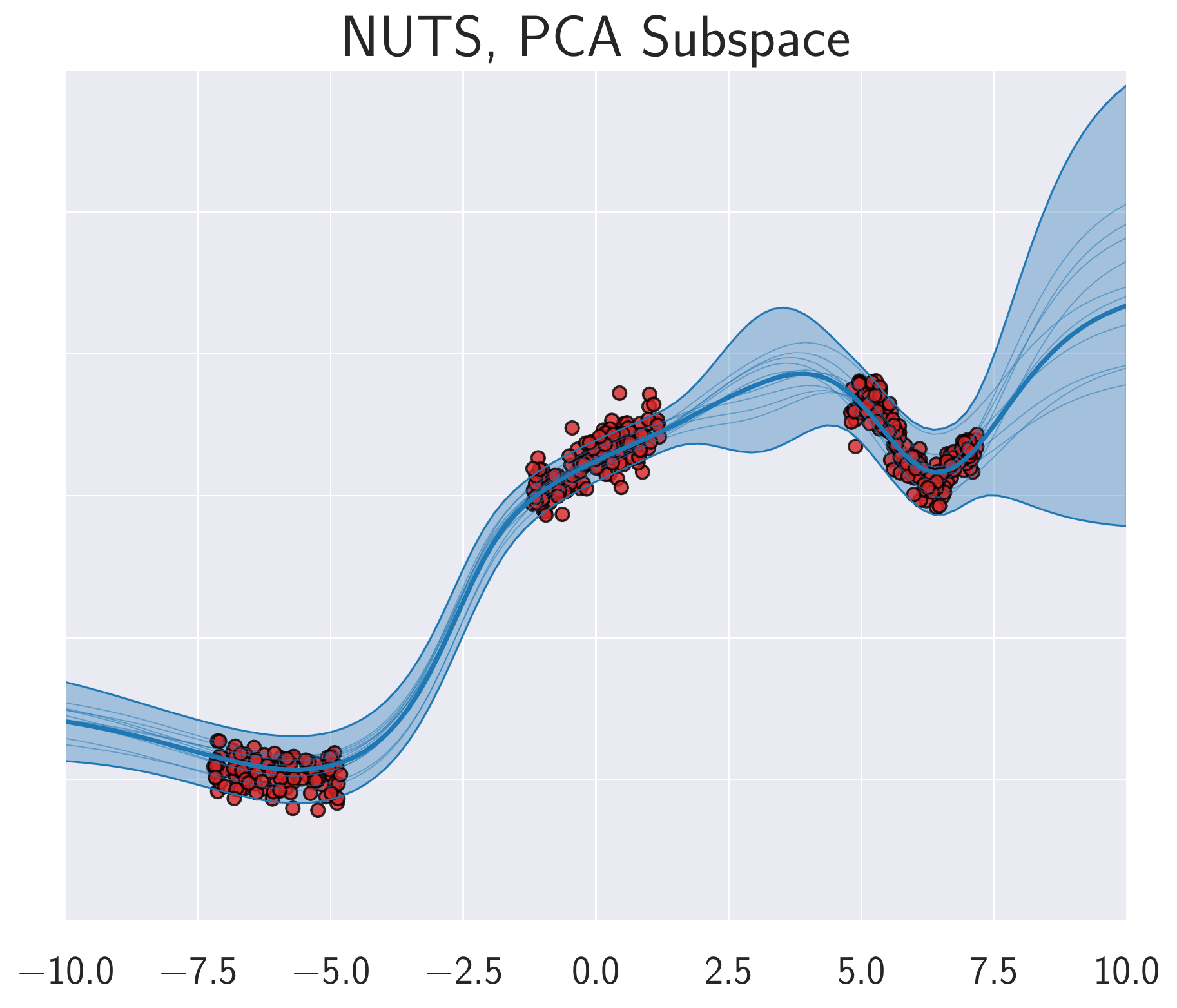}
	\end{subfigure}
	\quad
	\begin{subfigure}{0.30\textwidth}
		\centering
		\includegraphics[width=\textwidth]{figs/toyreg_nutscurve.pdf}
	\end{subfigure}
	\begin{subfigure}{0.30\textwidth}
		\centering
		\includegraphics[width=\textwidth]{figs/toyreg_essrand.pdf}
	\end{subfigure}
	\quad
	\begin{subfigure}{0.30\textwidth}
		\centering
		\includegraphics[width=\textwidth]{figs/toyreg_esspca.pdf}
	\end{subfigure}
	\quad
	\begin{subfigure}{0.30\textwidth}
		\centering
		\includegraphics[width=\textwidth]{figs/toyreg_esscurve.pdf}
	\end{subfigure}
	\begin{subfigure}{0.30\textwidth}
		\centering
		\includegraphics[width=\textwidth]{figs/toyreg_vi.pdf}
	\end{subfigure}
	\quad
	\begin{subfigure}{0.30\textwidth}
		\centering
		\includegraphics[width=\textwidth]{figs/toyreg_swag.pdf}
	\end{subfigure}
	\quad
	\begin{subfigure}{0.30\textwidth}
		\centering
		\includegraphics[width=\textwidth]{figs/toyreg_gp.pdf}
	\end{subfigure}
	\caption{Regression predictive distributions across inference methods and subspaces.
	Data is shown with red circles, the dark blue line shows predictive mean, the lighter blue lines show sample predictive functions, and the shaded region represents $\pm 3$ standard deviations of the predictive distribution at each point.} 
	\label{fig:supp_toyreg_trajectories}
\end{figure*}

\begin{figure*}
	\centering
	\begin{subfigure}{0.30\textwidth}
		\centering
		\includegraphics[width=\textwidth]{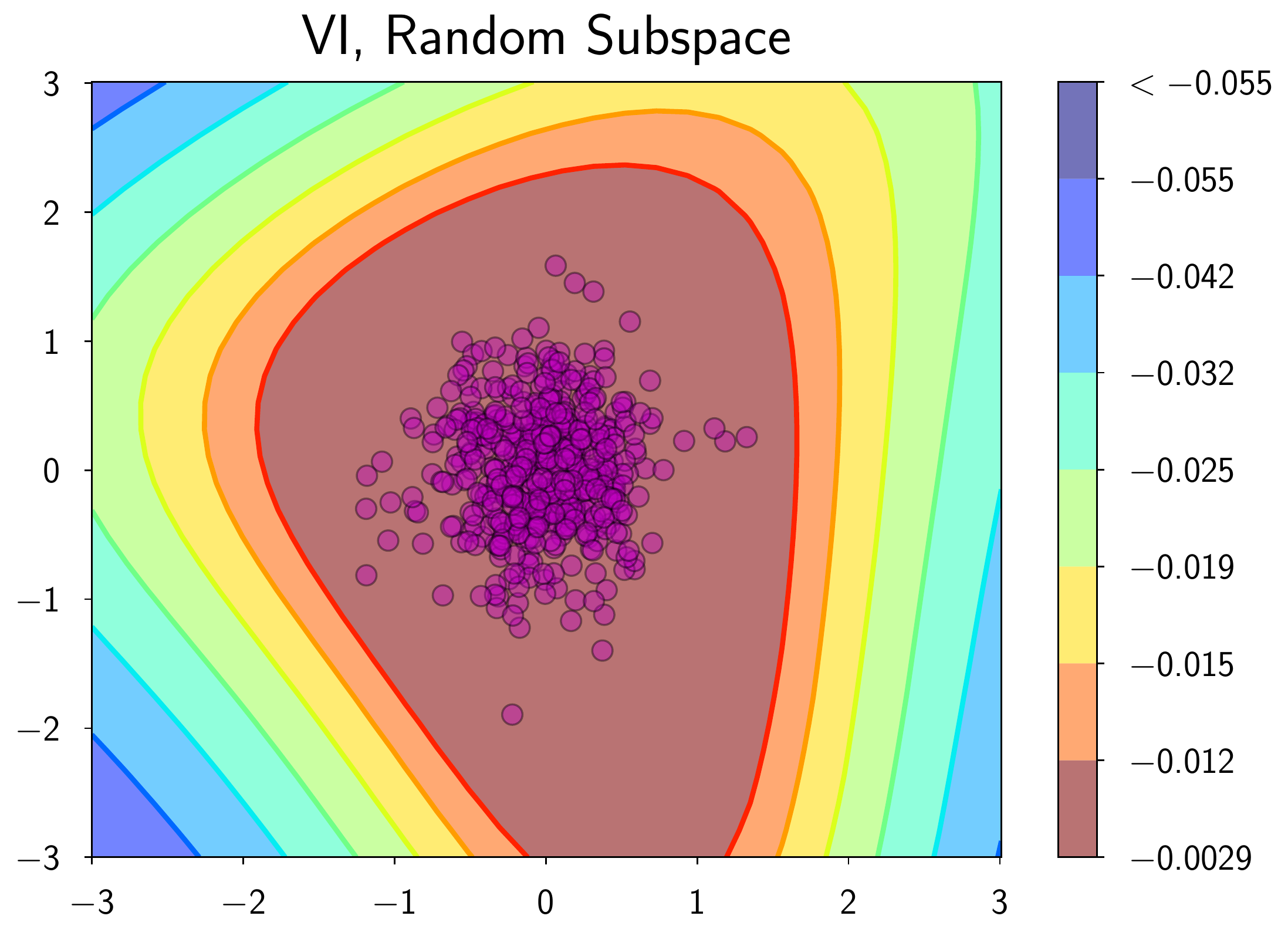}
	\end{subfigure}
	\quad
	\begin{subfigure}{0.30\textwidth}
		\centering
		\includegraphics[width=\textwidth]{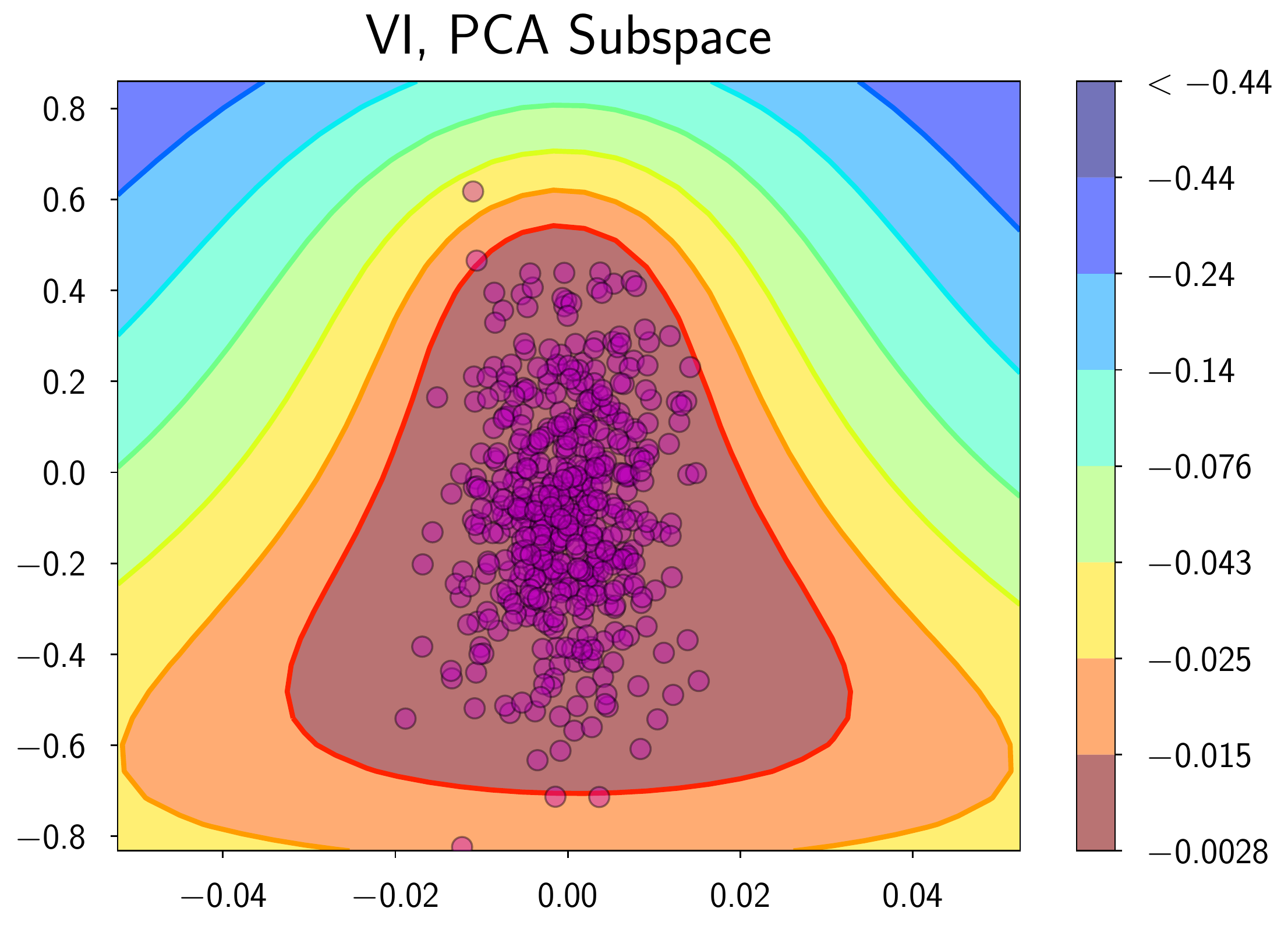}
	\end{subfigure}
	\quad
	\begin{subfigure}{0.30\textwidth}
		\centering
		\includegraphics[width=\textwidth]{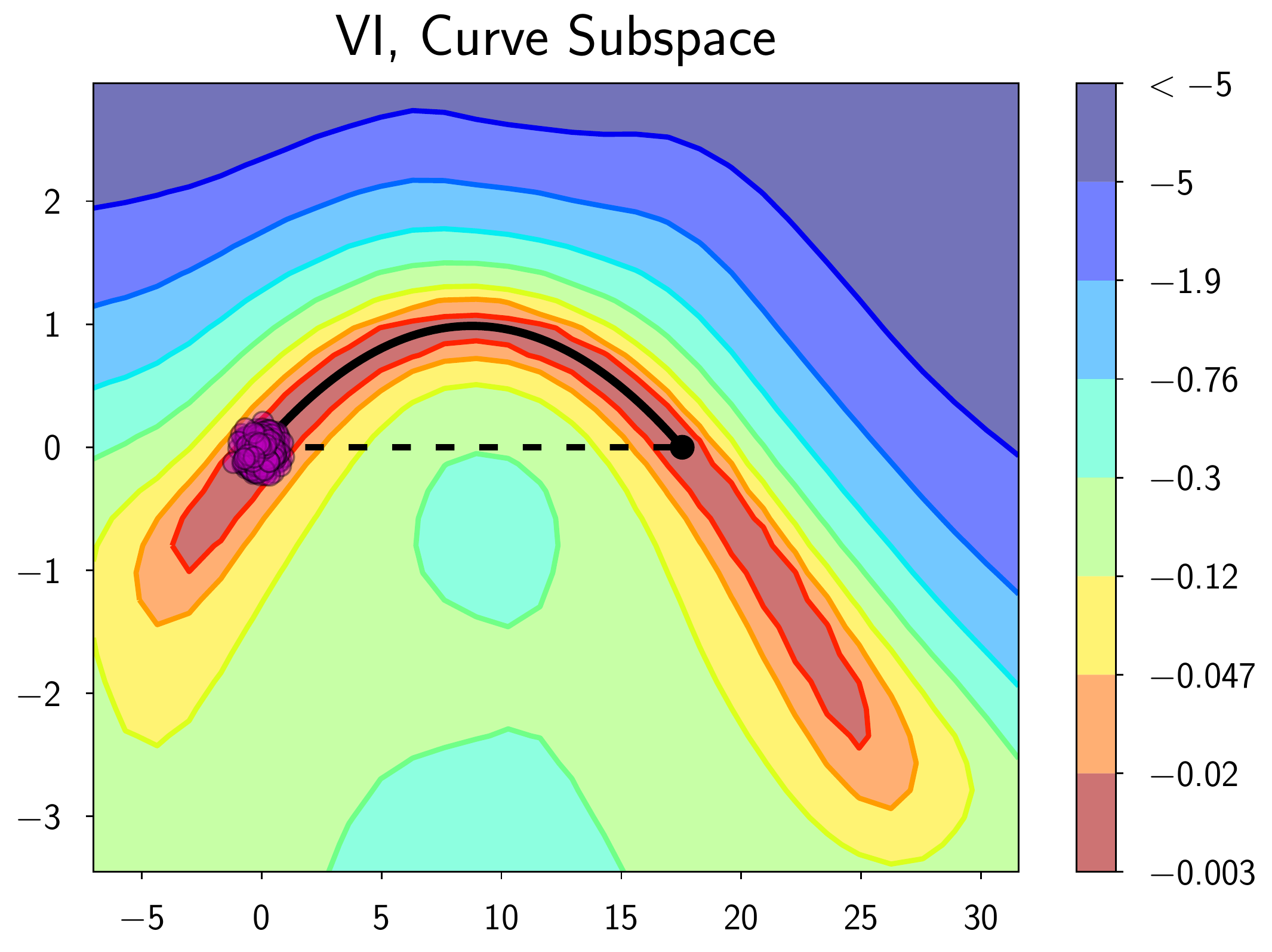}
	\end{subfigure}
	\begin{subfigure}{0.30\textwidth}
		\centering
		\includegraphics[width=\textwidth]{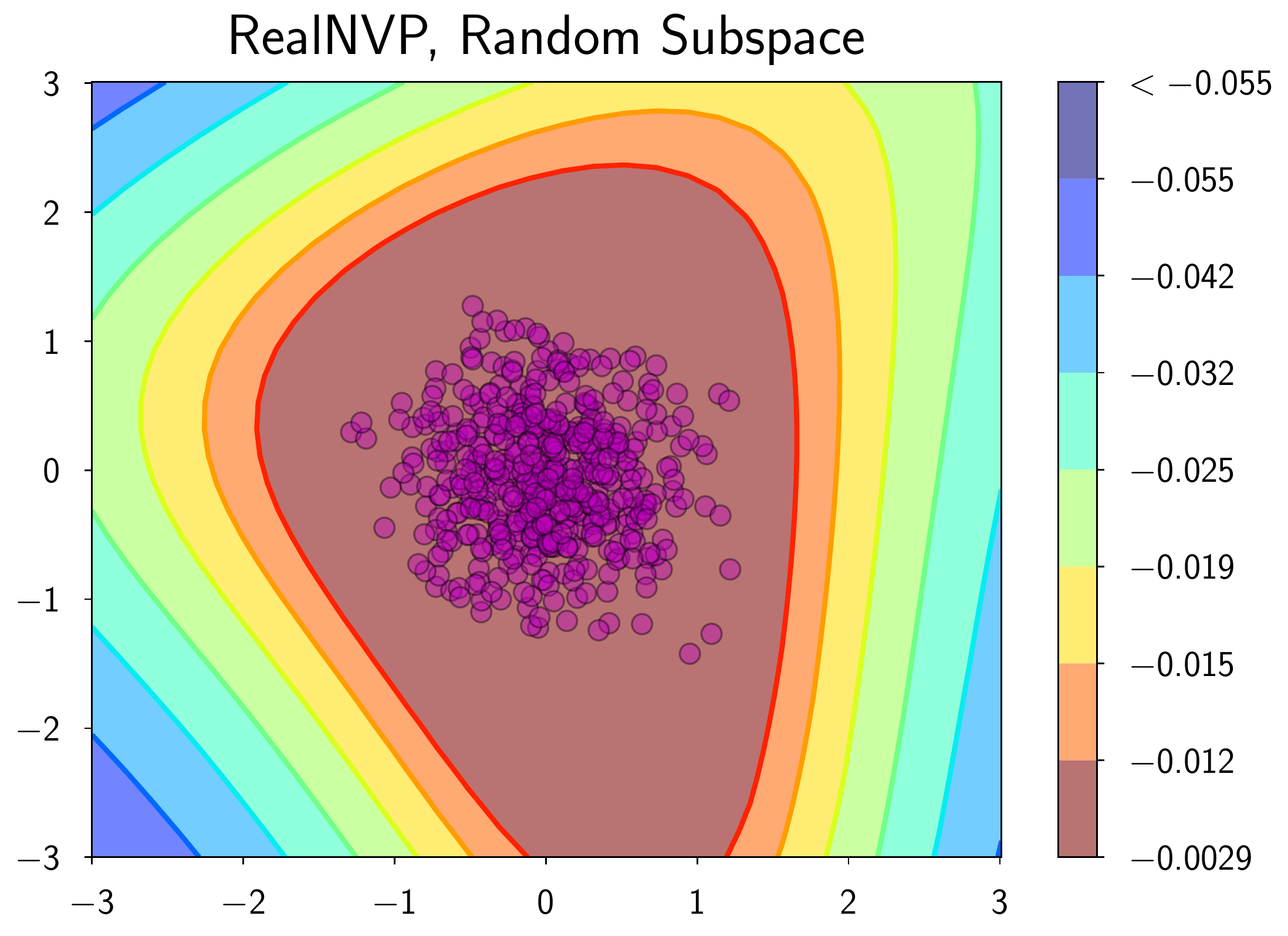}
	\end{subfigure}
	\quad
	\begin{subfigure}{0.30\textwidth}
		\centering
		\includegraphics[width=\textwidth]{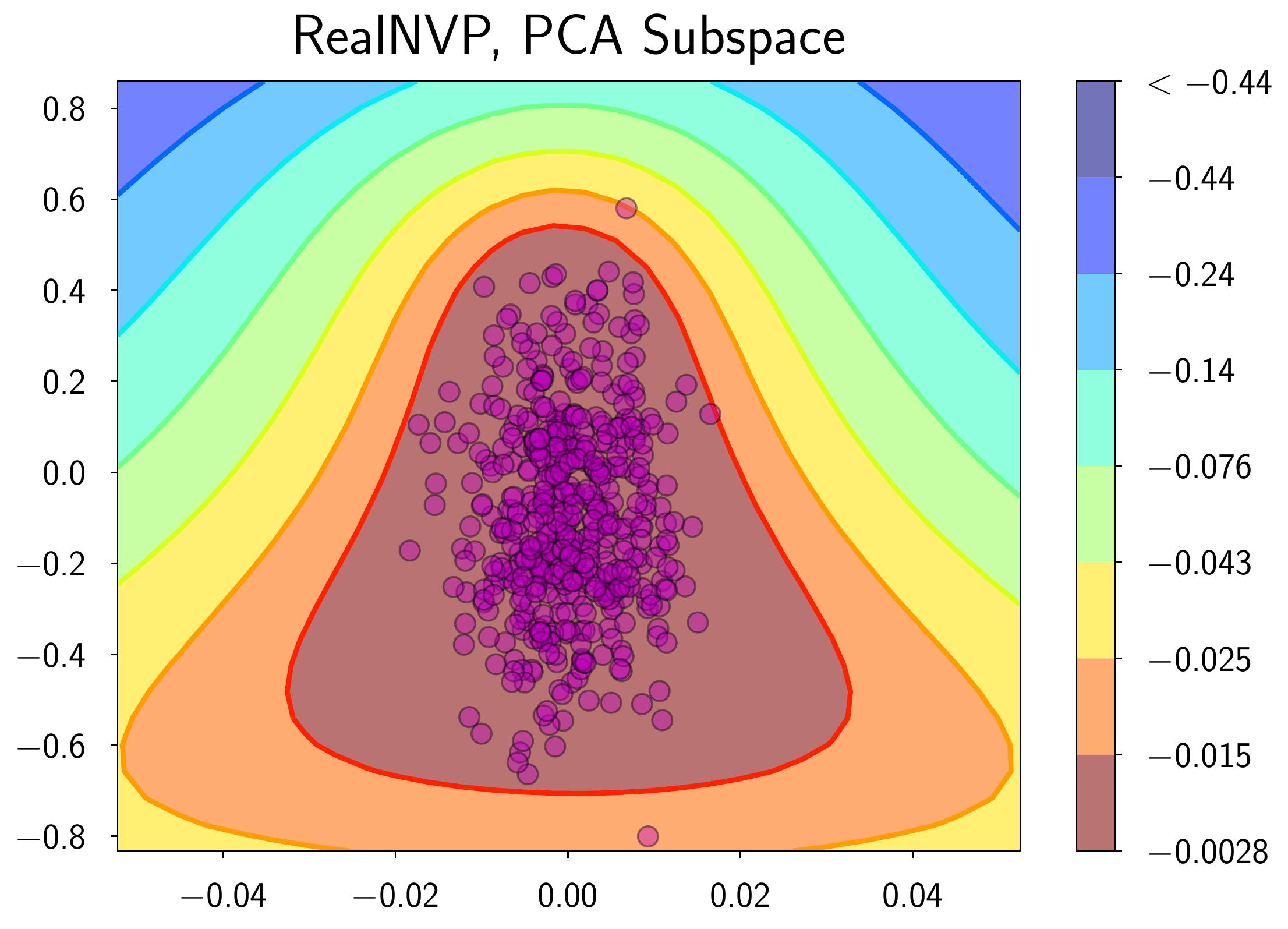}
	\end{subfigure}
	\quad
	\begin{subfigure}{0.30\textwidth}
		\centering
		\includegraphics[width=\textwidth]{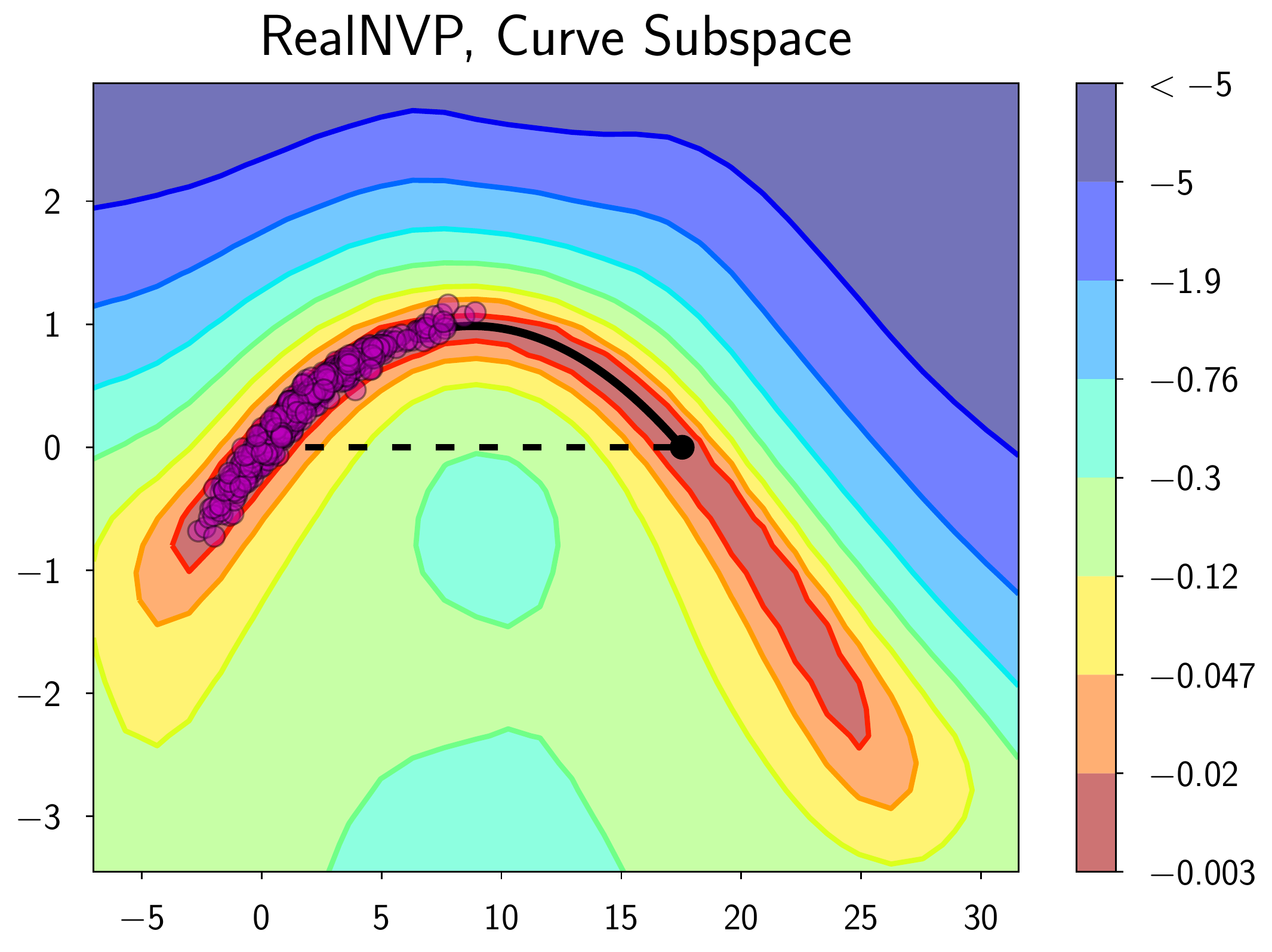}
	\end{subfigure}
	\begin{subfigure}{0.30\textwidth}
		\centering
		\includegraphics[width=\textwidth]{figs/toyreg_plane_nutsrand.pdf}
	\end{subfigure}
	\quad
	\begin{subfigure}{0.30\textwidth}
		\centering
		\includegraphics[width=\textwidth]{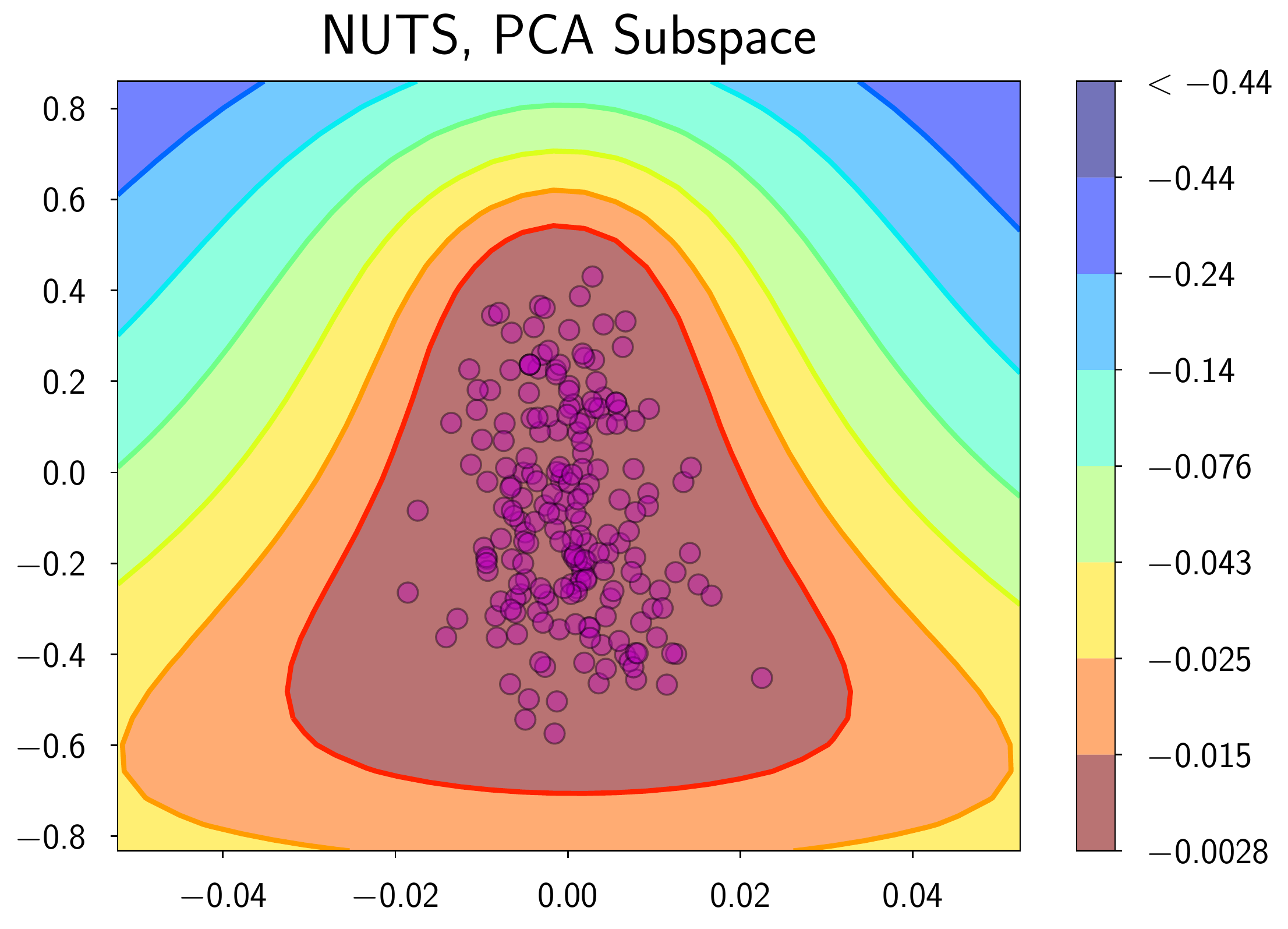}
	\end{subfigure}
	\quad
	\begin{subfigure}{0.30\textwidth}
		\centering
		\includegraphics[width=\textwidth]{figs/toyreg_plane_nutscurve.pdf}
	\end{subfigure}
	\begin{subfigure}{0.30\textwidth}
		\centering
		\includegraphics[width=\textwidth]{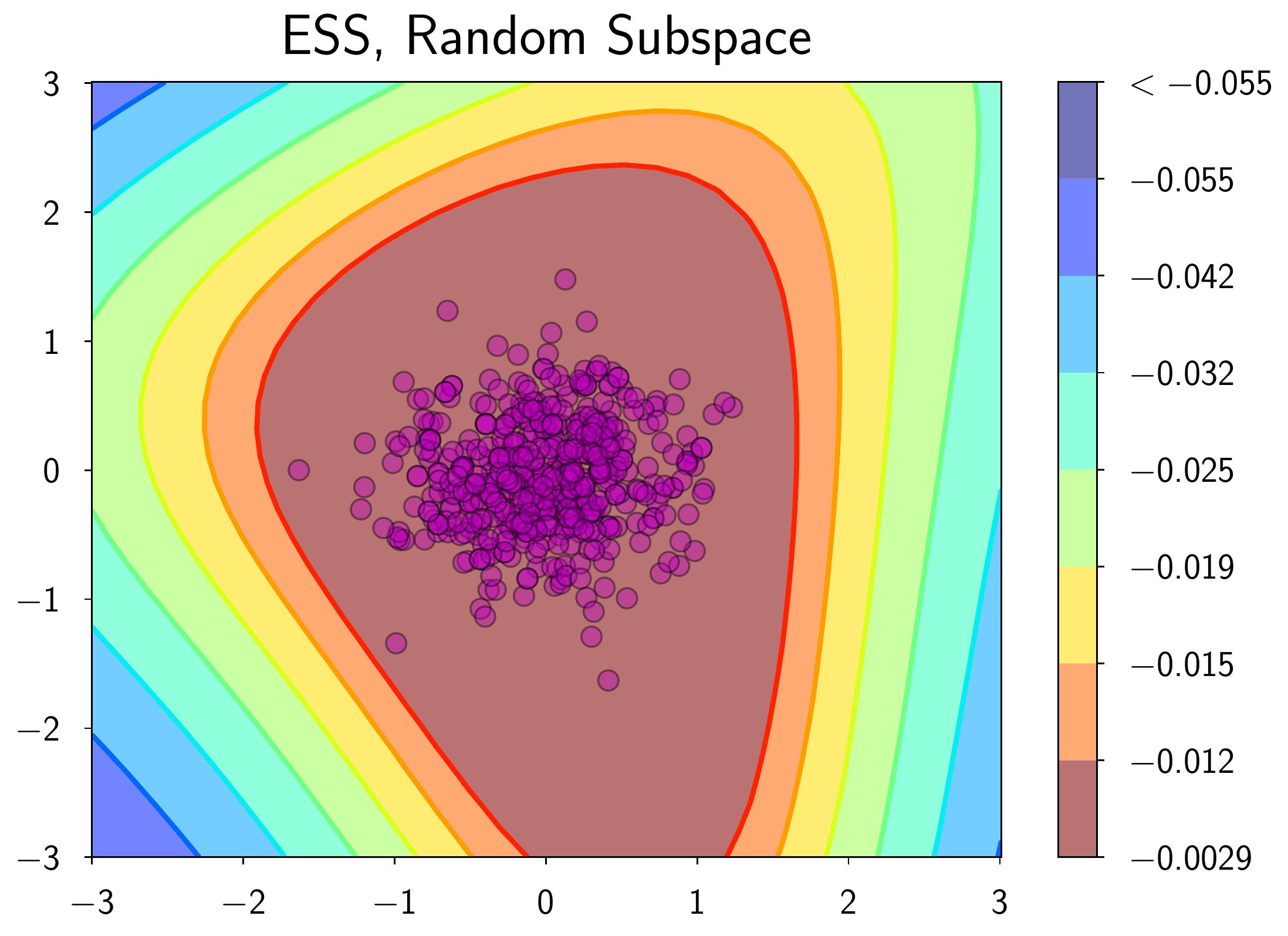}
	\end{subfigure}
	\quad
	\begin{subfigure}{0.30\textwidth}
		\centering
		\includegraphics[width=\textwidth]{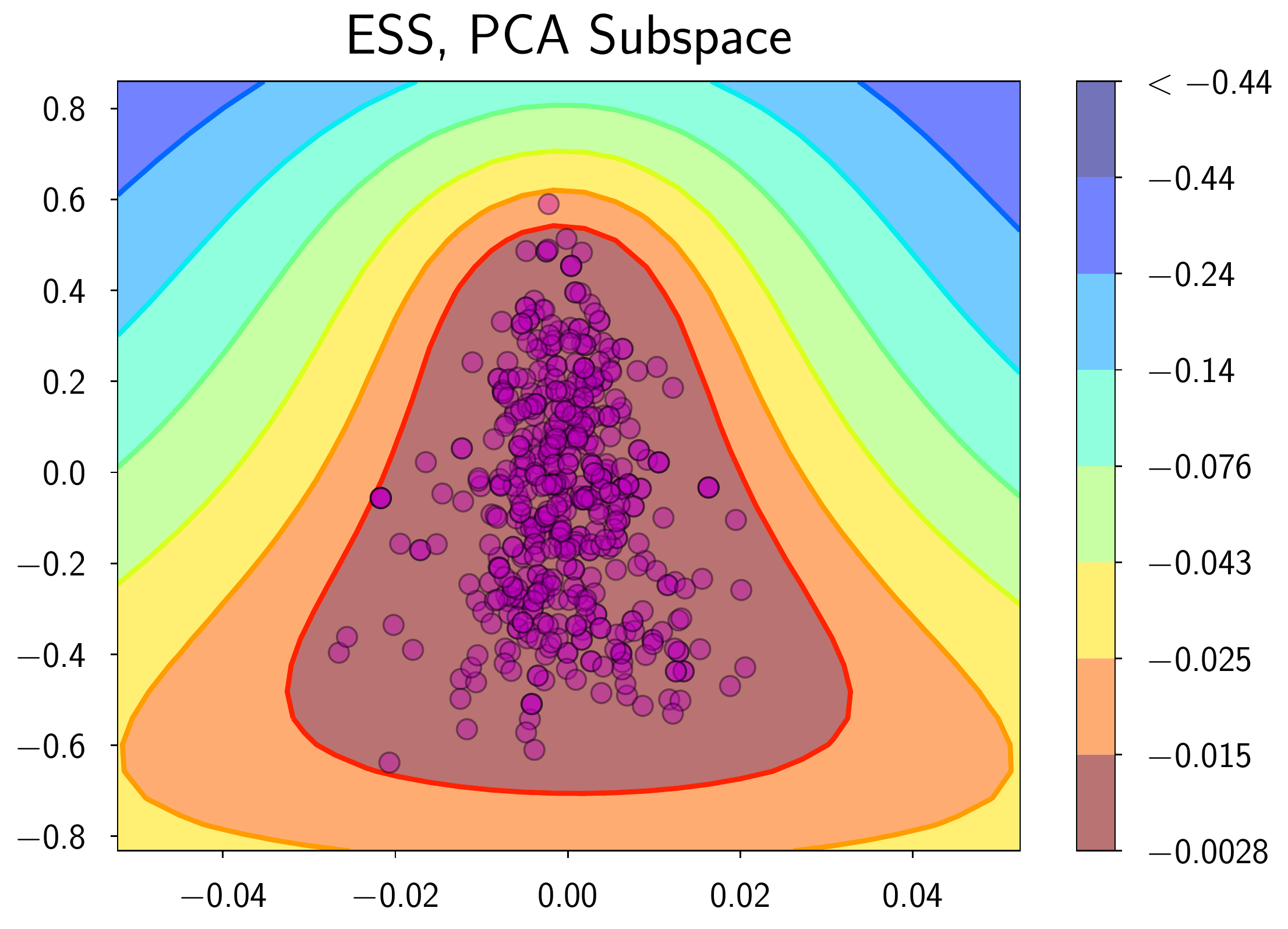}
	\end{subfigure}
	\quad
	\begin{subfigure}{0.30\textwidth}
		\centering
		\includegraphics[width=\textwidth]{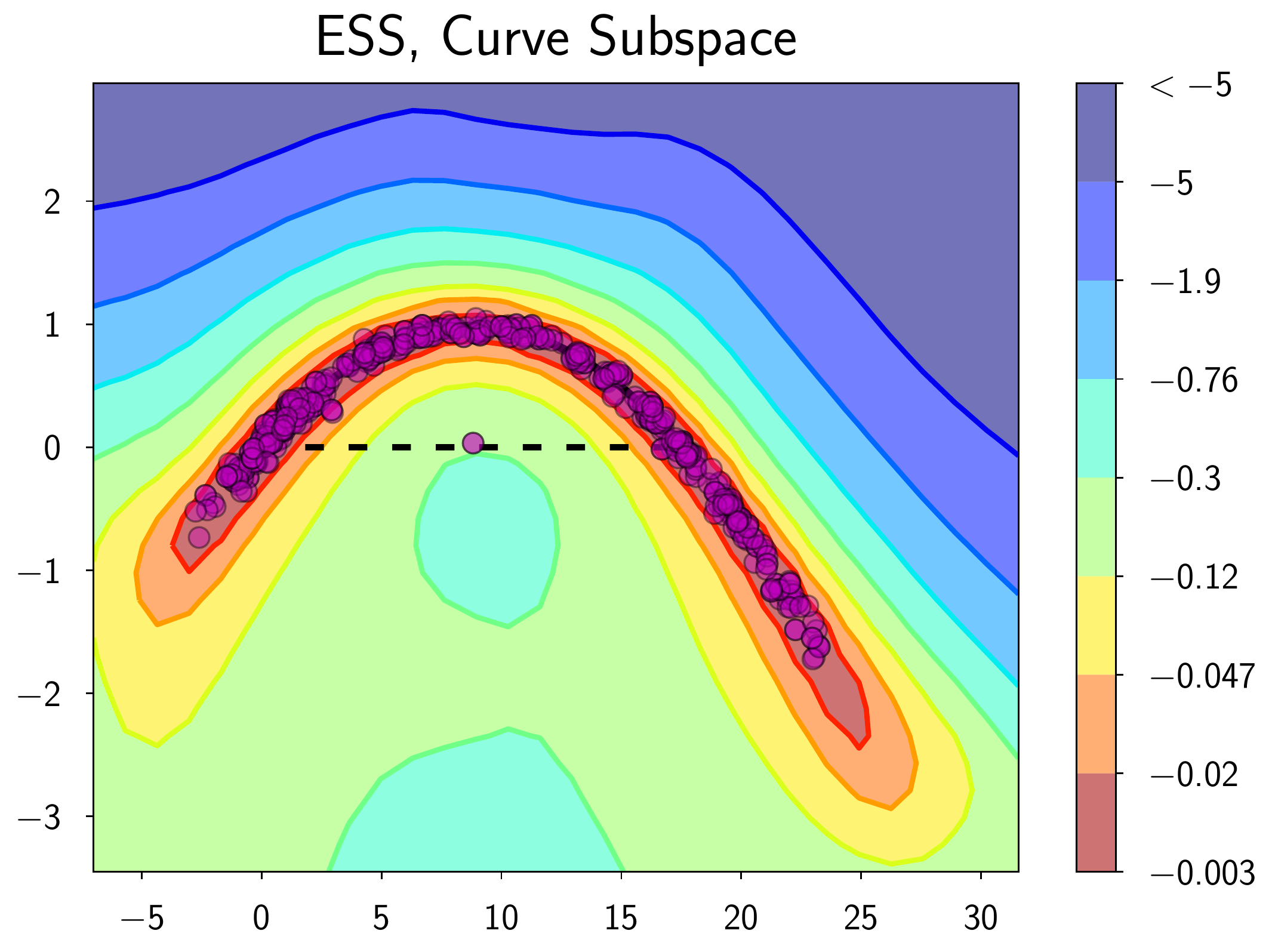}
	\end{subfigure}
	\caption{Posterior log-density surfaces and samples (magenta circles) for the synthetic regression problem across different subspaces and sampling methods.} 
	\label{fig:supp_toyreg_samples}
\end{figure*}

In Figure \ref{fig:supp_toyreg_trajectories} we present the predictive 
disribution plots for all the inference methods and subspaces. We additionally
visualize the samples over poterior density surfaces for each of the methods
in Figure \ref{fig:supp_toyreg_samples}.

\section{UCI REGRESSION EXPERIMENTAL DETAILS}

\subsection{SETUP}
In all experiments, we replicated over 20 trials reserving 90\% of the data for training and the other 10\% for testing, following the set-up of \citet{bui2016deep} and \citet{wilson2016deep}.

\subsubsection{Gaussian test likelihood}
\label{sec:testlike}

In Bayesian model averaging, we compute a Gaussian estimator $\mathcal N (y\vert\hat{\mu}, \hat\sigma^2)$ based on sample statistics\footnote{This is the same estimator used in \citet{wu2018fixing} and \citet{lakshminarayanan2017simple}.}, 
where $\hat{\mu} (x)= \frac{1}{J} \sum_{i=1}^J \mu(x; w_i)$, $\hat{\sigma}^2(x) = \frac{1}{J}\sum_{i=1}^J \left(\sigma^2(x; w_i) + \mu(x;w_i)^2\right) - \hat{\mu}(x)^2$, and $w_i$ are samples from the approximate posterior (see Section \ref{sec:sampling}). 

\subsubsection{Small Regression}
\label{sec:appendix_ucismall}

For the small UCI regression datasets, we use the architecture from \citet{wu2018fixing} with one hidden layer with 50 units. We manually tune learning rate and weight decay, and use batch size of $N / 10$ where $N$ is the dataset size. All models predict heteroscedastic uncertainty (i.e. output a variance). In Table \ref{tab:small_ll}, we compare subspace inference methods to deterministic VI (DVI, \citet{wu2018fixing}) and deep Gaussian processes with expectation propagation (DGP1-50  \citet{bui2016deep}).  ESS and VI in the PCA subspace outperform DVI on two out of five datasets.

\subsubsection{Large-Scale Regression}
\label{sec:appendix_ucilarge}

For the large-scale UCI regression tasks, we manually tuned hyper-parameters (batch size, learning rate, and epochs) to match the RMSE of the SGD DNN results in Table 1 of \citet{wilson2016deep}, starting with the parameters in the authors' released code.
We used heteroscedastic regression methods with a global variance parameter, e.g. the likelihood for a data point is given by: $\mathcal{N}(y_i, \mu(x_i; w), \sigma_g^2 + v(x_i; w)),$
optimizing $\sigma_g^2 = \text{softplus}(\tilde{\sigma}^2)$ in tandem with the network (which outputs both the mean and variance parameters).
$\sigma$ can be viewed as a global variance parameter, analogous to optimizing the jitter in Gaussian process regression. 
We additionally tried fitting models without a global variance parameter (e.g. standard heteroscedastic regression as is used in the SGD networks in \citet{wilson2016deep}), but found that they were typically more over-confident.

\begin{figure*}[!ht]
	\centering
    \includegraphics[width=0.99\textwidth]{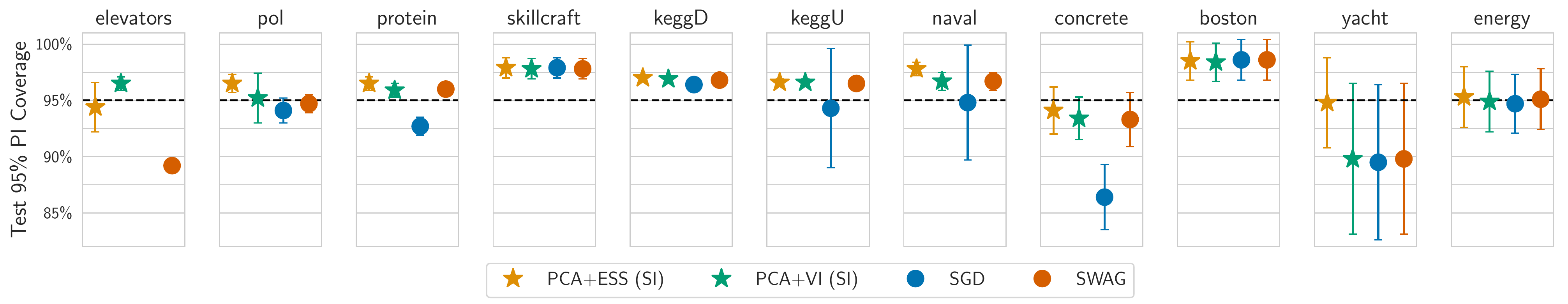}
	\caption{Coverage of 95\% prediction interval for models trained on UCI datasets. In most cases, subspace inference produces closer to 95\% coverage than models trained using SGD or SWAG.}
	\label{fig:biguci_calibration}
\end{figure*}
\begin{table*}[!h]
	\centering
	\caption{
	Unnormalized test log-likelihoods on small UCI datasets for Subspace Inference (SI), as well as direct comparisons to the numbers reported in deterministic variational inference (DVI, \citet{wu2018fixing}) and Deep Gaussian Processes with expectation propagation (DGP1-50, \citet{bui2016deep}), and variational inference (VI) with the re-parameterization trick \citep{kingma2015variational}.}
	\label{tab:small_ll}
	\resizebox{\textwidth}{!}{
		\begin{tabular}{llllllllll}
			\toprule
			dataset &           N &   D &       SGD &          PCA+ESS (SI) &           PCA+VI (SI) &         SWAG  & DVI & DGP1-50 & VI \\
			\midrule
			boston   &  506        & 13     &  -2.752 $\pm$ 0.132 &  -2.719 $\pm$ 0.132 &  -2.716 $\pm$ 0.133 &  -2.761 $\pm$ 0.132  & -2.41 $\pm$ 0.02 & \textbf{-2.33} $\pm$ 0.06 & -2.43 $\pm$0.03 \\
			concrete & 1030     &  8     &  -3.178 $\pm$ 0.198 &  -3.007 $\pm$ 0.086 &  \textbf{-2.994} $\pm$ 0.095 &  -3.013 $\pm$ 0.086   &  -3.06 $\pm$ 0.01  &     -3.13 $\pm$  0.03 & \textbf{-3.04} $\pm$0.02    \\
			energy  &    768       &   8  &  -1.736 $\pm$ 1.613 &  -1.563 $\pm$ 1.243 &  -1.715 $\pm$ 1.588 &  -1.679 $\pm$ 1.488      &  \textbf{-1.01} $\pm$ 0.06   &   -1.32 $\pm$ 0.03  & -2.38 $\pm$0.02 \\
			naval   &     11934   &   16  &   6.567 $\pm$ 0.185 &   6.541 $\pm$ 0.095 &   \textbf{6.708} $\pm$ 0.105 &   6.708 $\pm$ 0.105       &  6.29 $\pm$ 0.04    &  3.60 $\pm$ 0.33  & 5.87 $\pm$0.29\\
			yacht    &      308     & 6   &  -0.418 $\pm$ 0.426 &  \textbf{-0.225} $\pm$ 0.400 &  -0.396 $\pm$ 0.419 &  -0.404 $\pm$ 0.418     &   -0.47 $\pm$ 0.03   &  -1.39 $\pm$ 0.14 & -1.68 $\pm$0.04 \\
			\bottomrule
		\end{tabular}
	}
\end{table*}

\begin{table*}[!h]
	\centering
	\caption{
	RMSE on small UCI datasets. Subspace Inference (SI) typically performs comparably to SGD and SWAG.}
	\label{tab:small_rmse}
	\resizebox{0.6\textwidth}{!}{
		\begin{tabular}{lllll}
			\toprule
			{} &             SGD &         PCA+ESS (SI) &          PCA+VI (SI) &        SWAG \\
			\midrule
			boston   &  3.504 $\pm$ 0.975 & \textbf{3.453} $\pm$ 0.953 & 3.457 $\pm$ 0.951 & 3.517 $\pm$ 0.981 \\
			concrete &  5.194 $\pm$ 0.446 &  5.194 $\pm$ 0.448 &  \textbf{5.142} $\pm$ 0.418 &  5.233 $\pm$ 0.417 \\
			energy   &  1.602 $\pm$ 0.275 &  1.598 $\pm$ 0.274 &  \textbf{1.587} $\pm$ 0.272 &  1.594 $\pm$ 0.273 \\
			naval    &  \textbf{0.001} $\pm$ 0.000 &  \textbf{0.001} $\pm$ 0.000 &  \textbf{0.001} $\pm$ 0.000 &  \textbf{0.001} $\pm$ 0.000 \\
			yacht    &  0.973 $\pm$ 0.374 &  \textbf{0.972} $\pm$ 0.375 &  0.973 $\pm$ 0.375 &  0.973 $\pm$ 0.375 \\
			\bottomrule
		\end{tabular}
	}
\end{table*}

Following \citet{wilson2016deep}, for the UCI regression tasks with more than 6,000 data points, we used networks with the following structure: [1000, 1000, 500, 50, 2], while for skillcraft, we used a network with: [1000, 500, 50, 2].
We used a learning rate of $10^{-3}$, doubling the learning rate of bias parameters, a batch size of $400$, momentum of $0.9$, and weight decay of $4 \times 10^{-3}$, training for 200 epochs.
For skillcraft, we only trained for 100 epochs, using a learning rate of $5 \times 10^{-4}$ and for keggD, we used a learning rate of $10^{-4}$.
We used a standard normal prior with variance of $1.0$ in the subspace.

For all of our models, the likelihood for a data point is given by: $\mathcal{N}(y_i, \mu(x_i; w), \sigma + v(x_i; w)),$
optimizing $\sigma$ in tandem with the network (which outputs both the mean and variance parameters).
$\sigma$ can be viewed as a global variance parameter, analogous to optimizing the jitter in Gaussian process regression. 
We found fitting models without a global variance parameter often led to over-confident predictions.

In Table \ref{tab:bigrmse}, we report RMSE results compared to two types of approximate Gaussian processes \citep{salimbeni2018orthogonally,yang2015carte}; note that the results for OrthVGP are from Appendix Table F of \citet{salimbeni2018orthogonally} but scaled by the standard deviation of the respective dataset.
For the comparisons using Bayesian final layers \citep{riquelme2018deep}, we trained SGD nets with the same architecture and used the second-to-last layer (ignoring the final hidden unit layer of width two as it performed considerably worse) for the Bayesian approach and then followed the same hyper-parameter setup as in the authors' codebase \footnote{\url{https://github.com/tensorflow/models/tree/master/research/deep_contextual_bandits}} with $a = b = 6$ and $\lambda = 0.25$.

We repeated each model over 10 random train/test splits; each test set consisted of 10\% of the full dataset.
All data was pre-processed to have mean zero and variance one.
\renewcommand{\floatpagefraction}{.9}

\begin{table*}
  \centering
  \caption{Calibration on small-scale UCI datasets for Subspace Inference (SI). Bolded numbers are those closest to 95\% of the predicted coverage.}
  \label{tab:small_ca}
  \resizebox{0.65\textwidth}{!}{
\begin{tabular}{lllllll}
\toprule
{} &          N &   D &       SGD &         PCA+ESS (SI) &          PCA+VI (SI) &        SWAG \\
\midrule
boston   &  506        & 13    &    0.986 $\pm$ 0.018 & 0.985 $\pm$ 0.017 & \textbf{0.984} $\pm$ 0.017 & 0.986 $\pm$ 0.018 \\
concrete &   1030     &  8     &    0.864 $\pm$ 0.029 &  \textbf{0.941} $\pm$ 0.021 &  0.934 $\pm$ 0.019 &  0.933 $\pm$ 0.024   \\
energy   &    768       &   8    & 0.947 $\pm$ 0.026 &  0.953 $\pm$ 0.027 &  \textbf{0.949} $\pm$ 0.027 &  \textbf{0.951} $\pm$ 0.027            \\
naval    &     11934   &   16   &    \textbf{0.948} $\pm$ 0.051 &  0.978 $\pm$ 0.006 &  0.967 $\pm$ 0.008 &  0.967 $\pm$ 0.008        \\
yacht    &      308     & 6        &        0.895 $\pm$ 0.069 &  \textbf{0.948} $\pm$ 0.040 &  0.898 $\pm$ 0.067 &  0.898 $\pm$ 0.067    \\
\bottomrule
\end{tabular}
}
\end{table*}

\begin{table*}
	
	\centering
	\caption{RMSE comparison amongst methods on larger UCI regression tasks, as well as direct comparisons to the numbers reported in deep kernel learning with a spectral mixture kernel (DKL, \citep{wilson2016deep}), orthogonally decoupled variational GPs (OrthVGP, \citet{salimbeni2018orthogonally}), FastFood kernel GPs (FF, \citet{yang2015carte} from \citet{wilson2016deep}), and Bayesian final layers (NL, \citet{riquelme2018deep}). Subspace based inference typically outperforms SGD and approximate GPs and is competitive with DKL. }
	\label{tab:bigrmse}
	\resizebox{\textwidth}{!}{%
	\begin{tabular}{lrrllllllll}
		\toprule
		dataset &      N &   D &           SGD &  NL &    PCA+ESS (SI) &        PCA+VI (SI) &     SWAG & DKL & OrthVGP & FF \\
		\midrule
		elevators &  16599 &  18 &  $0.103\pm 0.035$ &  $0.101 \pm 0.002$ &  $0.089 \pm 0.002$ & $0.088 \pm 0.001$  &  $0.088 \pm 0.001$ & $\textbf{0.084}\pm0.02$ & 0.0952& $0.089\pm0.002$  \\
		keggD &  48827 &  20 &  $0.132 \pm 0.017$ &  $0.134 \pm 0.036$ &  $0.129 \pm 0.028$ &  $0.128 \pm 0.028$ &    $0.129 \pm 0.029$ & $\textbf{0.10}\pm0.01$ & 0.1198 & $0.12\pm0.00$  \\
		keggU &  63608 &  27 &  $0.186 \pm 0.034$  & $0.120 \pm 0.003$  & $0.160 \pm 0.043$ &  $0.160 \pm 0.043$ &   $ 0.160 \pm 0.043$ & $\textbf{0.11}\pm0.00$ & 0.1172 & $0.12\pm0.00$  \\
		protein &  45730 &   9 &  $0.436 \pm 0.011$ &  $0.447 \pm 0.012$ &  $0.425 \pm 0.017$ &  $0.418 \pm 0.021$ &   $ 0.415 \pm 0.018$ & $0.46\pm0.01$ & 0.46071 & $0.47\pm0.01$  \\
		skillcraft &   3338 &  19 &  $0.288\pm 0.014$ &  $0.253\pm 0.011$ &  $0.293\pm 0.015$ &  $0.293\pm 0.015$ &    $0.293\pm 0.015$ & $\textbf{0.25}\pm0.00$ & & $\textbf{0.25}\pm0.02$ \\
		pol &  15000 &  26 &  $3.900\pm 6.003$ &  $4.380 \pm 0.853$ &  $3.755 \pm 6.107$ &  $2.499 \pm 0.684$& $3.11\pm0.07*$ & 6.61749 &$4.30\pm0.2$  \\
		\bottomrule
	\end{tabular}
	}
	
\end{table*}

\begin{table*}
	\centering
	\caption{Normalized test log-likelihoods on larger UCI datasets. Subspace methods outperform an approximate GP approach (OrthVGP), SGD, and Bayesian final layers (NL), typically often out-performing SWAG.}
	\label{tab:bigtestll}
	\resizebox{\textwidth}{!}{%
\begin{tabular}{lrrllllll}
	\toprule
	dataset &      N &   D &            SGD &        PCA+ESS (SI) &         PCA+VI (SI) &       SWAG & OrthVGP & NL \\
	\midrule
	elevators &  16599 &  18 &  $-0.538 \pm 0.108$ &  $-0.351\pm 0.030$ &  $-0.325 \pm 0.019$ &  $-0.374 \pm 0.021$ & -0.4479 & $-0.698 \pm 0.039$ \\
	keggD &  48827 &  20 &   $1.012\pm0.154$ &   $1.074 \pm 0.034$ &   $1.085 \pm 0.031$ &   $1.080 \pm 0.035$ & \textbf{1.0224} & $0.935 \pm 0.265$ \\
	keggU &  63608 &  27 &   $0.602 \pm 0.224$ &   $0.752 \pm 0.025$ &   $0.757 \pm 0.028$ &   $0.749 \pm 0.029$ & 0.7007 & $0.670 \pm 0.038$ \\
	protein &  45730 &   9 &  $-0.854 \pm 0.085$ &  $-0.734 \pm 0.063$ &  $-0.712 \pm 0.057$ &  $-0.700 \pm 0.051$ & -0.9138 & $-0.884\pm 0.025$ \\
	skillcraft &   3338 &  19 &  $-1.162\pm 0.032$ & $-1.181 \pm 0.033$ &  $-1.179 \pm 0.033$ & $ -1.180 \pm 0.033$ & & $-1.002 \pm 0.050$ \\
	pol &  15000 &  26 &   $1.073 \pm 0.858$ &   $-0.185 \pm 2.779 $&   $1.764 \pm 0.271$ &   $1.533 \pm 1.084$ & 0.1586 & $-2.84 \pm 0.226$ \\
	\bottomrule
\end{tabular}
}

\end{table*}

\begin{table*}[!h]
	
	\centering
	\caption{Calibration on large-scale UCI datasets. Bolded numbers are those closest to 95 \% of the predicted coverage).}
	\label{tab:bigcalibration}
	\resizebox{0.7\textwidth}{!}{
	\begin{tabular}{llllllll}
		\toprule
		dataset &      N &   D &           SGD &    NL &   PCA+ESS (SI) &        PCA+VI (SI) &      SWAG \\
		\midrule

		elevators & 16599 & 18 & $0.857 \pm 0.031$ &  $0.834 \pm 0.012$ & $0.944 \pm 0.022$ &  $0.965 \pm 0.006$ & $0.892\pm 0.006$ \\
		keggD  & 48827 & 20 & $0.964 \pm 0.005$ &  $0.915 \pm 0.210$ &  $0.970 \pm 0.002$ & $0.969 \pm 0.002$ & $0.968 \pm 0.002$ \\
		keggU  & 63608 & 27 & $0.943 \pm 0.053$ & $0.964 \pm 0.002$ & $0.966 \pm 0.003$ & $0.966 \pm 0.003$ & $0.965 \pm 0.003$ \\
		protein  & 45730 &  9 & $0.927 \pm 0.008$ & $0.921 \pm 0.005$ & $0.965 \pm 0.006$ &  $0.959 \pm 0.006$ &  $0.960\pm0.006$ \\
		pol  & 15000 & 26 & $0.885\pm 0.118$  & $0.884\pm 0.118$ & $0.979\pm 0.012$ &  $0.979\pm 0.010$ &  
		$0.971\pm 0.014$ \\
		skillcraft   & 3338  & 19 & $0.979\pm 0.009$ &  $0.939\pm 0.013$ & $0.979\pm 0.009$ &  $0.978 \pm 0.009$ & $0.978\pm 0.009$ \\
		
		\bottomrule
	\end{tabular}}
	
\end{table*}

\begin{table*}
	\centering
	\caption{NLL for various versions of subspace inference, SWAG, temperature scaling, and dropout.}
	\label{tab:cifar_nll}
	\resizebox{\textwidth}{!}{%
		\begin{tabular}{lllllllll}
			\toprule
			Dataset   & Model           & PCA + VI (SI)               & PCA + ESS (SI)               & SWA               & SWAG            & KFAC-Laplace       & SWA-Dropout          & SWA-Temp          \\
			\midrule                                                                 
			CIFAR-10  & VGG-16          & $0.2052\pm0.0029$      & $0.2068\pm0.0029$      & $0.2621\pm0.0104$ & $\bm{0.2016}\pm0.0031$ & $0.2252\pm0.0032$ & $0.2328\pm0.0049$ & $0.2481\pm0.0245$      \\
			CIFAR-10  & PreResNet-164   & $0.1247\pm0.0025$      & $0.1252\pm0.0018$      & $0.1450\pm0.0042$ & $\bm{0.1232}\pm0.0022$ & $0.1471\pm0.0012$ & $0.1270\pm0.0000$ & $0.1347\pm0.0038$      \\
			CIFAR-10  & WideResNet28x10 & $0.1081\pm0.0003$      & $0.1090\pm0.0038$      & $0.1075\pm0.0004$ & $0.1122\pm0.0009$      & $0.1210\pm0.0020$ & $0.1094\pm0.0021$ & $\bm{0.1064}\pm0.0004$ \\
			CIFAR-100 & VGG-16          & $0.9904\pm0.0218$      & $1.015\pm0.0259$      & $1.2780\pm0.0051$ & $\bm{0.9480}\pm0.0038$ & $1.1915\pm0.0199$ & $1.1872\pm0.0524$ & $1.0386\pm0.0126$      \\
			CIFAR-100 & PreResNet-164   & $\bm{0.6640}\pm0.0025$ & $0.6858\pm0.0052$ & $0.7370\pm0.0265$ & $0.7081\pm0.0162$      & $0.7881\pm0.0025$ &                   & $0.6770\pm0.0191$ \\
			CIFAR-100 & WideResNet28x10 & $\bm{0.6052}\pm0.0090$ & $0.6096\pm0.0072$ & $0.6684\pm0.0034$ & $0.6078\pm0.0006$ & $0.7692\pm0.0092$ & $0.6500\pm0.0049$ & $0.6134\pm0.0023$      \\
			\bottomrule
		\end{tabular}%
	}	
\end{table*}

\begin{table*}
	\centering
	\caption{Accuracy for various versions of subspace inference, SWAG, temperature scaling, and dropout.}
	\label{tab:cifar_acc}
	\resizebox{\textwidth}{!}{
		\begin{tabular}{lllllllll}
			\toprule
			Dataset   & Model           & PCA + VI (SI)            & PCA + ESS (SI)          & SWA                   & SWAG               & KFAC-Laplace   & SWA-Dropout         & SWA-Temp       \\
			\midrule                                                                
			CIFAR-10  & VGG-16          & $93.61\pm0.02$      & $\bm{93.66}\pm0.08$ & $93.61\pm0.11$       & $93.60\pm0.10$      & $92.65\pm0.20$ & $93.23\pm0.36$      & $93.61\pm0.11$ \\
			CIFAR-10  & PreResNet-164   & $95.96\pm0.13$      & $95.98\pm0.09$      & $96.09\pm0.08$       & $96.03\pm0.02$      & $95.49\pm0.06$ & $\bm{96.18}\pm0.00$ & $96.09\pm0.08$ \\
			CIFAR-10  & WideResNet28x10 & $96.32\pm0.03$      & $ 96.38\pm0.05$     & $\bm{96.46}\pm0.04$  & $96.32\pm0.08$      & $96.17\pm0.00$ & $96.39\pm0.09$      & $96.46\pm0.04$ \\
			CIFAR-100 & VGG-16          & $\bm{74.83}\pm0.08$ & $74.62\pm0.37$      & $74.30\pm0.22$       & $74.77\pm0.09$      & $72.38\pm0.23$ & $72.50\pm0.54$      & $74.30\pm0.22$ \\
			CIFAR-100 & PreResNet-164   & $80.52\pm0.18$ & $\bm{80.54}\pm0.13$      & $80.19\pm0.52$       & $79.90\pm0.50$      & $78.51\pm0.05$ &                     & $80.19\pm0.52$ \\
			CIFAR-100 & WideResNet28x10 & $\bm{82.63}\pm0.26$ & $82.49\pm0.23$      & $82.40\pm0.16$       & $82.23\pm0.19$      & $80.94\pm0.41$ & $82.30\pm0.19$      & $82.40\pm0.16$ \\
			\bottomrule
		\end{tabular}
	}
\end{table*}

\section{IMAGE CLASSIFICATION RESULTS}

For the experiments on CIFAR datasets we follow the framework of
\citet{maddox2019simple}. We report the negative log-likelihood and accuracy
for our method and baselines in Tables \ref{tab:cifar_nll} and \ref{tab:cifar_acc}.

\subsection{EFFECT OF TEMPERATURE}
\label{sec:temp_effect}

\begin{figure*}
	\centering
	\begin{subfigure}{0.24\textwidth}
		\centering
		\includegraphics[width=1\textwidth]{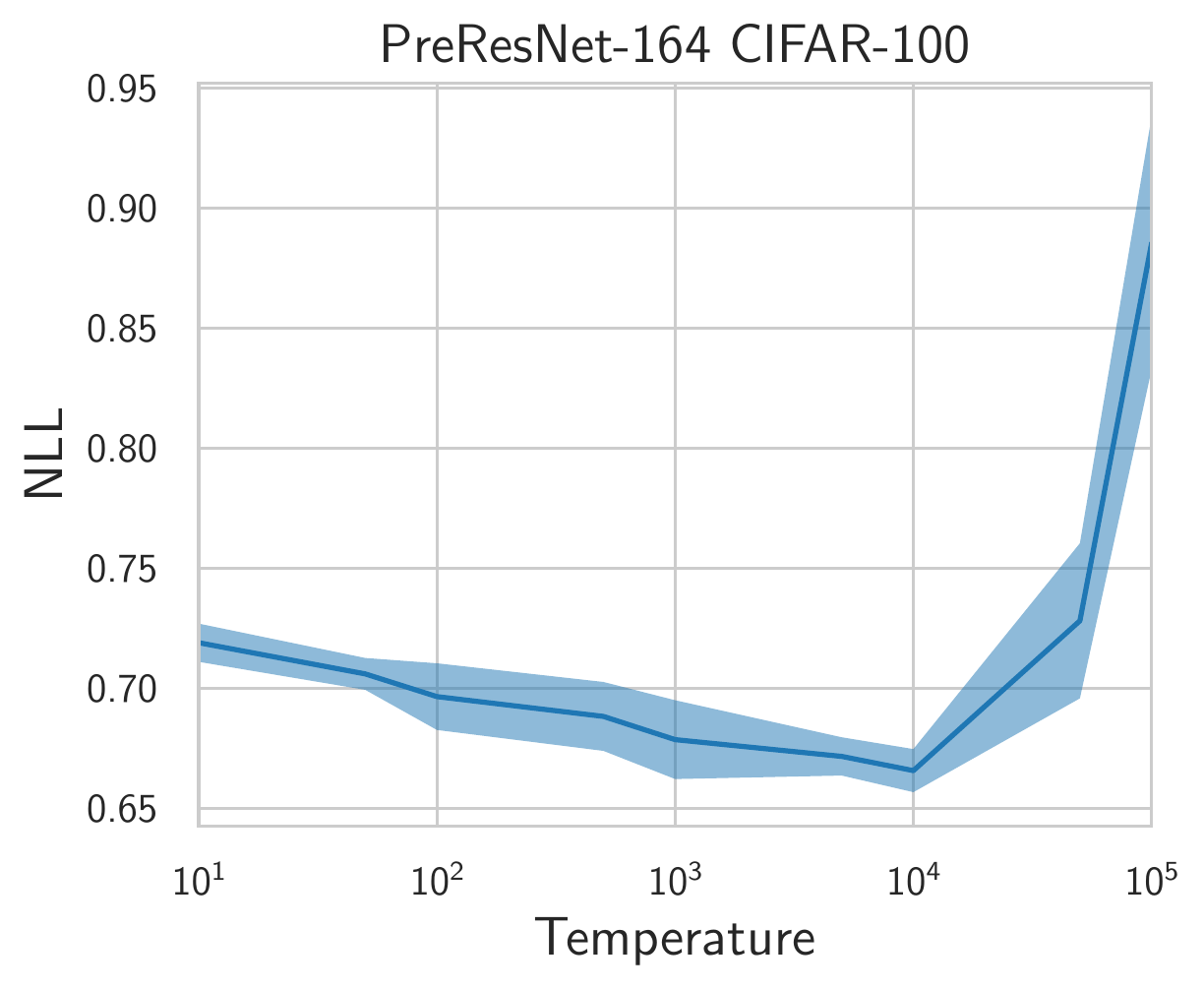}
		\caption{}
	\end{subfigure}
	\begin{subfigure}{0.24\textwidth}
		\centering
		\includegraphics[width=1\textwidth]{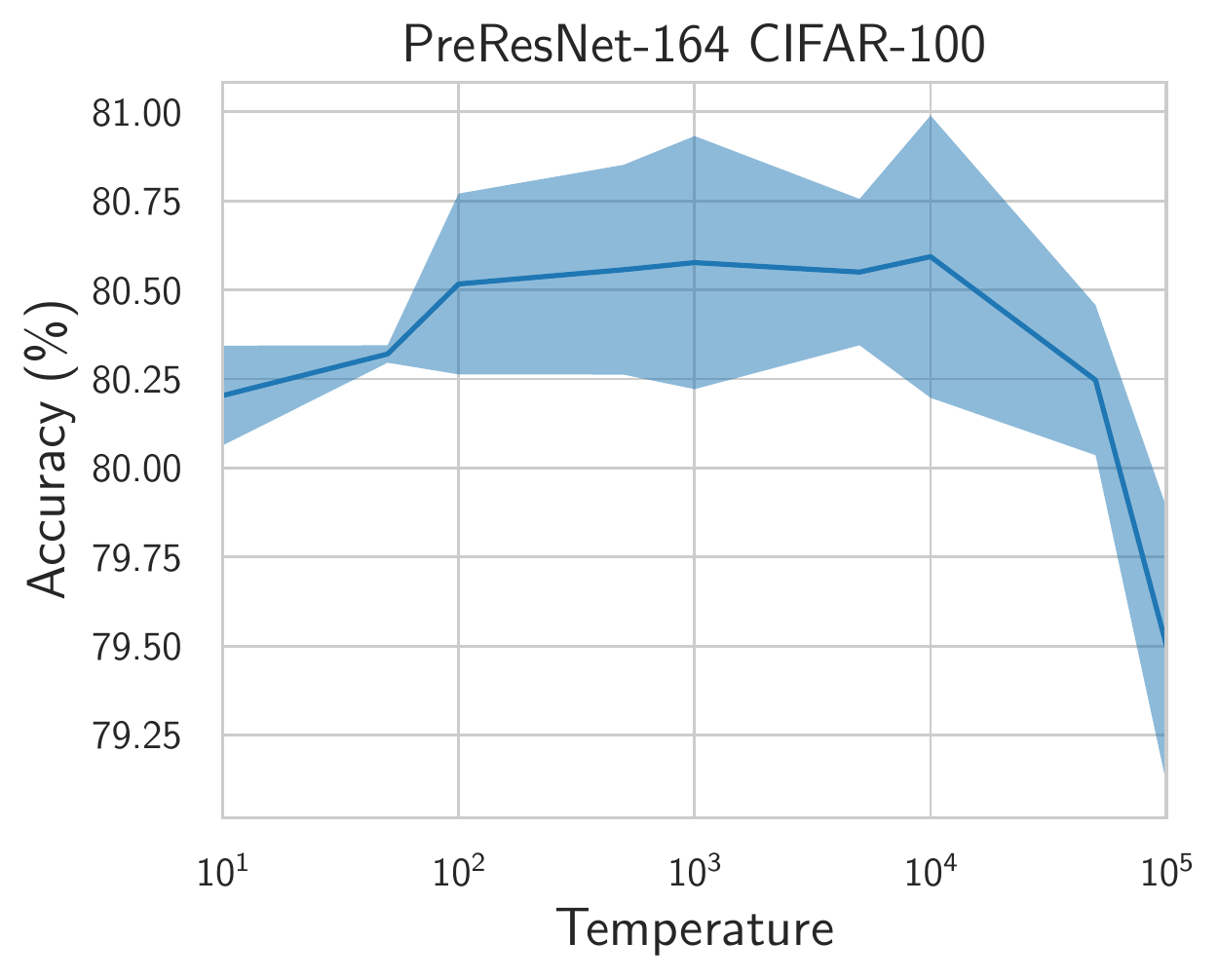}
		\caption{}
	\end{subfigure}
	\begin{subfigure}{0.24\textwidth}
		\centering
		\includegraphics[width=1\textwidth]{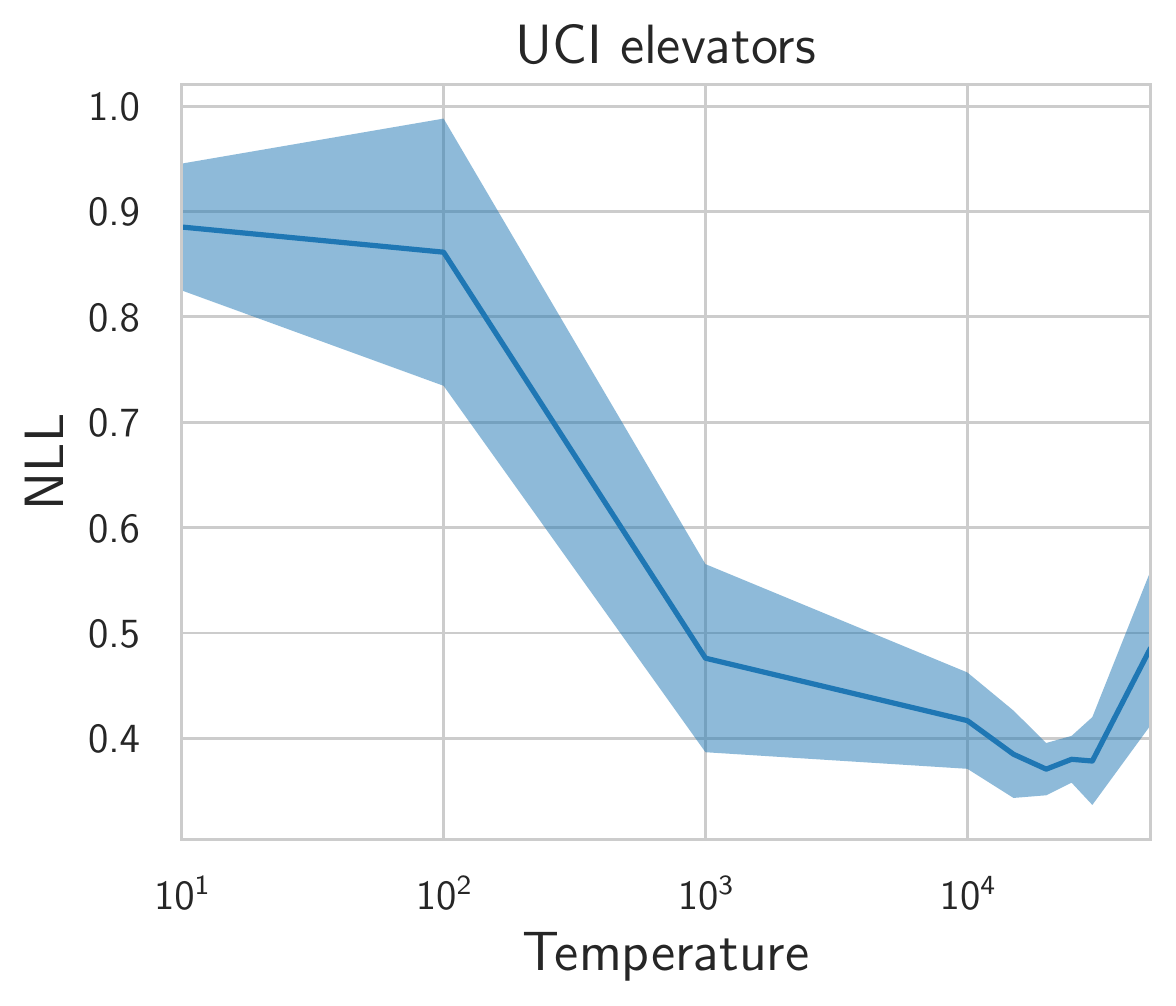}
		\caption{}
	\end{subfigure}
	\begin{subfigure}{0.24\textwidth}
		\centering
		\includegraphics[width=1\textwidth]{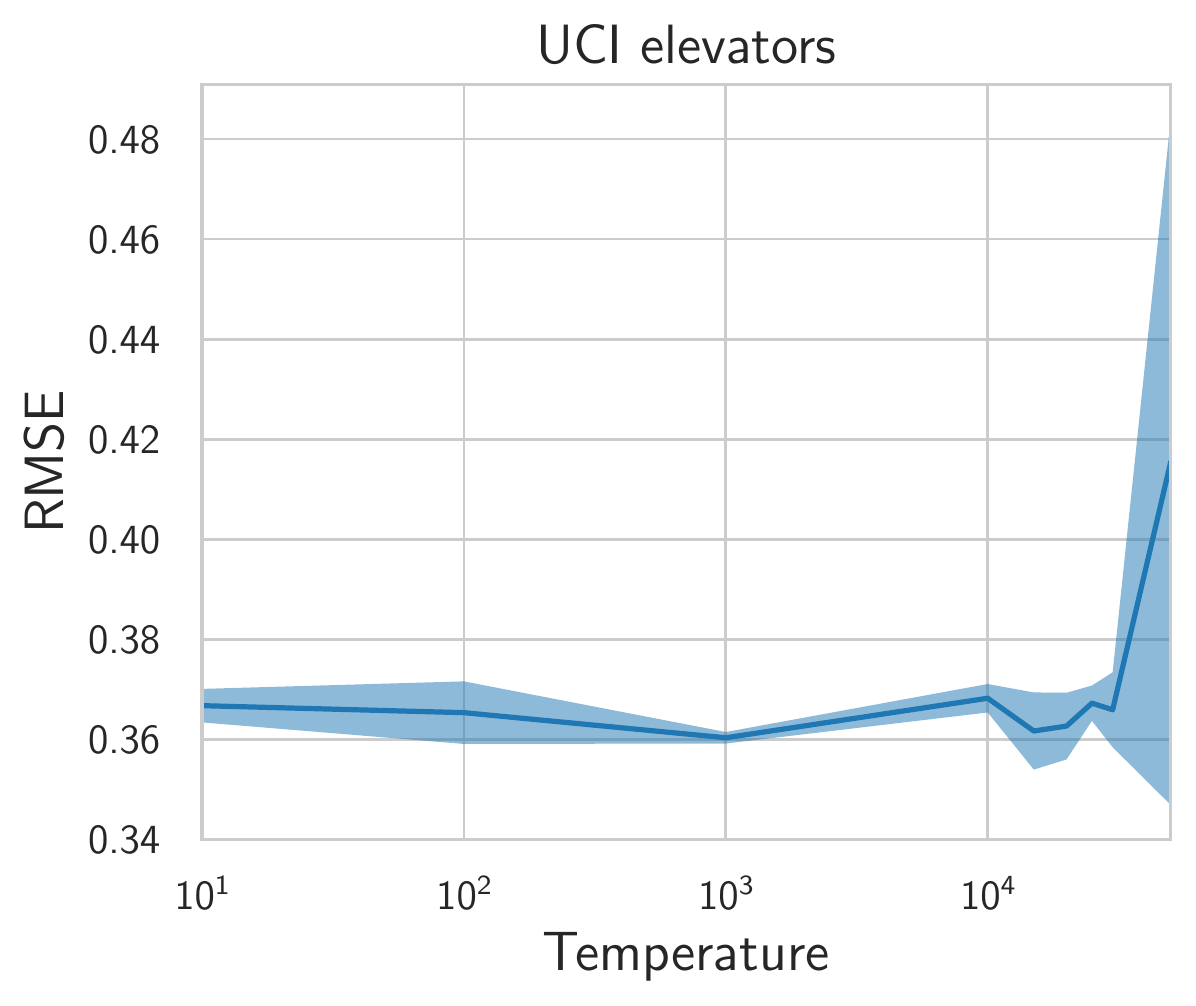}
		\caption{}
	\end{subfigure}
	
	\caption{\textbf{(a)}: Test negative log-likelihood and \textbf{(b)}: accuracy as a function of temperature in \eqref{eq:tempered_post} for PreResNet-164 on CIFAR-100. \textbf{(c)}: Test negative log-likelihood and \textbf{(d)}: RMSE as a function of temperature for our regression architecture (see Section \ref{sec:ucireg}) on UCI Elevators.
	We used ESS in a $5$-dimensional PCA subspace to construct this plot. 
	The dark blue line and shaded region show the mean $\pm$ 1 standard deviation over $3$ independent runs of the procedure.}
	\label{fig:temp_effect}
\end{figure*}

We study the effect of the temperature parameter $T$ defined in \eqref{eq:tempered_post} on the performance of subspace inference. 
We run elliptical slice sampling in a $5$-dimensional PCA subspace for a PreResNet-164 on CIFAR-100. 
We show test performance as a function of the temperature parameter in Figure \ref{fig:temp_effect} panels (a) and (b). 
Bayesian model averaging achieves strong results in the range $10^3 \le T \le 10^4$. 
We also observe that the value $T$ has a larger effect on uncertainty estimates and consequently NLL than on predictive accuracy.

We then repeat the same experiment on UCI elevators using the setting described in Section \ref{sec:uci_large}.
We show the results in Figure \ref{fig:temp_effect} panels (c), (d). 
Again, we observe that the performance is almost constant and close to optimal in a certain range of temperatures, and
the effect of temperature on likelihood is larger compared to RMSE.

\end{document}